\documentclass[acmlarge, screen=true]{acmart}

\renewcommand\footnotetextcopyrightpermission[1]{} 
\makeatletter
\let\@authorsaddresses\@empty
\makeatother

\makeatletter
\def\@ACM@checkaffil{
    \if@ACM@instpresent\else
    \ClassWarningNoLine{\@classname}{No institution present for an affiliation}%
    \fi
    \if@ACM@citypresent\else
    \ClassWarningNoLine{\@classname}{No city present for an affiliation}%
    \fi
    \if@ACM@countrypresent\else
        \ClassWarningNoLine{\@classname}{No country present for an affiliation}%
    \fi
}
\makeatother
\AtBeginDocument{%
  \providecommand\BibTeX{{%
    \normalfont B\kern-0.5em{\scshape i\kern-0.25em b}\kern-0.8em\TeX}}}

\setcopyright{acmlicensed}
\copyrightyear{2025}
\acmYear{2025}
\acmDOI{XXXXXXX.XXXXXXX}

\usepackage{xcolor}
\usepackage{comment}
\usepackage{lipsum}
\usepackage{titlesec}
\usepackage{mathtools}
\usepackage{comment}
\usepackage{array}
\usepackage{array}
\usepackage{makecell}
\usepackage{multirow}
\usepackage{booktabs}
\usepackage{pifont}
\usepackage{enumitem}
\setlist{nolistsep}
\usepackage[labelfont=bf]{caption}

\usepackage{amssymb}

\DeclareUnicodeCharacter{2212}{-}

\newcommand{\cmark}{\ding{51}}%
\newcommand{\xmark}{\ding{55}}%

\begin{document}

\title{Energy-Aware Deep Learning on Resource-Constrained Hardware}

\author{Josh Millar, Hamed Haddadi}
\affiliation{%
  \institution{Imperial College London}
}

\author{Anil Madhavapeddy}
\affiliation{%
  \institution{University of Cambridge}
}

\renewcommand{\shortauthors}{Millar et al.}

\begin{abstract}
The use of deep learning (DL) on Internet of Things (IoT) and mobile devices offers numerous advantages over cloud-based processing. However, such devices face substantial energy constraints to prolong battery-life, or may even operate intermittently via energy-harvesting. Consequently, \textit{energy-aware} approaches for optimizing DL inference and training on such resource-constrained devices have garnered recent interest. We present an overview of such approaches, outlining their methodologies, implications for energy consumption and system-level efficiency, and their limitations in terms of supported network types, hardware platforms, and application scenarios. We hope our review offers a clear synthesis of the evolving energy-aware DL landscape and serves as a foundation for future research in energy-constrained computing.
\end{abstract}

\maketitle


\vspace{-0.3cm}
\section{Introduction}
The Internet of Things (IoT) has become a vital component of our everyday lives, with a forecasted 30 billion smart devices connecting and controlling our homes, urban environments, and factories by 2027 \cite{iotanal}. 

IoT devices generate massive amounts of data, which often requires machine learning (ML) -based processing. Deep Neural Networks (DNNs) and other ML approaches are notoriously resource-intensive and so, traditionally, cloud-based processing has been relied upon to handle their computational load. However, considerable recent work has focused on enabling on-device deep learning (DL) inference and training. This offers numerous advantages over cloud-based processing, including reduced and predictable latency \cite{neurosurgeon}, independent of network congestion saturation, minimized bandwidth \cite{boomerang}, and improved data privacy. 

IoT, mobile, and other edge devices often face strict constraints in terms of memory and compute resources. Hence, numerous works focus on optimizing DNNs subject to such constraints. These generally focus on inference, employing lightweight DNN architectures \cite{lightweight1, lightweight2, lightweight3} and compression methods \cite{comp1, comp2, comp3}, as well as strategies for adapting inference dynamically based on input-complexity and workload. These strategies include multi-exit networks \cite{ee}, inference offloading \cite{off}, and the partitioning of inference between device and cloud \cite{partitioning}. Additional works focus on enabling on-device DNN training \cite{devt1, devt2, devt3, devt4, devt5, growth, 256KB} by regulating layer-wise growth \cite{growth} or eliminating redundant weights/activations to minimize updates \cite{256KB}. 

However, IoT devices also operate on strict energy budgets to prolong battery life. The energy consumption of a DNN is not directly proportional to its memory or computational footprint \cite{eyeriss} – so, while the above optimization methods improve DNN efficiency, they generally fail to optimize for energy-usage. This is vital, as energy-usage can impose a larger constraint on particular devices or applications than memory/compute ($e.g.$, if considering battery-less devices and/or deployment in remote or hard-to-reach environments).

We characterize \textit{energy-aware} DL as approaches for optimizing the design, inference, or training of DNNs in terms of energy consumption. Note that energy-awareness is not limited to energy-efficiency alone, also including approaches that can regulate inference energy consumption based on energy-availability, workload, and input complexity. This survey provides a concise review of such approaches, focusing on those with potential for deployment on highly-constrained IoT devices, particularly microcontroller units (MCUs). 
\subsection{Related Works and Contributions}

Numerous survey works explore ML and DL for resource-constrained computing \cite{tinymlreview}, yet specifically \textit{energy-aware} approaches have received comparatively limited attention.
Tekin et al. offer the only comprehensive survey addressing on-device ML from an energy consumption perspective \cite{energyreview}, but their IoT-centric review emphasizes use-case taxonomy and tooling over optimization, and overlooks energy-harvesting and intermittent computing.

Various benchmarking efforts also exist in this domain. MLPerfPower \cite{mlperfpower}, arguably the most prominent initiative, provides a benchmarking suite for standardizing the energy measurement of ML workloads at power scales ranging from microwatts to megawatts. It covers diverse platforms including mobile phones and specialized TPUs, like Nvidia's Jetson family, but has limited coverage on MCUs and ultra-low-power devices. Its benchmarks remain neither particularly up-to-date nor comprehensive regarding emerging energy-aware techniques.

Our work instead provides a concise yet detailed overview of algorithm-based techniques for energy-aware DNN design, training, inference, and deployment policies. We emphasize practical methodologies tailored to IoT devices operating under severe power constraints or intermittency conditions, offering a focused perspective on the evolving \textit{energy-aware} DL landscape.
\vspace{-0.2cm}
\subsection{Structure}
We cover energy-aware DNN design paradigms (\S\ref{section:design}), methods for energy-adaptive DNN inference (\S\ref{section:adaptive}), and on-device DNN training (\S\ref{section:perz}). We then explore various real-world DL applications in which energy-awareness is imperative and requires novel approaches; these include federated learning (FL) (\S\ref{section:fl}) and DL on intermittent energy-harvesting devices (\S\ref{section:intermittent}). Finally, \S\ref{section:hardware} outlines the various hardware platforms employed in the reviewed works. Each section offers insights into potential future directions. \S\ref{section:fw} provides a high-level overview of these. 



\section{Energy-Aware DNN Design}
\label{section:design}

Including energy as a design-time metric is key for guiding the design and optimization of efficient DNNs. However, balancing energy-efficiency with performance is a layered design challenge and requires sophisticated optimization techniques. The bulk of existing work has predominantly focused on using the number of DNN weights and/or operations as proxy for energy consumption; in particular multiply-and-accumulate (MAC) operations, which often account for over 99\% of total CNN operations \cite{alexnet, dally, xiangyu}. Unfortunately, while these proxies are correlated to the memory and computational footprints of DNNs, they do not accurately reflect their energy consumption ($e.g.$, SqueezeNet \cite{squeezenet} contains 50x fewer MACs than AlexNet \cite{alexnet}, yet exhibits greater energy consumption on various platforms \cite{eap, oscar}). This is because DNN energy consumption is enormously impacted by data movement; access to memory distant from the processor may cost 10 to 100+ times more than an FPU/ALU operation \cite{alu}. GoogLeNet’s \cite{googlenet} energy consumption breakdown reveals that only 10\% is attributed to network computation and the rest consumed by the movement of feature-maps \cite{eems}. Further, differences in memory architecture between platforms result in large variations in the comparative energy-efficiency of DNNs across devices. Therefore, energy-aware and platform-agnostic DNN compression and optimization methods \cite{deepcomp} are vital. 
Such methods can be largely categorised into the following: 
\begin{itemize}[label=--]
    \item \textit{pruning}: removing redundant weights from a network.
    \item \textit{quantization}: reducing the bit-precision of network weights (e.g., from 32-bit float to 8-bit integer).
    \item \textit{weight sharing}: reusing weights across network layers (e.g., shared filters in convolutional layers).
    \item \textit{knowledge distillation}: training a smaller network to emulate a larger one by minimizing the entropy, distance, or divergence between their outputs.
    \item \textit{low-rank decomposition}: approximating network weights with products of lower-rank matrices.
\end{itemize}

These optimisations generally reduce network memory footprint by removing or compressing DNN weights. However, as mentioned above, memory-focused compression does not guarantee optimal DNN energy consumption. There has been recent interest in energy-aware variants of such methods; among these, pruning and quantization have received the most attention in the context of energy-aware compression.


\footnotetext{\url{https://www.st.com/resource/en/datasheet/stm32l4r5zi.pdf}}
\subsection{Design-Time Optimizations}
\label{section:static}

Given the recognition that memory-focused compression does not guarantee optimal DNN energy consumption, there is a growing interest in the development of energy-aware variants of such compression methods that are directly optimized for energy use.

In energy-aware pruning (EAP), the pruning order is determined based on ranked layer-wise energy consumption \cite{eap}. EAP approaches vary predominantly by their means of estimating layer-wise energy consumption; approaches include joint estimation based on computation cost ($i.e.$ MACs) and memory accesses \cite{tien_ju2}, using CNN filter importance (nuclear-norm derived from SVD) as proxy \cite{nuke_norm}, and using energy-measurements derived from real hardware measurements \cite{tien_ju1}. GENESIS \cite{sonic} enables layer-wise DNN compression via rank decomposition and pruning, incorporating user-defined operator energy-costs. However, the effectiveness of EAP is limited by the fact that CNN energy consumption is dominated by convolutional operations and their extensive data movement, while pruning techniques are most effective on fully-connected (FC) layers.

There is also substantial recent work focusing on energy-aware quantization (EAQ) in the realm of in-memory DL accelerators. These accelerators utilize non-volatile memory to improve CNN energy-efficiency by performing convolution operations as analog-domain matrix-vector multiplications. This effectively reduces data movement. ECQ \cite{ecq} utilizes a genetic approach for exploring layer-wise weight/activation bitwidth(s), optimizing dynamic in-memory energy consumption at negligible performance loss. The ECQ approach requires only a network’s weight/activation distributions in order to estimate its energy consumption, making it applicable across various DNNs. The utilization of in-memory accelerators offers vast potential in reducing DNN inference energy consumption by minimizing data movement. However, such accelerators are not yet widely applicable on memory-constrained devices; Kim et al. \cite{pim_overview} provide a detailed overview of in-memory techniques for DL.


The key limitation of EA compression methods lies in their accuracy in predicting the layer/operation-wise energy consumption of DNNs \textit{pre-execution}. \S\ref{section:iee} provides an overview of existing energy-estimation methods, and their supported hardware platforms and DNN operators. Future work should additionally explore the impact of hardware variation on non-uniform quantization techniques.

\subsection{Neural Architecture Search}

\begin{table*}
  \centering
  \caption{\textmd{An overview of energy-aware neural architecture search methods. A dash indicates an out-of-scope area.}}
  \resizebox{13cm}{!}{\begin{tabular}{c|rrrr}
  \Xhline{1pt}
  \thead{\textbf{Method}} & \thead{\textbf{Eval. (Base) Arch. Space}} &	\thead{\textbf{Constrained Eval. Hardware }} & \textbf{Approach} & \thead{\textbf{Energy-Estimation}} \\
    \Xhline{1pt}
    \textbf{\cite{nascaps}}  &	\makecell[r]{DeepCaps \cite{deepcaps}, \\ CapsNet \cite{capsnet}}  &	\makecell[c]{-}  &	Evolutionary  &	\makecell[r]{MACs, \\ memory-accesses} \\
    \hline \\
    \textbf{\cite{split2}}  &	MobileNetV1/V2 \cite{mobilenetv1, mobilenetv2}  & \makecell[c]{-}  &	Gradient  &	\makecell[r]{FLOPs} \\
    \hline
    \textbf{\cite{chamnet}}  &	\makecell[r]{MobileNetV2, \\ MNASNet \cite{mnasnet}}  &	\makecell[r]{Snapdragon 835 (CPU)}  &	Evolutionary  &	\makecell[r]{Hardware-measurements} \\
    \hline
    \textbf{\cite{harvnet}}  &	\textmu NAS \cite{unas}  &	\makecell[r]{TI MSP430FR5994 (MCU)}  &	Super-Net  &	\makecell[r]{MACs} \\
    \hline
    \textbf{\cite{etnas}}  &	\makecell[r]{MobileNetV2, \\ ResNet50 \cite{resnet}, \\ YOLOv5 \cite{yolov5}}  &	\makecell[r]{Huawei Atlas200DK (NPU), \\ Zynq ZCU102 (FPGA)}  &	Super-Net  &	\makecell[r]{Hardware-measurements} \\
    \hline
  \end{tabular}}
  \label{tab:1}
  \vspace{-0.3cm}
\end{table*}

Traditional works enabling on-device inference focused on hand-crafted DNN architectures, including SqueezeNet \cite{squeezenet}, MobileNet \cite{mobilenetv1, mobilenetv2}, and EfficientNet \cite{efficientnet}. The design of these architectures required substantial manual tuning, combined with ingenuity; and while efficient, they lack versatility for diverse task-specific optimizations. Neural Architecture Search (NAS) methods revolutionized NN design by automating the discovery of NN architectures that optimize various performance metrics for a given task, while also considering computational/hardware constraints. Numerous works have demonstrated NAS's capability to reveal NN designs with state-of-the-art performance \cite{nas_survey}. Typical NAS methods explore the architecture space using evolutionary \cite{evolutionary} or reinforcement-learning (RL) \cite{reinforcement} approaches. These methods require vast computational resources and GPU hours, rendering them infeasible in most real-world applications. Consequently, more recent NAS methods use proxies for architecture evaluation to reduce search complexity at the expense of (marginally) suboptimal results.  Favoured among these methods is Differentiable NAS (DNAS), which utilizes gradient descent to optimize both DNN weights and parameters. Typically, DNAS employs super-nets, large DNNs that include various implementations of subnetworks, and explores the architecture space by evaluating subnetwork paths.  However, super-net DNAS methods are often limited in their capacity to model complex constraints, and their utility can be outweighed by evolutionary/learning-based NAS methods if the base architecture results in an excessively large super-net. ProxylessNAS \cite{proxylessnas} mitigates this issue by limiting subnetwork evaluation to binary paths. 

NAS excels in NN compression and optimization for a given task but is limited in its ability to jointly optimize performance and hardware-aware constraints. Typical EA-NAS approaches resort to using MACs, FLOPs, and other proxies for energy consumption. NASCaps \cite{nascaps}, for example, supports energy-aware NAS for traditional DNNs alongside convolutional capsule networks \cite{capsnet}, proxying energy consumption using a combination of MACs and memory-accesses. By integrating capsule layers and dynamic routing, it jointly optimizes capsule networks for energy consumption, accuracy, memory-footprint, and latency.

However, as previously mentioned, such proxies often result in an incomplete profile of DNN energy consumption. The difficulty in using energy as an optimization metric again lies in accurately estimating the consumption of a given architecture without physically executing it, which is infeasible given the vast design space. Additionally, the energy consumption of a DNN is highly device/platform dependent. Therefore, EA-NAS methods often utilize regression-based approaches, built on recorded energy measurements from a subset of NAS-generated reference architectures on the target platform, to \textit{predict} the energy consumption of candidate architectures. Liu et. al~\cite{split1} introduce NAS as a "continual splitting process", in which DNN operators are split for iterative loss minimization, and Wang et. al \cite{split2} extend this by incorporating benchmark energy-measurements of operator splitting to discover more energy-efficient architectures. ET-NAS \cite{etnas} uses energy consumption benchmarks of NN operators on various edge-side FPGA platforms, and under various parameter settings. ChamNet \cite{chamnet} is an evolutionary NAS method utilizing Gaussian-process regression and energy estimates derived from benchmark measurements for a range of hardware platforms, including mobile CPUs and DSPs. \S\ref{section:iee} provides an overview of existing energy-prediction methods for NAS, and their supported hardware platforms and DNN operators. 

Tabular energy consumption benchmarks for various architectures are slowly developing. However, they currently remain limited to common NN operators, architectures, and datasets. HW-NAS-Bench \cite{hw_nas_bench} collects measured/estimated hardware metrics, including energy consumption, of explored candidate networks within the spaces of NAS-Bench-201 \cite{nas_bench_201} and FBNet \cite{fbnet}, on six diverse COTS devices. These include the NVIDIA Jetson TX2 GPU, Raspberry Pi 4, an Edge TPU Dev Board, Google Pixel 3, and Eyeriss (ASIC) \cite{eyeriss}. Ta et. al \cite{deepen} conduct hardware energy measurements of various edge-DNNs, including but not limited to MobileNet v1/v2 \cite{mobilenetv1, mobilenetv2}, SqueezeNet \cite{squeezenet}, GoogLeNet \cite{googlenet} and ProxylessNAS \cite{proxylessnas}, on various other mobile CPUs and GPUs. Devices used include the OnePlus 8, Xiaomi Redmi Note8, and the Huawei Mate40 Pro and P40 Pro/Lite. 

Generalizing to new architectures, without complicating the NAS space with hardware specifications, involves one- or few-shot learning methods \cite{fs1, fs2}. The extension of existing benchmarks via few-shot networks and surrogate models, predicting learning dynamics in conjunction with energy measurements, can help bridge the gap between hardware-agnostic NAS optimization and practical energy-efficient architecture search \cite{ec_nas}.

Table \ref{tab:1} provides an overview of EA-NAS methods, including the architectural space and hardware-platforms on which they were evaluated, as well as their energy-estimation methodologies.

Future work in EA-NAS should involve the development of energy-estimation methods that can accurately predict DNN energy consumption based solely on architectural features ($i.e.$, without requiring execution). The development of hardware-agnostic techniques capable of optimizing DNNs for various hardware platforms and profiles with minimal re-training is also essential, as well as efficient hardware-profiling methods to enable the scalable deployment of HW-NAS. This introduces a cross-domain learning problem, involving the prediction of absolute DNN energy consumption across hardware platforms; it prompts the development of a domain-invariant means of profiling DNNs, such as ASTs \cite{cdmpp}.

\subsection{Predicting Inference Energy}
\label{section:iee}
The accurate hardware-agnostic prediction of DNN energy consumption is key in guiding the design of energy-efficient DNNs. There exist numerous works in this area, but no universally applicable method. The bulk of methods employ regression-based approaches, trained on hardware-specific benchmark DNN or layer-wise energy measurements, and utilize features such as MACs, FLOPs, and layer configurations as input. However, these approaches are constrained by the DNN operations they support, and their performance is currently extremely platform-dependent. Future work should prioritize the refinement of cross-platform estimation techniques, and the optimization of energy measurement pipelines \cite{mlonmcu}. Table \ref{tab:2} provides an overview of the various existing methods, including their hardware evaluation platforms and supported DNN operations. Tu et. al's \cite{deepen} prediction method is currently SOTA in terms of generalization performance on unseen networks and edge devices, using a RF regression model built on kernel-level measurements on various mobile Cortex A-series CPUs and Mali G-series GPUs. SyNergy \cite{synergy} and DeepWear \cite{deepwear} are the only other methods built for constrained/edge devices, evaluated on the Nvidia Jetson TX1 and Cortex A-series CPUs, as well as the Mali-400 mobile GPU. 

\begin{table*}
  \centering
  \setlength\extrarowheight{-5pt}
  \caption{\textmd{An overview of DNN energy-estimation methodologies. A dash indicates an out-of-scope area.}}
  \renewcommand{\arraystretch}{0}
   \addtolength{\tabcolsep}{-3pt}
  \resizebox{13cm}{!}{\begin{tabular}{c|rrrrr}
    \Xhline{1pt}
  \thead{\textbf{Method}} & \thead{\textbf{Granularity}} &	\thead{\textbf{Inputs}} & \textbf{Application} & \thead{\textbf{Eval. Platform}} & \thead{\textbf{Supported Operators}} \\
    \Xhline{1pt}
    \hline
    \textbf{\cite{synergy}}	& \makecell[r]{Layer, \\ Network} & MACs & Inference & CPU (Cortex A-series) & \makecell[r]{\texttt{conv}, \texttt{fc}} \\
    \hline
    \textbf{\cite{delight}} & Network & Layer parameters & \makecell[r]{Inference, \\ Training} & CPU-GPU (Nvidia Tegra K1) & \makecell[c]{-} \\
    \hline
    \textbf{\cite{deepwear}} & Layer & Layer parameters & \makecell[r]{Inference, \\ Training} 
 & \makecell[r]{CPU (Snapdragon 400 \& 805),\\ GPU (Mali-400)}	& \makecell[r]{\texttt{conv}, \texttt{fc}, \texttt{relu}} \\
 \hline
 \textbf{\cite{eyeriss}}	& Layer & \makecell[r]{MACs, \\ memory accesses} &
 Inference & ASIC (Eyeriss) \cite{eyeriss} & \makecell[r]{\texttt{conv}, \texttt{fc}} \\
 \hline
 \textbf{\cite{deepen}} & \makecell[r]{Layer, \\ Network} & Layer parameters & Inference & \makecell[r]{CPU (Cortex A-series), \\ GPU (Mali-G series \& Adreno 610)} & \makecell[r]{\texttt{conv}, \texttt{fc}, \texttt{relu}, \texttt{bn}} \\
    \hline
  \end{tabular}}
  \label{tab:2}
  \vspace{-0.3cm}
\end{table*}


\section{Energy-Adaptive Inference}
\label{section:adaptive}
Realizing EA inference entails an inference-time trade-off between loss/accuracy and energy consumption. Numerous approaches optimize DNN inference by utilizing lightweight architectures and static optimizations. However, these static optimizations, covered in \S\ref{section:static}, lack inference-time dynamicity. Hence, there’s a need for adaptive inference techniques that can adjust energy consumption dynamically based on variations in energy-availability, workload, and input complexity. The latter is motivated by the recognition that if inputs are not all equally \textit{hard}, a non-adaptive network incurs unnecessary processing for \textit{easy} inputs or suffers performance loss for difficult ones \cite{big_little, tann, pagliari1, pagliari2, panda}. Here, we discuss techniques for enabling EA inference, broadly categorized into DNN right-sizing, multi-exit, and inference offloading.

\subsection{DNN Right-Sizing}

Lighter networks can often handle the bulk of input instances nearly as well as the optimally-trained full network. Thus, cascaded networks of increasing complexity can be sequentially executed for efficient resource utilization with negligible accuracy loss. By overlapping the parameters of cascaded DNNs, recent works have harnessed the versatility inherent in employing multiple networks of increasing capacity with the memory-footprint of a single network; such parameter-sharing approaches are collectively known as DNN \textit{right-sizing}.

Notable right-sizing approaches include input-dependent layer-wise pruning \cite{tann, lin}, DNN modularization and inference-time module selection \cite{odena}, and input-dependent quantization \cite{pagliari1}. The latter re-configures DNN precision based on input difficulty, unlike traditional static quantization which sets the bit-width uniformly across the entire network or different network subsets. This prevents accuracy loss for difficult inputs that cannot be handled at low precision whilst reducing energy consumption for the bulk of inputs that can be accurately processed with a less conservative quantization. Additionally, input-dependent quantization does not require entire network re-training in order to maintain the base non-quantized accuracy at varying precision levels. It can be universally applied to any pre-trained CNN. Feature boosting and suppression \cite{fbs}, which predictively ``boosts” layer channels at runtime and suppresses those deemed unimportant, is another universally applicable channel-based approach. DeLight \cite{delight} employs input-adaptive pre-processing by projecting input data to an ensemble of lower-dimensional subspaces based on their context and energy resource-constraints.

HydraNets \cite{hydranets} consist of variable execution branches specialized towards extracting features of different input classes. They utilize a gating mechanism to choose the appropriate execution branches for a given input and ensemble techniques to aggregate branch outputs to produce a final inference. The various network branches are trained jointly for better accuracy. This joint training is enabled by independent batch-normalization (BN) of feature means and variances, which prevents feature aggregation inconsistency across the different branches/widths \cite{slimmable}. Given that BN layers typically constitute less than 1\% of network parameters, this incurs negligible extra computational/memory overhead. Daghero et. al~\cite{har} extends HydraNets for energy-adaptive inference by decomposing a network into branches of varying complexity, assigning ``difficulty levels" to inputs, and processing them with the corresponding combination of network branches. 

HydraNets, alongside the other works detailed in this section, focus on CNNs. There exist non-energy-focused works on RNNs, utilizing dynamic entropy-based beam-width\footnote{BW refers to the number of candidate sequences evaluated at each step.} (BW) modification for adaptive inference \cite{bw1, bw2}. Recent works have also demonstrated adaptive inference on transformer-based DNNs, including cascaded transformers with increasing numbers of patches \cite{patches}, and input-based patch, block, and head selection \cite{pbh}; however, these are not generally applicable to constrained devices, and none directly considers energy as a metric.


\subsection{Multi-Exit Networks}
\label{section:ee}

Proliferation in DNN complexity is helpful only when classifying difficult inputs, which are inherently rare. The features learned at earlier network layers are often sufficient for accurately classifying the vast majority of the data population \cite{figurnov, branchy, skipnet}. Multi-exit networks offer a form of adaptive right-sizing where inputs can exit the network at intermediate layers, bypassing the unnecessary processing of subsequent layers. Such networks obviate the need for re-execution in the case of a failed inference and can be combined with fixed optimizations.

\begin{figure*}[h!]
\begin{center}
\includegraphics[width=7cm]{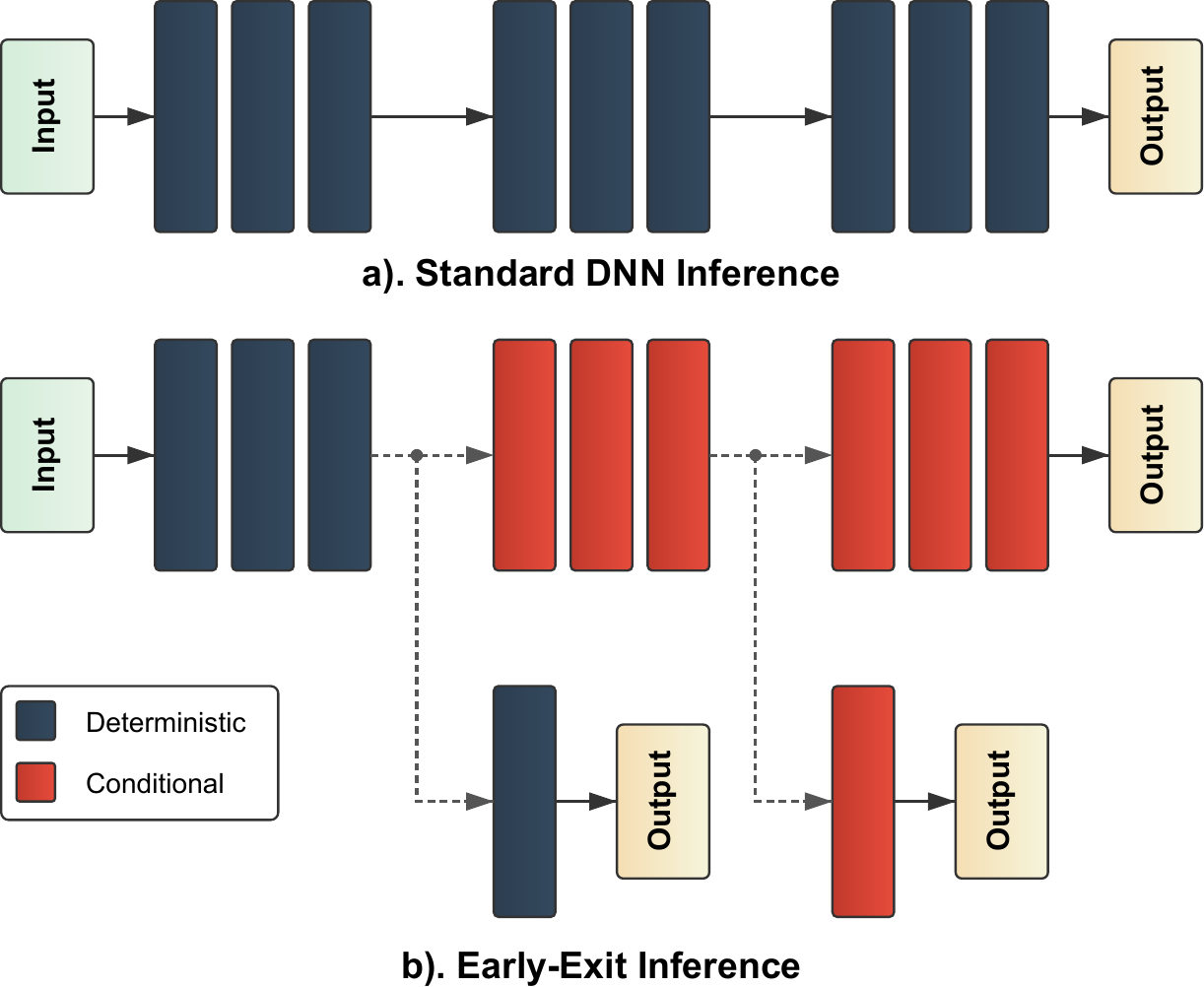}
\caption{\textmd{\textit{Conventional} DNN architecture alongside an EE variant. The dashed lines indicate conditional execution paths.}}\label{fig:2}
\vspace{-0.1cm}
\end{center}
\end{figure*}

The design of a multi-exit network includes the addition of side-branch exit heads to 
a \textit{backbone} network; Fig. \ref{fig:2} below outlines a conventional DNN alongside a multi-exit variant of the same architecture. With trivial modifications, any DNN can be made an implicit ensemble of networks by adding exit layers. The exit-point(s) of a multi-exit network can be determined either pre-execution (subnet-based inference \cite{branchy}) or at inference time (progressive inference \cite{konstantin}). The former allows for more reliable estimation of inference energy requirements, enabling dynamic voltage and frequency scaling (DVFS). However, inference always continues until the selected exit layer, potentially resulting in wasted energy for \textit{easy} inputs and/or overthinking \cite{kaya}. Instead, with progressive inference, the exit-point is determined during network execution, typically based on the uncertainty of the exit classification or following an energy-management policy.  This can reduce energy consumption via dynamic adaptation to input complexity, but necessitates more intricate hardware-level adjustments to do so.

\subsubsection{Multi-Exit Design}

Introducing exit layers to pre-trained backbone DNNs provides versatility when choosing the backbone design, as well as the case-driven optimization of exit layers \cite{cdl}. However, more recent works have explored generalised structures \cite{ee1, ee2, kaya, ee4, ee5, ee6} and the co-design of end-to-end multi-exit networks \cite{ee7}. By co-designing and jointly training the backbone NN and exit layers, there comes greater flexibility in optimization and the opportunity for improved performance at the cost of restricted task adaptability.

Another design decision is the granularity and positioning of exit layers. Coarse-grained exits reduce overheads, but can miss exit opportunities. Conversely, fine-grained exits increase overheads, leading to heightened energy requirements and increased parameters. The extent of these overheads depends on the depth of the exit-layer(s), the network architecture, the use-case, and the exit policy. Zero Time Waste (ZTW) \cite{ztw} addresses the overheads associated with failed exits, at the expense of (marginally) increased energy consumption and memory-footprint, by adding direct connections between exit-layers and recycling their outputs in an ensemble-like manner. Additionally, fine-grained exits can negatively impact convergence when jointly training the backbone NN and exit layers. PredictiveExit \cite{pred_exit} uses a low-cost prediction engine to forecast where an input instance will exit the network, reducing the energy overheads of fine-grained exits by skipping the execution of pre-placed exit layers. This enables runtime configuration adjustment based on DVFS for energy-saving during inference. 

NAS approaches have recently been extended to the co-optimized design of multi-exit DNNs to meet energy, computation, and/or memory constraints. HarvNAS \cite{harvnet} configures multi-exit network architectures for battery-less devices by jointly optimizing the backbone network, exit layers, and inference policy subject to energy and memory constraints. However, the addition of exit layers increases the already vast design space with placement configurations. \S\ref{section:design} covers EA architecture design and NAS methods in more detail.

Recent works have explored multi-exits for compute-intensive Transformer architectures \cite{transformer1, transformer2, transformer3}, but largely overlook energy consumption as a metric, whilst other architectures are mostly unexplored.

\begin{figure*}
\centering
\includegraphics[width=10.5cm]{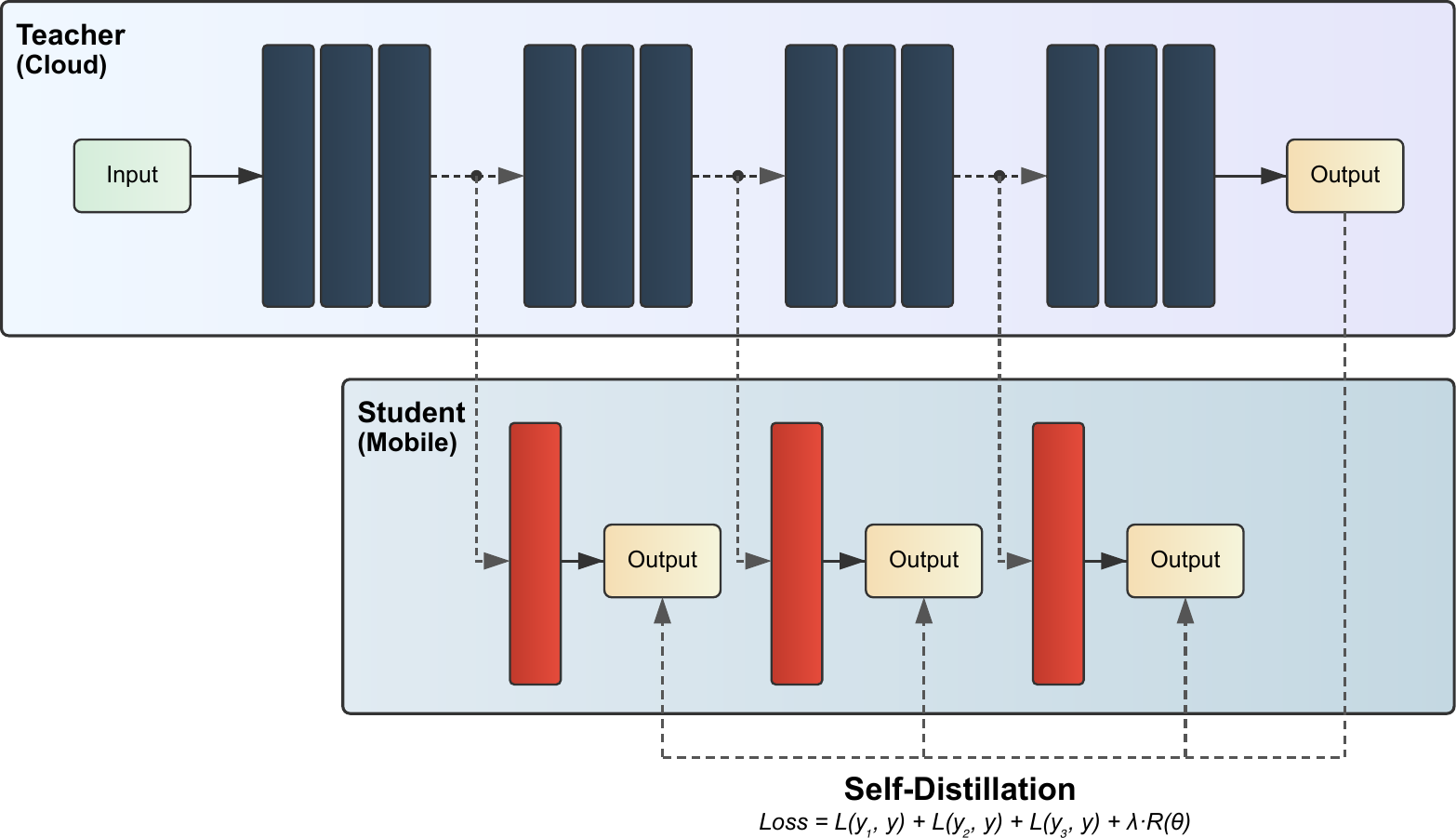}
  \caption{\textmd{Multi-exit last-layer distillation training, with a backbone network trained in the cloud and its exits trained on-device.}} \label{fig:3}
  \vspace{-0.5cm}
\end{figure*}
\subsubsection{Deployment of Multi-Exit Networks}
\label{section:harvsched}
Initial multi-exit works such as CDL \cite{cdl} utilize exit policies based on static thresholds, quantifying the network’s inference confidence using metrics like top-1 prediction score \cite{konstantin}, prediction entropy \cite{branchy}, and score margin (SM), the margin between the $1^{st}$ and $2^{nd}$ inference output \textit{scores} \cite{big_little, cdl}. However, imposing a global threshold on network outputs corresponds to implicitly assuming all classes are equally difficult to process. In scenarios for which this assumption does not hold, class-dependent thresholds become necessary \cite{ee6} (\S\ref{section:inf}). 

Rule-based policies enable the consideration of more nuanced energy-aware objectives, such as maximizing throughput given an energy budget \cite{jiang}. Zygarde \cite{zygarde} is a scheduler for multi-exit inference tasks on energy-harvesting devices, constrained by the need for immediate task execution alongside varying energy-availability. Reinforcement learning (RL) has also been used to generate policies that additionally consider historical device operation and the longer-term effects of scheduling decisions. HarvSched, a RL–based exit policy scheduler for energy-harvesting devices, learns its policy considering both instantaneous factors ($i.e.$, harvesting status and energy-storage level) alongside device operational history \cite{harvnet}. HarvSched demonstrates superior accuracy plus \textit{fewer missed events} than Zygarde when evaluated in a test scenario (randomly-triggered events on an Arduino Uno equipped with MSP430-based solar harvester, 0.47F capacity supercapacitor, and TI PMIC).

The on-device training of exit-layers requires efficient fine-tuning approaches, covered in  \S\ref{section:perz}. 

\subsection{Energy-Adaptive Inference Policies}
\label{section:inf}

Effective DNN right-sizing requires an adaptive inference policy that decides on the optimal network configuration for each inference. Such \textit{binary} policies are inherently non-differentiable. Consequently, most utilize RL, typically using either REINFORCE \cite{reinforce} or Q-learning \cite{qlearning}; the latter is particularly well-suited to resource-constrained devices, as inference involves a table-based look-up. However, RL necessitates pre-deployment, or cumbersome trial-and-error learning when pre-deployment isn't feasible. Recent works are beginning to introduce non-RL approaches; these include approximating the gradients of non-differentiable policy functions with continuous ones \cite{hua}, employing gated mixture-of-experts approaches \cite{shazeer} (which utilize noisy-ranking on the gating networks), or training policy networks \cite{blockdrop}. 

These policies also necessitate a means of evaluating inference uncertainty. The works detailed above all use SM as proxy \cite{big_little}, with the rationale being that the margin between the $1^{st}$ and $2^{nd}$ inference output \textit{scores}, in a classification task, is proportional to the likelihood of correct top-1 inference. However, imposing a global threshold on network outputs corresponds to implicitly assuming all inputs are equally difficult to process. In scenarios where this assumption does not hold, class-dependent thresholds become necessary; Daghero et. al \cite{ee6} defines these as follows:
\begin{equation} 
        th_c = \text{argmin}_{th_c}(FP_c(th_c)+\alpha \cdot E_c(th_c)) 
\end{equation}
where $th_c$ denotes the threshold for class $c$ and $FP_c(th_c)$ the false positive rate. $E_c(th_c)$ is a measure proportional to the energy consumption associated with processing an input of class $C$, quantified by the weighted number of invocations of the various cascade DNNs. Additionally, they propose a method for updating these class-dependent thresholds at runtime. 

Beyond multi-exit networks, energy-based thresholding is not largely applied. This is where DNN energy consumption is profiled across various network configurations ($i.e.$, branches, modules, bit-widths), and the network is adapted accordingly based on current/forecasted energy-availability. This is particularly relevant for energy-harvesting devices, where energy-availability fluctuates. \S\ref{section:intermittent} covers general DL techniques for intermittent and energy-harvesting devices in detail.

\subsection{Inference Offloading}

 
While data transmission is energy-expensive, device- and cloud-only approaches are typically not optimal in terms of inference energy consumption and latency \cite{jointdnn}. Given sufficient energy-availability and task deadline, and a relatively stable connection, \textit{offloading} inference from device to a cloud server may be advantageous, as a larger-capacity network can handle full or partial inference.

\begin{figure}[!htbp]
\begin{center}
\includegraphics[width=6cm]{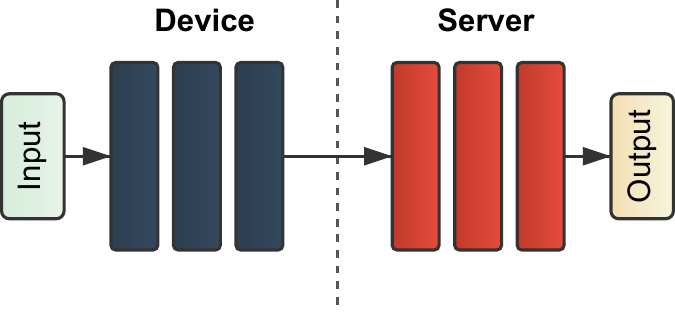}
\vspace{-0.3cm}
\caption{\textmd{DNN partitioning between device and server.}}\label{fig:4}
\vspace{-0.3cm}
\end{center}
\end{figure}

Typical inference offloading methods involve layer-wise DNN partitioning. Fig. \ref{fig:4} provides an illustration of this. The optimal partitioning of a DNN is dependent on its layer topology, the inference task, the connection bandwidth, and the energy profile of the target device. Neurosurgeon \cite{neurosurgeon} profiles the edge-device and cloud-server to generate energy/latency-cost estimations for the execution of different DNN layer types. Using these estimations, and other relevant device factors ($e.g.$, the current bandwidth), Neurosurgeon partitions DNN computation between device and cloud-server to optimize for energy consumption and end-to-end latency. EEoC \cite{eeoc} computes the partitioning using real measurements, instead of estimations, of the costs of DNN layer execution and layer output transmission on the target device. JointDNN \cite{jointdnn} uses ILP to find the optimal partitioning layer, with respect to energy-availability, bandwidth saturation/congestion, and quality-of-service. It also evaluates the benefit of compressing DNN layer outputs to reduce transmission energy costs. Boomerang \cite{boomerang} minimizes the overall energy cost of inference by evaluating the trade-off between on-device latency and energy cost versus the transmission energy cost for partitioning multi-exit networks.

Alternatively, depth-wise partitioning methods vertically partition (convolutional) layers to reduce memory footprint whilst enhancing parallelism in communication and task offloading \cite{deepthings}. However, vertical tile-based partitioning, unlike depth-wise layer partitioning methods, can result in increased communication overheads and dependencies as it requires compiling results from adjacent partitions.

Khoshsirat et al. \cite{divide_and_save} explore dynamically partitioning inference between containers on the same device to reduce energy consumption, and provide insights on how to optimally allocate inference tasks into these containers from an energy standpoint. Their approach is evaluated on Nvidia Jetson TX2 and AGX Orin edge devices and demonstrates energy-savings. However, it is only applicable if inference can be partitioned into independent tasks; a future direction should be to investigate container-wise partitioning of dependent tasks. Additionally, the effectiveness of such partitioning diminishes as devices become increasingly resource-constrained. 

Inference offloading is particularly suitable for wearable devices ($e.g.$, smart watches) as they are typically paired with a mobile device. The communication between the wearable and offloading devices can be efficiently realized by short-range radio ($e.g.$, BLE). DeepWear \cite{deepwear} enables the context-aware partial offloading of CNN and RNN inference from wearables to their paired mobile devices via local BLE connectivity. This involves minimizing the weighted energy consumption of all feasible DNN partitions. DeepWear has been successfully applied to COTS wearables, including the LG Urbane smartwatch, and mobile devices such as the Google Nexus 6.

Additionally, offloading has been extended to battery-less energy-harvesting devices; Sabovic et. al \cite{sabovic} demonstrate its use for the application of person detection on an Arduino Nano (with 1.5F supercapacitor, AEM10941 solar harvester and PMIC, and BLE for wireless communication). Their device estimates the latencies of device-only and cloud inference (no support for DNN partitioning) and decides to offload based on the task-deadline and current harvesting status. Future work could explore learning-based offloading approaches that consider historical device operation and energy-availability.

\vspace{-0.3cm}
\section{On-Device Training}
\label{section:perz}

The above sections generally focus on optimizing on-device inference with an offline-trained network. However once deployed, data- or context-shifts can result in degraded network performance in the target environment \cite{shannon_strikes_again}. This issue is especially relevant for \textit{in-the-wild} sensing applications, where environmental conditions can be highly variable and result in distortion ($e.g.$, fog, snow, rain) and/or degradation ($e.g.$, defocus). Additionally, a deployed classifier may need to include new classes or reduce its existing class set to optimize its performance in the target environment. While not \textit{energy-aware}, methods that contribute to enhancing the out-of-distribution (OOD) generalizability of pre-trained networks are key for IoT deployments. These include causality- \cite{causal1, causal2, causal3}, gradient operation- \cite{gradient1, gradient2, gradient3}, and learning-based methods ($e.g.$ zero-shot, meta, and representation learning) \cite{learning1, learning2, learning3, learning4, learning5}. However, explicit fine-tuning commonly outperforms generalization in mitigating data/context shifts in real-world evaluations \cite{kumar, andreassen, wiles, innout, salman}.

Typically, DNNs are fine-tuned in the cloud, as gradient backpropagation is energy-expensive and prohibitively memory-intensive on resource-constrained devices (due to limited SRAM). 
However, this massively increases the communication overheads of the IoT device, increasing bandwidth requirements, and is not viable for ultra-low-power devices or those deployed in remote environments. Hence, there is substantial recent work towards optimizing DNN fine-tuning requirements, using unsupervised and self-supervised methods, to meet the energy budgets of constrained devices. 

MiniLearn \cite{minilearn} enables IoT devices to fine-tune and optimize pre-trained integer quantized networks using intermediate compressed outputs of quantized layers. It fine-tunes dequantized hidden (convolutional and FC) layers while keeping the rest of the pre-trained network in its quantized integer precision, reducing energy-usage without compromising performance. However, MiniLearn operates in the constrained context where network performance is optimized by reducing its class set and relies on labelled target data. LifeLearner \cite{lifelearner} is a rehearsal-based meta continual-learning (CL) approach that is optimized for data-scarcity and energy-efficiency to enable realistic on-device CL (demonstrated on a STM32H747 MCU). Typical CL approaches require full-network backpropagation, and sufficient representative data to be kept on-device for meta-testing. These requirements are not viable in a constrained on-device scenario. Similar to MiniLearn, LifeLearner introduces a new rehearsal strategy, in which rehearsals are extracted as compressed outputs of intermediate network layers. By optimizing the intermediate rehearsal layer for maximum compression, LifeLearner reduces the computational- and energy-costs of fine-tuning. Additionally, LifeLearner utilizes bitmap compression and product-quantization of rehearsal-layer outputs to reduce memory-footprint.  

Decreasing the number of fine-tuning parameters reduces its energy-expense \cite{autofreeze, flextuning, clint} and can improve performance by reducing overfitting \cite{kirkpatrick, elsa, spottune, catastrophy}. Hence, recent works have also explored quantifying the contributions of network layers/parameters towards counteracting data shifts. NEq \cite{neurons_at_equilibrium} reduces fine-tuning energy consumption by iteratively determining the weights to fine-tune. It computes the cosine-similarity between weights at epochs \textit{t} and \textit{t-1}, and uses this difference to quantify their ``relative velocity". Then, NEq selects to fine-tune weights not in equilibrium, where equilibrium can be defined as:
\begin{equation}
|v_i^t| < \epsilon, \; \; \; \; \; \; \epsilon \le 0
\end{equation}
Here, $v_i^t$ is the relative velocity of parameter $i$ at epoch $t$ and $\beta$ is an arbitrary threshold. Quélennec et. al ~\cite{neurons_under_budget} replaces the NEq threshold with a fine-tuning energy budget $B^w$, and greedily selects the $n$ highest velocity weights that can fit within the defined budget. However, efficiently determining which layers/parameters to fine-tune is an ongoing challenge \cite{surgical}. NEq, and other gradient-based selection methods \cite{surgical}, require a full-network backpropagation to compute the initial parameter velocities (which mightn't be viable on extremely energy- or memory-constrained devices), alongside fine-grained estimates of parameter-wise energy consumption. The NEq method applied layer-wise could support better estimates, and the layer-wise velocity can be initially approximated by random-sampling of intra-layer neurons.

Recent works utilize multi-exit networks to support efficient personalization by fine-tuning exit layers whilst maintaining the network backbone \cite{distill4}. Typically, this is done using a self-supervised knowledge distillation approach (see Fig \ref{fig:3}) in which labels are generated by the output of the last exit. However, Fawden et. al \cite{fawden} analyze per-exit performance, demonstrating that it is not valid to assume the last exit is always the most accurate. They instead recommend selecting input instances on which to fine-tune based on their predictive uncertainty. Additionally, they interpret exits as an ensemble of networks with a joint loss. 

\section{Applications}
\label{section:apps}
\subsection{DL on Energy-Harvesting Devices}
\label{section:intermittent}
The utilization of DL on battery-less devices, which harvest energy from environmental sources ($e.g.$ solar, vibration, RF), has garnered recent interest. These devices offer numerous advantages over their battery-powered counterparts. They are largely maintenance-free and easy-to-recycle, promoting sustainability and suitability for large-scale deployments in hard-to-reach environments. Additionally, capacitors are more resistant to degradation than batteries, prolonging the lifetime of these devices to potentially decades. Given that data transmission is more energy-expensive than performing local inference, and that battery-less devices operate on extremely low energy-budgets, their deployment is impractical without on-device inference and task-scheduling. However, battery-less devices operate intermittently as energy is sporadically available, with recurring power failures that result in on-off behaviour. This presents challenges in efficiently guaranteeing inference \textit{correctness} in the event of power failure, alongside reducing its likelihood via task-scheduling based on fluctuating energy-availability.

Despite being a relatively new area of interest, there are numerous works demonstrating the real-world feasibility of DL on energy-harvesting devices. These include CNN-based facial recognition on solar-powered devices \cite{face_rec1, face_rec2}, a capacitance-based gesture recognition wearable \cite{capband}, and DL-driven bioacoustics monitoring in underwater environments using harvested acoustic energy \cite{underwater}. 

\subsubsection{Intermittence-Safe Inference}

Battery-less devices accumulate charge using harvested energy until voltage $\texttt{V}{\texttt{on}}$ is reached; the device then powers on and operates until charge depletes to $\texttt{V}{\texttt{off}}$. This cycle repeats periodically or aperiodically. $\texttt{V}{\texttt{on}}$ is typically set pre-deployment. Future work could explore adjusting capacitor capacity at runtime using a policy-based or RL approach to match energy fluctuations and workload.

During intermittent execution, power failures reset volatile memory, hampering computational progress, and can lead to non-volatile memory inconsistencies ($e.g.$, partial variable updates can occur because of an interrupted partial execution and lead to incorrect results in the subsequent re-execution). Numerous methods exist for guaranteeing correct execution on intermittent systems, ranging from using periodic checkpoints to store intermediate states \cite{int1, int2}, non-volatile processors that maintain state across power cycles \cite{int3, int4} and methods based on executing atomic tasks \cite{int5, int6}. However, these methods introduce computational (and therefore energy) overheads, especially those dealing with non-volatile processing \cite{alfred}. This is problematic for computation-heavy DNN inference/training. SONIC \cite{sonic} is an intermittence-safe task-based API designed explicitly for DNN inference, aiming to exploit its regularity to minimize overheads. SONIC "breaks the rules" of conventional task-based programming paradigms by allowing the persistence of non-volatile memory across loop iterations. However, a combination of loop-ordered buffering and sparse-undo logging ensures that loop iterations are idempotent, guaranteeing safe resumption of interrupted tasks. Additionally, loop continuation eliminates wasted computation in the event of power failure. 
SONIC also presents the first demonstration of DNN inference on a constrained energy-harvesting system, using the COTS MSP430 MCU with a RF-energy harvester.

\subsubsection{Inference Scheduling}
\label{section:harvsched2}
While SONIC and other intermittence-safe frameworks guarantee correct inference in the event of a power failure, they do not aid in reducing downtime. DL on energy-harvesting platforms also requires energy-reactive inference strategies and inference task-scheduling, given cyclical/fluctuating energy-availability. Existing inference strategies largely focus on multi-exit networks. ePerceptive, a notable energy-reactive inference framework, built on SONIC, outputs lower-fidelity inferences at intermediate layers in response to energy fluctuations whilst guaranteeing valid inferences in the case of power failure \cite{eperceptive}.  ePerceptive also dynamically adjusts input resolution(s) to optimize performance under energy constraints. 

General task-scheduling for computing on energy-harvesting devices is well-studied \cite{sched1, sched2, sched3, sched4, sched5}. ACES \cite{ACES}, for example, uses RL to optimise the duty cycles of battery-less nodes in order to maximise utility and increase node lifetime, under fluctuating energy-availability. DL inference schedulers can be generally categorized as rule- or RL-based. Zygarde\cite{zygarde}, as discussed above, is an online rule-based scheduler for multi-exit networks that schedules exit-decisions based on an objective considering task-deadline, inference uncertainty and the current energy-harvesting status of the device. Zygarde partitions inference into mandatory and optional layers, and uses k-means to estimate inference uncertainty. The limitation of Zygarde and other non-trainable schedulers is that they fail to consider historical device operation. HarvSched, also discussed above (\S\ref{section:harvsched}), is a RL–based scheduler that learns its scheduling policy based on current factors ($e.g.$, energy-storage level) and device operational history \cite{harvnet}. This facilitates adaptive scheduling decisions based on evolving conditions. However, RL-based schedulers need pre-deployment in order to learn their policies, or cumbersome trial-and-error learning when pre-deployment isn't feasible.


The methods discussed above enable on-device inference using an offline-trained network. However, on-device training/fine-tuning is also necessary for counteracting real-world data- and context-shifts. Given that backpropagation is computationally more expensive than inference, fine-tuning only on non-redundant inputs is important; blindly selecting inputs on which to fine-tune can result in wasted energy if the inputs do not contribute towards improving DNN performance. 
Lee et. al \cite{intermittent_learning} assess input relevance with a combination of three metrics, based on input uncertainty, diversity, and representation, to demonstrate unsupervised/self-supervised learning on intermittently-powered COTS MCUs (AVR, PIC, MSP-430). \S\ref{section:perz} covers on-device training and fine-tuning methods in more detail.

\subsection{Energy-Aware Federated Learning} 
\label{section:fl}
Federated Learning (FL) is a decentralized ML approach wherein DNNs are trained collaboratively across devices whilst keeping their raw data localized. This is realized via communicating weight updates to base servers that coordinate to aggregate a global network. FL offers scalability in massively-distributed environments, which can span over a large number of devices. FedAvg, the most common FL approach, performs rounds of local training on device subsets to accommodate their heterogeneous resource constraints and latencies. FedAvg has been demonstrated to guarantee convergence in various heterogeneous scenarios, including unbalanced and non-i.i.d distributions \cite{fedavg1, fedavg2, fedavg3}. Luo et. al~\cite{cefld} demonstrates the effectiveness of FedAvg on mobile devices.

The key to effective FL across heterogeneous resource-constrained devices is energy-aware scheduling. This involves a trade-off between computation- and communication energy-costs, which are dependent on local DNN efficiency, the number of participating devices, and the number of local iterations in each FL round. Most works on energy-aware FL focus on optimizing the energy costs of on-device network training and are largely based on static DNN optimizations. Q-FedUpdate \cite{qfedupdate} supports training with full integer quantization on energy-efficient mobile Digital Signal Processors (DSPs). These are found universally on mobile phones and most other IoT devices. Their use for DNN execution can yield 11.3x/4.0x reduction in energy usage over mobile CPU/GPUs. Training is supported by maintaining a global full-precision ($i.e.$ FP32) network, which continuously accumlates network updates instead of having updates erased by quantization. Q-FedUpdate is integrated with efficient batch quantization as well as a pipelining-based approach to enable simultaneous CPU-based quantization and DSP training. The latter reduces computational overheads associated with repeated network quantization. Other works focus on reducing FL communication-energy costs; approaches include gradient quantization via SignSGD \cite{signsgd}, gradient sparsification \cite{sparse1, sparse2, sparse3}, and dynamic batch-sizing \cite{dbs1, dbs2, dbs3}, as well as dynamically adjusting the number of local iterations between two global aggregation rounds \cite{agg}. In addition, energy-aware device scheduling can reduce communication costs. Sun et. al~\cite{denizzz}, for example, maximize the number of devices scheduled for gradient update at each iteration under an energy-constraint. Zeng et. al~\cite{radio} use energy-efficient bandwidth allocation and scheduling to reduce the energy consumption of FL over wireless networks.


Numerous recent works on energy-aware FL also concentrate on DNN partitioning to enable adaptivity towards heterogenous device energy budgets. Yang et. al~\cite{yang} utilize layer-wise partitioning, partitioning local DNNs into shallow and deep-layers, with shallow layers updated more frequently in FL rounds as they capture more general features. Rong et al.~\cite{rong} instead partition CNNs into subnetworks via width-wise partitioning, grouping filters into blocks and exchanging filter blocks among subnetworks periodically to ensure equal training across all devices while reducing training costs. Given the various participating devices, their energy-budgets, and their CNN partitions, a heuristic approach is used to optimize the allocation of subnetworks to devices. The devices are ranked by their communication latency, and subnetworks by their number of weights, and matched accordingly. A "resource configuration search" then determines the minimum communication latency fitting the energy-budget of each device, taking into account time constraints for FL round completion.

Other recent approaches model the FL trade-off as a joint optimization problem to minimize energy-cost while ensuring convergence, considering the various heterogenous devices and their energy budgets. Tran et. al~\cite{nguyen} formulate a joint-optimization problem based on the sum of computation and communication energy. However, their approach requires the synchronous upload of weights from all devices, which is unrealistic, and does not provide convergence guarantees. Luo et. al \cite{cefld} jointly optimize the number of participating devices and local iterations to minimize energy cost. They propose a sampling-based control solution that realizes minimal overhead and provides convergence guarantees. Yang et. al~\cite{zhaohui} derive energy consumption models of different FL approaches based on their convergence rates, utilizing these to form a novel joint-optimization problem.

FL on intermittent energy-harvesting devices poses unique challenges, as devices can only participate when they have sufficient energy available. Given the energy-generation of participating devices is non-uniform, device scheduling ($i.e.$, when to participate) becomes important. One approach is to let each device participate as soon as it can do so. However, Guler et. al \cite{susfl} demonstrate that this approach can bias the global FL network towards devices with more frequent energy-availability, resulting in an overall loss in performance. Another approach is to wait until all devices have sufficient energy to participate in the FL round, then use conventional FL sampling. This mitigates bias but requires waiting on the device with the slowest energy generation, which can result in slow convergence. Guler et. al introduce a FL protocol, with convergence guarantees, in which devices decide whether to participate in a given FL round via a stochastic process based on their energy profiles. This approach requires no communication between devices and is therefore scalable to large networks. However, it assumes knowledge of the energy-renewal cycles of the participating devices ($i.e.$, the number of FL rounds needed to generate enough energy to participate), necessitating pre-deployment (and cycles may vary over time).

Future work should involve the development of energy-aware DNN optimizations that can adapt to 
device heterogeneity to maximize energy-savings ($e.g.$, applying different quantization granularities based on the capabilities and energy-constraints of different participating devices \cite{5g}).

\section{Hardware Trends \& Gaps}
\label{section:hardware}
While the reviewed approaches span a variety of hardware platforms, most focus on accessible mobile and edge devices. Few approaches directly target MCUs, despite representing highly constrained platforms, likely due to the difficulty of implementing DL on such resource-limited architectures. The vast majority of approaches instead target mobile CPUs/GPUs, which is unsurprising given that these devices offer greater resource availability, making them more accessible for evaluation, yet still represent energy-constrained environments. Moreover, only a small number of approaches that do include evaluation on MCU hardware do so on non-hobbyist platforms \cite{harvnet, zygarde, ee6, sabovic, lifelearner}. 
Among the exceptions are those focusing on energy-harvesting, typically targeting extremely low-power off-grid systems. The STM32 and MSP430 families of MCU are most commonly evaluated. 

There is a particularly notable gap in energy-aware NAS approaches designed for or evaluated on heterogeneous MCU architectures. There are also limited evaluations on FPGAs/ASICs \cite{etnas, eyeriss}, but these platforms are generally associated with higher costs, more specialized use cases, and demand expertise for effective deployment. Offloading approaches generally transition tasks from mobile to edge/cloud environments, moving computations from a mobile CPU/GPU/DSP to more powerful devices like the Nvidia Jetson.

\begin{table}[ht]
    \centering
    \vspace{-0.1cm}
    \caption{Embedded DL frameworks and their supported platforms.}
    \vspace{-0.1cm}
    \label{table:frameworks}
    \resizebox{8cm}{!}{
    \begin{tabular}{@{}llc@{}}
        \toprule
        \textbf{Framework}       & \textbf{Supported Platforms}                    & \textbf{Open-Source} \\ \midrule
        TFLite-Micro \cite{tflitemicro}            & \makecell[l]{ARM Cortex-M series,\\ Xtensa (incl. ESP32),\\ RISC-V}                       & \cmark                               \\ 
        EdgeML \cite{edgeml}                 & \makecell[l]{ARM Cortex-M,\\ AVR RISC}                        & \cmark                              \\ 
        $\mu$Tensor \cite{utensor}                  & ARM Cortex-M                                  & \cmark                              \\ 
        $\mu$TVM \cite{utvm}                 & ARM Cortex-M                                  & \cmark                              \\ 
        CMSIS-NN \cite{cmsisnn}              & ARM Cortex-M                                  & \cmark                              \\ 
        STM32CubeMX \cite{stm32cubemx}           & ARM Cortex-M (STM32 series)                  & \xmark                             \\ 
        TAILS/SONIC \cite{sonic}            & TI MSP430                                     & \cmark                              \\ 
        \bottomrule
    \end{tabular}}
    \vspace{-0.1cm}
\end{table}

Broader platform and DNN-operation support, and improved compilation pipelines, of lightweight ML frameworks are essential for advancing MCU-based hardware evaluation; Table \ref{table:frameworks} summarizes current DL frameworks for MCUs and their supported platforms.

\section{Future Work and Directions}
\label{section:fw}
Having provided an outline of the evolving energy-aware ML \textit{landscape}, we identify a number of overarching directions for future work in this rapidly evolving field:
\vspace{-0.2cm}
\paragraph{Cross-Platform Energy Estimation}
Current EA approaches struggle to generalize across hardware platforms and network architectures, with their fundamental limitation being the accuracy of estimating the parameter/layer-wise energy consumption of a DNN, on a piece of hardware, without executing it. Similarly, finding universal correlations is incredibly challenging given the heterogeneity of hardware platforms, and the impact of their variations ($e.g.$, memory access hierarchy, loop nesting, and data flow) on DNN energy consumption. Future work should prioritize the development of universal, execution-free energy estimation approaches, leveraging architectural representations of DNNs to enable improved generalisation and cross-network/device invariance. For example, abstract syntax trees (ASTs), as used in CDMPP \cite{cdmpp}, offer a promising path for encoding the hierarchical layout of DNNs in a hardware-agnostic manner. Going forward, extending such representations to incorporate energy-relevant metadata could enable fine-grained, cross-platform energy optimizations.
\vspace{-0.2cm}
\paragraph{Automated Energy Profiling}
There remains a clear need for low-cost, hardware-agnostic methods that can generate reliable power and energy profiles at scale. Currently, the scalability and cross-device validity of energy prediction methods relies on profiling a large number of heterogeneous platforms and DNN operators; this necessitates community collaboration in building a \textit{holistic} energy benchmarking dataset. However, the accuracy of built-in power monitors for mobile/IoT profiling is insufficient, requiring the use of expensive external power monitors. As such, this also prompts the development of automated hardware-agnostic profiling methods for improving the efficiency of collecting device measurements at scale.
\vspace{-0.2cm}
\paragraph{Platform-Specific Optimizations}
In addition to generalization, energy-aware methods should also adapt to platform heterogeneity, accounting for both static hardware constraints and dynamic runtime conditions, such as energy availability or thermal limits, to maximize energy-savings. Future work should explore hierarchical, policy-based, or RL approaches for adapting DNN optimization granularity (e.g., adjusting quantization precision, modifying compute scheduling, or pruning subnetwork paths) based on task requirements, energy constraints, and runtime feedback. Such strategies become imperative when deploying a pre-trained network on a large number of heterogeneous devices ($e.g.$, in a FL setting).
\vspace{-0.2cm}
\paragraph{Energy-Aware Training Methodologies} While most attention has focused on inference-time energy optimization, energy-aware DNN training remains largely unexplored. Much opportunity exists for innovation; for instance, differentiable energy-aware regularizers could guide optimization toward architectures that are inherently more efficient across diverse hardware. Similarly, progressive training approaches that gradually increase NN complexity based on task difficulty could reduce unnecessary computation. Such work also includes developing $\mu$NPU architectures capable of supporting low-power, on-chip accelerated backpropagation.
\vspace{-0.2cm}
\paragraph{Multi-Modal Inference} Driven by hardware trends, such as increased on-chip memory bandwidth (e.g., SRAM or embedded DRAM) alongside tightly-coupled CPU/NPU architectures, there are growing opportunities for dynamic, multi-modal inference strategies. Sparse mixture-of-experts (MoE) architectures are one such approach, where compact networks are stored and dynamically switched in and out or combined based on input context or energy budget. This allows efficient handling of diverse modalities (e.g., vision, audio) while enabling context-sensitive inference without compromising energy efficiency. Future work should investigate how MoE strategies can be optimized for energy-adaptivity on resource-constrained platforms.
\vspace{-0.2cm}
\section{Conclusion}

This work has outlined key directions in energy-aware DL, highlighting the interplay between NN efficiency and hardware constraints. While enormous progress has been made in energy-efficient NN design and deployment, major challenges remain — particularly in accurate, hardware-agnostic energy estimation, adaptive optimization across diverse hardware, and scalable energy profiling with minimal overhead. Moving forward, addressing such challenges will require co-ordinated efforts spanning both hardware design and DL theory. 

We anticipate the continued emergence of specialized hardware accelerators, alongside developments in reduced data movement architectures, in-memory compute, and hardware-specific NN design, to push the boundaries of energy-efficient ML. We hope this survey acts as a useful foundation for further research at the intersection of DL and energy-constrained computing, guiding more informed and impactful designs across diverse and resource-constrained platforms.

\bibliographystyle{ACM-Reference-Format}
\bibliography{references}


\begin{thebibliography}{191}


\ifx \showCODEN    \undefined \def \showCODEN     #1{\unskip}     \fi
\ifx \showISBNx    \undefined \def \showISBNx     #1{\unskip}     \fi
\ifx \showISBNxiii \undefined \def \showISBNxiii  #1{\unskip}     \fi
\ifx \showISSN     \undefined \def \showISSN      #1{\unskip}     \fi
\ifx \showLCCN     \undefined \def \showLCCN      #1{\unskip}     \fi
\ifx \shownote     \undefined \def \shownote      #1{#1}          \fi
\ifx \showarticletitle \undefined \def \showarticletitle #1{#1}   \fi
\ifx \showURL      \undefined \def \showURL       {\relax}        \fi
\providecommand\bibfield[2]{#2}
\providecommand\bibinfo[2]{#2}
\providecommand\natexlab[1]{#1}
\providecommand\showeprint[2][]{arXiv:#2}

\bibitem[Abadade et~al\mbox{.}(2023)]%
        {tinymlreview}
\bibfield{author}{\bibinfo{person}{Youssef Abadade}, \bibinfo{person}{Anas Temouden}, \bibinfo{person}{Hatim Bamoumen}, \bibinfo{person}{Nabil Benamar}, \bibinfo{person}{Yousra Chtouki}, {and} \bibinfo{person}{Abdelhakim~Senhaji Hafid}.} \bibinfo{year}{2023}\natexlab{}.
\newblock \showarticletitle{{A Comprehensive Survey on TinyML}}.
\newblock \bibinfo{journal}{\emph{IEEE Access}}  \bibinfo{volume}{11} (\bibinfo{year}{2023}), \bibinfo{pages}{96892--96922}.
\newblock
\href{https://doi.org/10.1109/ACCESS.2023.3294111}{doi:\nolinkurl{10.1109/ACCESS.2023.3294111}}


\bibitem[Andreassen et~al\mbox{.}(2021)]%
        {andreassen}
\bibfield{author}{\bibinfo{person}{Anders Andreassen}, \bibinfo{person}{Yasaman Bahri}, \bibinfo{person}{Behnam Neyshabur}, {and} \bibinfo{person}{Rebecca Roelofs}.} \bibinfo{year}{2021}\natexlab{}.
\newblock \showarticletitle{{The Evolution of Out-of-Distribution Robustness Throughout Fine-Tuning}}.
\newblock \bibinfo{journal}{\emph{CoRR}}  \bibinfo{volume}{abs/2106.15831} (\bibinfo{year}{2021}).
\newblock
\showeprint[arXiv]{2106.15831}
\urldef\tempurl%
\url{https://arxiv.org/abs/2106.15831}
\showURL{%
\tempurl}


\bibitem[Apache(2024)]%
        {utvm}
\bibfield{author}{\bibinfo{person}{Apache}.} \bibinfo{year}{2024}\natexlab{}.
\newblock \bibinfo{title}{{MicroTVM}}.
\newblock \bibinfo{howpublished}{\url{https://tvm.apache.org/docs/topic/microtvm/index.html}}.
\newblock


\bibitem[ARM(2024)]%
        {cmsisnn}
\bibfield{author}{\bibinfo{person}{ARM}.} \bibinfo{year}{2024}\natexlab{}.
\newblock \bibinfo{title}{{CMSIS-NN}}.
\newblock \bibinfo{howpublished}{\url{https://github.com/ARM-software/CMSIS-NN}}.
\newblock


\bibitem[Bakhtiarifard et~al\mbox{.}(2024)]%
        {ec_nas}
\bibfield{author}{\bibinfo{person}{Pedram Bakhtiarifard}, \bibinfo{person}{Christian Igel}, {and} \bibinfo{person}{Raghavendra Selvan}.} \bibinfo{year}{2024}\natexlab{}.
\newblock \bibinfo{title}{{EC-NAS: Energy Consumption Aware Tabular Benchmarks for Neural Architecture Search}}.
\newblock
\showeprint[arxiv]{2210.06015}~[cs.LG]


\bibitem[Balaji et~al\mbox{.}(2018)]%
        {learning5}
\bibfield{author}{\bibinfo{person}{Yogesh Balaji}, \bibinfo{person}{Swami Sankaranarayanan}, {and} \bibinfo{person}{Rama Chellappa}.} \bibinfo{year}{2018}\natexlab{}.
\newblock \showarticletitle{{MetaReg: Towards Domain Generalization using Meta-Regularization}}. In \bibinfo{booktitle}{\emph{Advances in Neural Information Processing Systems}}, \bibfield{editor}{\bibinfo{person}{S.~Bengio}, \bibinfo{person}{H.~Wallach}, \bibinfo{person}{H.~Larochelle}, \bibinfo{person}{K.~Grauman}, \bibinfo{person}{N.~Cesa-Bianchi}, {and} \bibinfo{person}{R.~Garnett}} (Eds.), Vol.~\bibinfo{volume}{31}. \bibinfo{publisher}{Curran Associates, Inc.}
\newblock
\urldef\tempurl%
\url{https://proceedings.neurips.cc/paper_files/paper/2018/file/647bba344396e7c8170902bcf2e15551-Paper.pdf}
\showURL{%
\tempurl}


\bibitem[Berestizshevsky and Even(2018)]%
        {konstantin}
\bibfield{author}{\bibinfo{person}{Konstantin Berestizshevsky} {and} \bibinfo{person}{Guy Even}.} \bibinfo{year}{2018}\natexlab{}.
\newblock \showarticletitle{{Sacrificing Accuracy for Reduced Computation: Cascaded Inference Based on Softmax Confidence}}.
\newblock \bibinfo{journal}{\emph{CoRR}}  \bibinfo{volume}{abs/1805.10982} (\bibinfo{year}{2018}).
\newblock
\showeprint[arXiv]{1805.10982}
\urldef\tempurl%
\url{http://arxiv.org/abs/1805.10982}
\showURL{%
\tempurl}


\bibitem[Bragagnolo et~al\mbox{.}(2022)]%
        {neurons_at_equilibrium}
\bibfield{author}{\bibinfo{person}{Andrea Bragagnolo}, \bibinfo{person}{Enzo Tartaglione}, {and} \bibinfo{person}{Marco Grangetto}.} \bibinfo{year}{2022}\natexlab{}.
\newblock \bibinfo{title}{{To update or not to update? Neurons at equilibrium in deep models}}.
\newblock
\showeprint[arxiv]{2207.09455}~[cs.LG]


\bibitem[Cai et~al\mbox{.}(2022)]%
        {devt1}
\bibfield{author}{\bibinfo{person}{Han Cai}, \bibinfo{person}{Ji Lin}, \bibinfo{person}{Yujun Lin}, \bibinfo{person}{Zhijian Liu}, \bibinfo{person}{Haotian Tang}, \bibinfo{person}{Hanrui Wang}, \bibinfo{person}{Ligeng Zhu}, {and} \bibinfo{person}{Song Han}.} \bibinfo{year}{2022}\natexlab{}.
\newblock \showarticletitle{{Enable Deep Learning on Mobile Devices: Methods, Systems, and Applications}}.
\newblock \bibinfo{journal}{\emph{ACM Transactions on Design Automation of Electronic Systems}} \bibinfo{volume}{27}, \bibinfo{number}{3} (\bibinfo{date}{March} \bibinfo{year}{2022}), \bibinfo{pages}{1–50}.
\newblock
\showISSN{1557-7309}
\href{https://doi.org/10.1145/3486618}{doi:\nolinkurl{10.1145/3486618}}


\bibitem[Cai et~al\mbox{.}(2018)]%
        {proxylessnas}
\bibfield{author}{\bibinfo{person}{Han Cai}, \bibinfo{person}{Ligeng Zhu}, {and} \bibinfo{person}{Song Han}.} \bibinfo{year}{2018}\natexlab{}.
\newblock \showarticletitle{{ProxylessNAS: Direct Neural Architecture Search on Target Task and Hardware}}.
\newblock \bibinfo{journal}{\emph{CoRR}}  \bibinfo{volume}{abs/1812.00332} (\bibinfo{year}{2018}).
\newblock
\showeprint[arXiv]{1812.00332}
\urldef\tempurl%
\url{http://arxiv.org/abs/1812.00332}
\showURL{%
\tempurl}


\bibitem[Chen et~al\mbox{.}(2016b)]%
        {devt2}
\bibfield{author}{\bibinfo{person}{Tianqi Chen}, \bibinfo{person}{Bing Xu}, \bibinfo{person}{Chiyuan Zhang}, {and} \bibinfo{person}{Carlos Guestrin}.} \bibinfo{year}{2016}\natexlab{b}.
\newblock \bibinfo{title}{{Training Deep Nets with Sublinear Memory Cost}}.
\newblock
\showeprint[arxiv]{1604.06174}~[cs.LG]


\bibitem[Chen et~al\mbox{.}(2024)]%
        {transformer3}
\bibfield{author}{\bibinfo{person}{Yanxi Chen}, \bibinfo{person}{Xuchen Pan}, \bibinfo{person}{Yaliang Li}, \bibinfo{person}{Bolin Ding}, {and} \bibinfo{person}{Jingren Zhou}.} \bibinfo{year}{2024}\natexlab{}.
\newblock \bibinfo{title}{{EE-LLM: Large-Scale Training and Inference of Early-Exit Large Language Models with 3D Parallelism}}.
\newblock
\showeprint[arxiv]{2312.04916}~[cs.LG]


\bibitem[Chen et~al\mbox{.}(2020)]%
        {yang}
\bibfield{author}{\bibinfo{person}{Yang Chen}, \bibinfo{person}{Xiaoyan Sun}, {and} \bibinfo{person}{Yaochu Jin}.} \bibinfo{year}{2020}\natexlab{}.
\newblock \showarticletitle{{Communication-Efficient Federated Deep Learning With Layerwise Asynchronous Model Update and Temporally Weighted Aggregation}}.
\newblock \bibinfo{journal}{\emph{IEEE Transactions on Neural Networks and Learning Systems}} \bibinfo{volume}{31}, \bibinfo{number}{10} (\bibinfo{date}{Oct.} \bibinfo{year}{2020}), \bibinfo{pages}{4229–4238}.
\newblock
\showISSN{2162-2388}
\href{https://doi.org/10.1109/tnnls.2019.2953131}{doi:\nolinkurl{10.1109/tnnls.2019.2953131}}


\bibitem[Chen et~al\mbox{.}(2016a)]%
        {eyeriss}
\bibfield{author}{\bibinfo{person}{Yu-Hsin Chen}, \bibinfo{person}{Joel Emer}, {and} \bibinfo{person}{Vivienne Sze}.} \bibinfo{year}{2016}\natexlab{a}.
\newblock \showarticletitle{{Eyeriss: A Spatial Architecture for Energy-Efficient Dataflow for Convolutional Neural Networks}}. In \bibinfo{booktitle}{\emph{2016 ACM/IEEE 43rd Annual International Symposium on Computer Architecture (ISCA)}}. \bibinfo{pages}{367--379}.
\newblock
\href{https://doi.org/10.1109/ISCA.2016.40}{doi:\nolinkurl{10.1109/ISCA.2016.40}}


\bibitem[Colin and Lucia(2016)]%
        {int6}
\bibfield{author}{\bibinfo{person}{Alexei Colin} {and} \bibinfo{person}{Brandon Lucia}.} \bibinfo{year}{2016}\natexlab{}.
\newblock \showarticletitle{Chain: tasks and channels for reliable intermittent programs}. In \bibinfo{booktitle}{\emph{Proceedings of the 2016 ACM SIGPLAN International Conference on Object-Oriented Programming, Systems, Languages, and Applications}} (Amsterdam, Netherlands) \emph{(\bibinfo{series}{OOPSLA 2016})}. \bibinfo{publisher}{Association for Computing Machinery}, \bibinfo{address}{New York, NY, USA}, \bibinfo{pages}{514–530}.
\newblock
\showISBNx{9781450344449}
\href{https://doi.org/10.1145/2983990.2983995}{doi:\nolinkurl{10.1145/2983990.2983995}}


\bibitem[Daghero et~al\mbox{.}(2020)]%
        {ee6}
\bibfield{author}{\bibinfo{person}{Francesco Daghero}, \bibinfo{person}{Alessio Burrello}, \bibinfo{person}{Daniele~Jahier Pagliari}, \bibinfo{person}{Luca Benini}, \bibinfo{person}{Enrico Macii}, {and} \bibinfo{person}{Massimo Poncino}.} \bibinfo{year}{2020}\natexlab{}.
\newblock \showarticletitle{{Energy-Efficient Adaptive Machine Learning on IoT End-Nodes With Class-Dependent Confidence}}. In \bibinfo{booktitle}{\emph{2020 27th IEEE International Conference on Electronics, Circuits and Systems (ICECS)}}. \bibinfo{pages}{1--4}.
\newblock
\href{https://doi.org/10.1109/ICECS49266.2020.9294863}{doi:\nolinkurl{10.1109/ICECS49266.2020.9294863}}


\bibitem[Daghero et~al\mbox{.}(2022)]%
        {har}
\bibfield{author}{\bibinfo{person}{Francesco Daghero}, \bibinfo{person}{Alessio Burrello}, \bibinfo{person}{Chen Xie}, \bibinfo{person}{Marco Castellano}, \bibinfo{person}{Luca Gandolfi}, \bibinfo{person}{Andrea Calimera}, \bibinfo{person}{Enrico Macii}, \bibinfo{person}{Massimo Poncino}, {and} \bibinfo{person}{Daniele~Jahier Pagliari}.} \bibinfo{year}{2022}\natexlab{}.
\newblock \showarticletitle{{Human Activity Recognition on Microcontrollers with Quantized and Adaptive Deep Neural Networks}}.
\newblock \bibinfo{journal}{\emph{ACM Transactions on Embedded Computing Systems}} \bibinfo{volume}{21}, \bibinfo{number}{4} (\bibinfo{date}{July} \bibinfo{year}{2022}), \bibinfo{pages}{1–28}.
\newblock
\showISSN{1558-3465}
\href{https://doi.org/10.1145/3542819}{doi:\nolinkurl{10.1145/3542819}}


\bibitem[Dai et~al\mbox{.}(2019)]%
        {chamnet}
\bibfield{author}{\bibinfo{person}{Xiaoliang Dai}, \bibinfo{person}{Peizhao Zhang}, \bibinfo{person}{Bichen Wu}, \bibinfo{person}{Hongxu Yin}, \bibinfo{person}{Fei Sun}, \bibinfo{person}{Yanghan Wang}, \bibinfo{person}{Marat Dukhan}, \bibinfo{person}{Yunqing Hu}, \bibinfo{person}{Yiming Wu}, \bibinfo{person}{Yangqing Jia}, \bibinfo{person}{Peter Vajda}, \bibinfo{person}{Matt Uyttendaele}, {and} \bibinfo{person}{Niraj~K. Jha}.} \bibinfo{year}{2019}\natexlab{}.
\newblock \showarticletitle{{ChamNet: Towards Efficient Network Design Through Platform-Aware Model Adaptation}}. In \bibinfo{booktitle}{\emph{Proceedings of the IEEE/CVF Conference on Computer Vision and Pattern Recognition (CVPR)}}.
\newblock


\bibitem[Delgado and Famaey(2022)]%
        {sched5}
\bibfield{author}{\bibinfo{person}{Carmen Delgado} {and} \bibinfo{person}{Jeroen Famaey}.} \bibinfo{year}{2022}\natexlab{}.
\newblock \showarticletitle{{Optimal Energy-Aware Task Scheduling for Batteryless IoT Devices}}.
\newblock \bibinfo{journal}{\emph{IEEE Transactions on Emerging Topics in Computing}} \bibinfo{volume}{10}, \bibinfo{number}{3} (\bibinfo{year}{2022}), \bibinfo{pages}{1374--1387}.
\newblock
\href{https://doi.org/10.1109/TETC.2021.3086144}{doi:\nolinkurl{10.1109/TETC.2021.3086144}}


\bibitem[Dey et~al\mbox{.}(2019)]%
        {partitioning}
\bibfield{author}{\bibinfo{person}{Swarnava Dey}, \bibinfo{person}{Arijit Mukherjee}, \bibinfo{person}{Arpan Pal}, {and} \bibinfo{person}{Balamuralidhar P}.} \bibinfo{year}{2019}\natexlab{}.
\newblock \showarticletitle{{Embedded Deep Inference in Practice: Case for Model Partitioning}}. In \bibinfo{booktitle}{\emph{Proceedings of the 1st Workshop on Machine Learning on Edge in Sensor Systems}} (New York, NY, USA) \emph{(\bibinfo{series}{SenSys-ML 2019})}. \bibinfo{publisher}{Association for Computing Machinery}, \bibinfo{address}{New York, NY, USA}, \bibinfo{pages}{25–30}.
\newblock
\showISBNx{9781450370110}
\href{https://doi.org/10.1145/3362743.3362964}{doi:\nolinkurl{10.1145/3362743.3362964}}


\bibitem[Dong et~al\mbox{.}(2023)]%
        {etnas}
\bibfield{author}{\bibinfo{person}{Dong Dong}, \bibinfo{person}{Hongxu Jiang}, \bibinfo{person}{Xuekai Wei}, \bibinfo{person}{Yanfei Song}, \bibinfo{person}{Xu Zhuang}, {and} \bibinfo{person}{Jason Wang}.} \bibinfo{year}{2023}\natexlab{}.
\newblock \showarticletitle{{ETNAS: An energy consumption task-driven neural architecture search}}.
\newblock \bibinfo{journal}{\emph{Sustainable Computing: Informatics and Systems}}  \bibinfo{volume}{40} (\bibinfo{year}{2023}), \bibinfo{pages}{100926}.
\newblock
\showISSN{2210-5379}
\href{https://doi.org/10.1016/j.suscom.2023.100926}{doi:\nolinkurl{10.1016/j.suscom.2023.100926}}


\bibitem[Dong and Yang(2020)]%
        {nas_bench_201}
\bibfield{author}{\bibinfo{person}{Xuanyi Dong} {and} \bibinfo{person}{Yi Yang}.} \bibinfo{year}{2020}\natexlab{}.
\newblock \bibinfo{title}{{NAS-Bench-201: Extending the Scope of Reproducible Neural Architecture Search}}.
\newblock
\showeprint[arxiv]{2001.00326}~[cs.CV]


\bibitem[Du et~al\mbox{.}(2020)]%
        {learning2}
\bibfield{author}{\bibinfo{person}{Ying{-}Jun Du}, \bibinfo{person}{Jun Xu}, \bibinfo{person}{Huan Xiong}, \bibinfo{person}{Qiang Qiu}, \bibinfo{person}{Xiantong Zhen}, \bibinfo{person}{Cees G.~M. Snoek}, {and} \bibinfo{person}{Ling Shao}.} \bibinfo{year}{2020}\natexlab{}.
\newblock \showarticletitle{{Learning to Learn with Variational Information Bottleneck for Domain Generalization}}.
\newblock \bibinfo{journal}{\emph{CoRR}}  \bibinfo{volume}{abs/2007.07645} (\bibinfo{year}{2020}).
\newblock
\showeprint[arXiv]{2007.07645}
\urldef\tempurl%
\url{https://arxiv.org/abs/2007.07645}
\showURL{%
\tempurl}


\bibitem[Eastwood et~al\mbox{.}(2022)]%
        {clint}
\bibfield{author}{\bibinfo{person}{Cian Eastwood}, \bibinfo{person}{Ian Mason}, {and} \bibinfo{person}{Christopher K.~I. Williams}.} \bibinfo{year}{2022}\natexlab{}.
\newblock \showarticletitle{{Unit-level surprise in neural networks}}. In \bibinfo{booktitle}{\emph{Proceedings on "I (Still) Can't Believe It's Not Better!" at NeurIPS 2021 Workshops}} \emph{(\bibinfo{series}{Proceedings of Machine Learning Research}, Vol.~\bibinfo{volume}{163})}, \bibfield{editor}{\bibinfo{person}{Melanie~F. Pradier}, \bibinfo{person}{Aaron Schein}, \bibinfo{person}{Stephanie Hyland}, \bibinfo{person}{Francisco J.~R. Ruiz}, {and} \bibinfo{person}{Jessica~Z. Forde}} (Eds.). \bibinfo{publisher}{PMLR}, \bibinfo{pages}{33--40}.
\newblock
\urldef\tempurl%
\url{https://proceedings.mlr.press/v163/eastwood22a.html}
\showURL{%
\tempurl}


\bibitem[Eshratifar et~al\mbox{.}(2018)]%
        {jointdnn}
\bibfield{author}{\bibinfo{person}{Amir~Erfan Eshratifar}, \bibinfo{person}{Mohammad~Saeed Abrishami}, {and} \bibinfo{person}{Massoud Pedram}.} \bibinfo{year}{2018}\natexlab{}.
\newblock \showarticletitle{{JointDNN: An Efficient Training and Inference Engine for Intelligent Mobile Cloud Computing Services}}.
\newblock \bibinfo{journal}{\emph{CoRR}}  \bibinfo{volume}{abs/1801.08618} (\bibinfo{year}{2018}).
\newblock
\showeprint[arXiv]{1801.08618}
\urldef\tempurl%
\url{http://arxiv.org/abs/1801.08618}
\showURL{%
\tempurl}


\bibitem[Fang et~al\mbox{.}(2020)]%
        {ee4}
\bibfield{author}{\bibinfo{person}{Biyi Fang}, \bibinfo{person}{Xiao Zeng}, \bibinfo{person}{Faen Zhang}, \bibinfo{person}{Hui Xu}, {and} \bibinfo{person}{Mi Zhang}.} \bibinfo{year}{2020}\natexlab{}.
\newblock \showarticletitle{{FlexDNN: Input-Adaptive On-Device Deep Learning for Efficient Mobile Vision}}. In \bibinfo{booktitle}{\emph{2020 IEEE/ACM Symposium on Edge Computing (SEC)}}. \bibinfo{pages}{84--95}.
\newblock
\href{https://doi.org/10.1109/SEC50012.2020.00014}{doi:\nolinkurl{10.1109/SEC50012.2020.00014}}


\bibitem[Fawden et~al\mbox{.}(2023)]%
        {fawden}
\bibfield{author}{\bibinfo{person}{Terry Fawden}, \bibinfo{person}{Lorena Qendro}, {and} \bibinfo{person}{Cecilia Mascolo}.} \bibinfo{year}{2023}\natexlab{}.
\newblock \showarticletitle{{Uncertainty-Informed On-Device Personalisation Using Early Exit Networks on Sensor Signals}}.
\newblock \bibinfo{journal}{\emph{2023 31st European Signal Processing Conference (EUSIPCO)}} (\bibinfo{year}{2023}), \bibinfo{pages}{1305--1309}.
\newblock
\urldef\tempurl%
\url{https://api.semanticscholar.org/CorpusID:261116963}
\showURL{%
\tempurl}


\bibitem[Figurnov et~al\mbox{.}(2016)]%
        {figurnov}
\bibfield{author}{\bibinfo{person}{Michael Figurnov}, \bibinfo{person}{Maxwell~D. Collins}, \bibinfo{person}{Yukun Zhu}, \bibinfo{person}{Li Zhang}, \bibinfo{person}{Jonathan Huang}, \bibinfo{person}{Dmitry~P. Vetrov}, {and} \bibinfo{person}{Ruslan Salakhutdinov}.} \bibinfo{year}{2016}\natexlab{}.
\newblock \showarticletitle{{Spatially Adaptive Computation Time for Residual Networks}}.
\newblock \bibinfo{journal}{\emph{CoRR}}  \bibinfo{volume}{abs/1612.02297} (\bibinfo{year}{2016}).
\newblock
\showeprint[arXiv]{1612.02297}
\urldef\tempurl%
\url{http://arxiv.org/abs/1612.02297}
\showURL{%
\tempurl}


\bibitem[Fraternali et~al\mbox{.}(2020)]%
        {ACES}
\bibfield{author}{\bibinfo{person}{Francesco Fraternali}, \bibinfo{person}{Bharathan Balaji}, \bibinfo{person}{Yuvraj Agarwal}, {and} \bibinfo{person}{Rajesh~K. Gupta}.} \bibinfo{year}{2020}\natexlab{}.
\newblock \showarticletitle{{ACES: Automatic Configuration of Energy Harvesting Sensors with Reinforcement Learning}}.
\newblock \bibinfo{journal}{\emph{ACM Transactions on Sensor Networks}} \bibinfo{volume}{16}, \bibinfo{number}{4} (\bibinfo{date}{July} \bibinfo{year}{2020}), \bibinfo{pages}{1–31}.
\newblock
\showISSN{1550-4867}
\href{https://doi.org/10.1145/3404191}{doi:\nolinkurl{10.1145/3404191}}


\bibitem[Gao et~al\mbox{.}(2018)]%
        {fbs}
\bibfield{author}{\bibinfo{person}{Xitong Gao}, \bibinfo{person}{Yiren Zhao}, \bibinfo{person}{Lukasz Dudziak}, \bibinfo{person}{Robert~D. Mullins}, {and} \bibinfo{person}{Cheng{-}Zhong Xu}.} \bibinfo{year}{2018}\natexlab{}.
\newblock \showarticletitle{{Dynamic Channel Pruning: Feature Boosting and Suppression}}.
\newblock \bibinfo{journal}{\emph{CoRR}}  \bibinfo{volume}{abs/1810.05331} (\bibinfo{year}{2018}).
\newblock
\showeprint[arXiv]{1810.05331}
\urldef\tempurl%
\url{http://arxiv.org/abs/1810.05331}
\showURL{%
\tempurl}


\bibitem[Gim and Ko(2022)]%
        {devt3}
\bibfield{author}{\bibinfo{person}{In Gim} {and} \bibinfo{person}{JeongGil Ko}.} \bibinfo{year}{2022}\natexlab{}.
\newblock \showarticletitle{{Memory-efficient DNN training on mobile devices}}. In \bibinfo{booktitle}{\emph{Proceedings of the 20th Annual International Conference on Mobile Systems, Applications and Services}} (Portland, Oregon) \emph{(\bibinfo{series}{MobiSys '22})}. \bibinfo{publisher}{Association for Computing Machinery}, \bibinfo{address}{New York, NY, USA}, \bibinfo{pages}{464–476}.
\newblock
\showISBNx{9781450391856}
\href{https://doi.org/10.1145/3498361.3539765}{doi:\nolinkurl{10.1145/3498361.3539765}}


\bibitem[Giordano et~al\mbox{.}(2020)]%
        {face_rec1}
\bibfield{author}{\bibinfo{person}{Marco Giordano}, \bibinfo{person}{Philipp Mayer}, {and} \bibinfo{person}{Michele Magno}.} \bibinfo{year}{2020}\natexlab{}.
\newblock \showarticletitle{{A Battery-Free Long-Range Wireless Smart Camera for Face Detection}}. In \bibinfo{booktitle}{\emph{Proceedings of the 8th International Workshop on Energy Harvesting and Energy-Neutral Sensing Systems}} (Virtual Event, Japan) \emph{(\bibinfo{series}{ENSsys '20})}. \bibinfo{publisher}{Association for Computing Machinery}, \bibinfo{address}{New York, NY, USA}, \bibinfo{pages}{29–35}.
\newblock
\showISBNx{9781450381291}
\href{https://doi.org/10.1145/3417308.3430273}{doi:\nolinkurl{10.1145/3417308.3430273}}


\bibitem[Gobieski et~al\mbox{.}(2018)]%
        {sonic}
\bibfield{author}{\bibinfo{person}{Graham Gobieski}, \bibinfo{person}{Nathan Beckmann}, {and} \bibinfo{person}{Brandon Lucia}.} \bibinfo{year}{2018}\natexlab{}.
\newblock \showarticletitle{{Intelligence Beyond the Edge: Inference on Intermittent Embedded Systems}}.
\newblock \bibinfo{journal}{\emph{CoRR}}  \bibinfo{volume}{abs/1810.07751} (\bibinfo{year}{2018}).
\newblock
\showeprint[arXiv]{1810.07751}
\urldef\tempurl%
\url{http://arxiv.org/abs/1810.07751}
\showURL{%
\tempurl}


\bibitem[Google(2024)]%
        {tflitemicro}
\bibfield{author}{\bibinfo{person}{Google}.} \bibinfo{year}{2024}\natexlab{}.
\newblock \bibinfo{title}{{TFLite-Micro}}.
\newblock \bibinfo{howpublished}{\url{https://github.com/tensorflow/tflite-micro}}.
\newblock


\bibitem[Guler and Yener(2021)]%
        {susfl}
\bibfield{author}{\bibinfo{person}{Basak Guler} {and} \bibinfo{person}{Aylin Yener}.} \bibinfo{year}{2021}\natexlab{}.
\newblock \showarticletitle{{Sustainable Federated Learning}}.
\newblock \bibinfo{journal}{\emph{CoRR}}  \bibinfo{volume}{abs/2102.11274} (\bibinfo{year}{2021}).
\newblock
\showeprint[arXiv]{2102.11274}
\urldef\tempurl%
\url{https://arxiv.org/abs/2102.11274}
\showURL{%
\tempurl}


\bibitem[Guo et~al\mbox{.}(2018)]%
        {spottune}
\bibfield{author}{\bibinfo{person}{Yunhui Guo}, \bibinfo{person}{Honghui Shi}, \bibinfo{person}{Abhishek Kumar}, \bibinfo{person}{Kristen Grauman}, \bibinfo{person}{Tajana Rosing}, {and} \bibinfo{person}{Rog{\'{e}}rio~Schmidt Feris}.} \bibinfo{year}{2018}\natexlab{}.
\newblock \showarticletitle{{SpotTune: Transfer Learning through Adaptive Fine-tuning}}.
\newblock \bibinfo{journal}{\emph{CoRR}}  \bibinfo{volume}{abs/1811.08737} (\bibinfo{year}{2018}).
\newblock
\showeprint[arXiv]{1811.08737}
\urldef\tempurl%
\url{http://arxiv.org/abs/1811.08737}
\showURL{%
\tempurl}


\bibitem[Han et~al\mbox{.}(2020)]%
        {sparse1}
\bibfield{author}{\bibinfo{person}{Pengchao Han}, \bibinfo{person}{Shiqiang Wang}, {and} \bibinfo{person}{Kin~K. Leung}.} \bibinfo{year}{2020}\natexlab{}.
\newblock \showarticletitle{{Adaptive Gradient Sparsification for Efficient Federated Learning: An Online Learning Approach}}.
\newblock \bibinfo{journal}{\emph{CoRR}}  \bibinfo{volume}{abs/2001.04756} (\bibinfo{year}{2020}).
\newblock
\showeprint[arXiv]{2001.04756}
\urldef\tempurl%
\url{https://arxiv.org/abs/2001.04756}
\showURL{%
\tempurl}


\bibitem[Han et~al\mbox{.}(2016a)]%
        {deepcomp}
\bibfield{author}{\bibinfo{person}{Song Han}, \bibinfo{person}{Huizi Mao}, {and} \bibinfo{person}{William~J. Dally}.} \bibinfo{year}{2016}\natexlab{a}.
\newblock \showarticletitle{{Deep Compression: Compressing Deep Neural Network with Pruning, Trained Quantization and Huffman Coding}}. In \bibinfo{booktitle}{\emph{4th International Conference on Learning Representations, {ICLR} 2016, San Juan, Puerto Rico, May 2-4, 2016, Conference Track Proceedings}}, \bibfield{editor}{\bibinfo{person}{Yoshua Bengio} {and} \bibinfo{person}{Yann LeCun}} (Eds.).
\newblock
\urldef\tempurl%
\url{http://arxiv.org/abs/1510.00149}
\showURL{%
\tempurl}


\bibitem[Han et~al\mbox{.}(2016b)]%
        {comp1}
\bibfield{author}{\bibinfo{person}{Song Han}, \bibinfo{person}{Huizi Mao}, {and} \bibinfo{person}{William~J. Dally}.} \bibinfo{year}{2016}\natexlab{b}.
\newblock \bibinfo{title}{{Deep Compression: Compressing Deep Neural Networks with Pruning, Trained Quantization and Huffman Coding}}.
\newblock
\showeprint[arxiv]{1510.00149}~[cs.CV]


\bibitem[Han et~al\mbox{.}(2015)]%
        {dally}
\bibfield{author}{\bibinfo{person}{Song Han}, \bibinfo{person}{Jeff Pool}, \bibinfo{person}{John Tran}, {and} \bibinfo{person}{William~J. Dally}.} \bibinfo{year}{2015}\natexlab{}.
\newblock \showarticletitle{{Learning both Weights and Connections for Efficient Neural Networks}}.
\newblock \bibinfo{journal}{\emph{CoRR}}  \bibinfo{volume}{abs/1506.02626} (\bibinfo{year}{2015}).
\newblock
\showeprint[arXiv]{1506.02626}
\urldef\tempurl%
\url{http://arxiv.org/abs/1506.02626}
\showURL{%
\tempurl}


\bibitem[Hasan(2022)]%
        {iotanal}
\bibfield{author}{\bibinfo{person}{Mohammad Hasan}.} \bibinfo{year}{2022}\natexlab{}.
\newblock \showarticletitle{{State of IoT-Spring 2022}}.
\newblock \bibinfo{journal}{\emph{IOT Analytics, See https://iot-analytics. com/product/state-of-iot-spring-2022 website}} (\bibinfo{year}{2022}).
\newblock


\bibitem[He et~al\mbox{.}(2015a)]%
        {xiangyu}
\bibfield{author}{\bibinfo{person}{Kaiming He}, \bibinfo{person}{Xiangyu Zhang}, \bibinfo{person}{Shaoqing Ren}, {and} \bibinfo{person}{Jian Sun}.} \bibinfo{year}{2015}\natexlab{a}.
\newblock \showarticletitle{{Deep Residual Learning for Image Recognition}}.
\newblock \bibinfo{journal}{\emph{CoRR}}  \bibinfo{volume}{abs/1512.03385} (\bibinfo{year}{2015}).
\newblock
\showeprint[arXiv]{1512.03385}
\urldef\tempurl%
\url{http://arxiv.org/abs/1512.03385}
\showURL{%
\tempurl}


\bibitem[He et~al\mbox{.}(2015b)]%
        {resnet}
\bibfield{author}{\bibinfo{person}{Kaiming He}, \bibinfo{person}{Xiangyu Zhang}, \bibinfo{person}{Shaoqing Ren}, {and} \bibinfo{person}{Jian Sun}.} \bibinfo{year}{2015}\natexlab{b}.
\newblock \bibinfo{title}{{Deep Residual Learning for Image Recognition}}.
\newblock
\showeprint[arxiv]{1512.03385}~[cs.CV]


\bibitem[Hester et~al\mbox{.}(2017)]%
        {int5}
\bibfield{author}{\bibinfo{person}{Josiah Hester}, \bibinfo{person}{Kevin Storer}, {and} \bibinfo{person}{Jacob Sorber}.} \bibinfo{year}{2017}\natexlab{}.
\newblock \showarticletitle{{Timely Execution on Intermittently Powered Batteryless Sensors}}. \bibinfo{pages}{1--13}.
\newblock
\href{https://doi.org/10.1145/3131672.3131673}{doi:\nolinkurl{10.1145/3131672.3131673}}


\bibitem[Hicks(2017)]%
        {int2}
\bibfield{author}{\bibinfo{person}{Matthew Hicks}.} \bibinfo{year}{2017}\natexlab{}.
\newblock \showarticletitle{{Clank: Architectural support for intermittent computation}}. In \bibinfo{booktitle}{\emph{2017 ACM/IEEE 44th Annual International Symposium on Computer Architecture (ISCA)}}. \bibinfo{pages}{228--240}.
\newblock
\href{https://doi.org/10.1145/3079856.3080238}{doi:\nolinkurl{10.1145/3079856.3080238}}


\bibitem[Howard et~al\mbox{.}(2017)]%
        {mobilenetv1}
\bibfield{author}{\bibinfo{person}{Andrew~G. Howard}, \bibinfo{person}{Menglong Zhu}, \bibinfo{person}{Bo Chen}, \bibinfo{person}{Dmitry Kalenichenko}, \bibinfo{person}{Weijun Wang}, \bibinfo{person}{Tobias Weyand}, \bibinfo{person}{Marco Andreetto}, {and} \bibinfo{person}{Hartwig Adam}.} \bibinfo{year}{2017}\natexlab{}.
\newblock \bibinfo{title}{{MobileNets: Efficient Convolutional Neural Networks for Mobile Vision Applications}}.
\newblock
\showeprint[arxiv]{1704.04861}~[cs.CV]


\bibitem[Hu et~al\mbox{.}(2024)]%
        {cdmpp}
\bibfield{author}{\bibinfo{person}{Hanpeng Hu}, \bibinfo{person}{Junwei Su}, \bibinfo{person}{Juntao Zhao}, \bibinfo{person}{Yanghua Peng}, \bibinfo{person}{Yibo Zhu}, \bibinfo{person}{Haibin Lin}, {and} \bibinfo{person}{Chuan Wu}.} \bibinfo{year}{2024}\natexlab{}.
\newblock \showarticletitle{{CDMPP: A Device-Model Agnostic Framework for Latency Prediction of Tensor Programs}}. In \bibinfo{booktitle}{\emph{Proceedings of the Nineteenth European Conference on Computer Systems}} \emph{(\bibinfo{series}{EuroSys ’24})}. \bibinfo{publisher}{ACM}.
\newblock
\href{https://doi.org/10.1145/3627703.3629572}{doi:\nolinkurl{10.1145/3627703.3629572}}


\bibitem[Hua et~al\mbox{.}(2018)]%
        {hua}
\bibfield{author}{\bibinfo{person}{Weizhe Hua}, \bibinfo{person}{Christopher~De Sa}, \bibinfo{person}{Zhiru Zhang}, {and} \bibinfo{person}{G.~Edward Suh}.} \bibinfo{year}{2018}\natexlab{}.
\newblock \showarticletitle{{Channel Gating Neural Networks}}.
\newblock \bibinfo{journal}{\emph{CoRR}}  \bibinfo{volume}{abs/1805.12549} (\bibinfo{year}{2018}).
\newblock
\showeprint[arXiv]{1805.12549}
\urldef\tempurl%
\url{http://arxiv.org/abs/1805.12549}
\showURL{%
\tempurl}


\bibitem[Huang et~al\mbox{.}(2017a)]%
        {ee2}
\bibfield{author}{\bibinfo{person}{Gao Huang}, \bibinfo{person}{Danlu Chen}, \bibinfo{person}{Tianhong Li}, \bibinfo{person}{Felix Wu}, \bibinfo{person}{Laurens van~der Maaten}, {and} \bibinfo{person}{Kilian~Q. Weinberger}.} \bibinfo{year}{2017}\natexlab{a}.
\newblock \showarticletitle{{Multi-Scale Dense Convolutional Networks for Efficient Prediction}}.
\newblock \bibinfo{journal}{\emph{CoRR}}  \bibinfo{volume}{abs/1703.09844} (\bibinfo{year}{2017}).
\newblock
\showeprint[arXiv]{1703.09844}
\urldef\tempurl%
\url{http://arxiv.org/abs/1703.09844}
\showURL{%
\tempurl}


\bibitem[Huang et~al\mbox{.}(2017b)]%
        {ee7}
\bibfield{author}{\bibinfo{person}{Gao Huang}, \bibinfo{person}{Danlu Chen}, \bibinfo{person}{Tianhong Li}, \bibinfo{person}{Felix Wu}, \bibinfo{person}{Laurens van~der Maaten}, {and} \bibinfo{person}{Kilian~Q. Weinberger}.} \bibinfo{year}{2017}\natexlab{b}.
\newblock \showarticletitle{{Multi-Scale Dense Convolutional Networks for Efficient Prediction}}.
\newblock \bibinfo{journal}{\emph{CoRR}}  \bibinfo{volume}{abs/1703.09844} (\bibinfo{year}{2017}).
\newblock
\showeprint[arXiv]{1703.09844}
\urldef\tempurl%
\url{http://arxiv.org/abs/1703.09844}
\showURL{%
\tempurl}


\bibitem[Iandola et~al\mbox{.}(2016)]%
        {squeezenet}
\bibfield{author}{\bibinfo{person}{Forrest~N. Iandola}, \bibinfo{person}{Matthew~W. Moskewicz}, \bibinfo{person}{Khalid Ashraf}, \bibinfo{person}{Song Han}, \bibinfo{person}{William~J. Dally}, {and} \bibinfo{person}{Kurt Keutzer}.} \bibinfo{year}{2016}\natexlab{}.
\newblock \showarticletitle{{SqueezeNet: AlexNet-level accuracy with 50x fewer parameters and {\textless}1MB model size}}.
\newblock \bibinfo{journal}{\emph{CoRR}}  \bibinfo{volume}{abs/1602.07360} (\bibinfo{year}{2016}).
\newblock
\showeprint[arXiv]{1602.07360}
\urldef\tempurl%
\url{http://arxiv.org/abs/1602.07360}
\showURL{%
\tempurl}


\bibitem[Islam and Nirjon(2020)]%
        {zygarde}
\bibfield{author}{\bibinfo{person}{Bashima Islam} {and} \bibinfo{person}{Shahriar Nirjon}.} \bibinfo{year}{2020}\natexlab{}.
\newblock \showarticletitle{{Zygarde: Time-Sensitive On-Device Deep Inference and Adaptation on Intermittently-Powered Systems}}.
\newblock \bibinfo{journal}{\emph{Proc. ACM Interact. Mob. Wearable Ubiquitous Technol.}} \bibinfo{volume}{4}, \bibinfo{number}{3}, Article \bibinfo{articleno}{82} (\bibinfo{date}{sep} \bibinfo{year}{2020}), \bibinfo{numpages}{29}~pages.
\newblock
\href{https://doi.org/10.1145/3411808}{doi:\nolinkurl{10.1145/3411808}}


\bibitem[Jaafra et~al\mbox{.}(2019)]%
        {reinforcement}
\bibfield{author}{\bibinfo{person}{Yesmina Jaafra}, \bibinfo{person}{Jean {Luc Laurent}}, \bibinfo{person}{Aline Deruyver}, {and} \bibinfo{person}{Mohamed {Saber Naceur}}.} \bibinfo{year}{2019}\natexlab{}.
\newblock \showarticletitle{{Reinforcement learning for neural architecture search: A review}}.
\newblock \bibinfo{journal}{\emph{Image and Vision Computing}}  \bibinfo{volume}{89} (\bibinfo{year}{2019}), \bibinfo{pages}{57--66}.
\newblock
\showISSN{0262-8856}
\href{https://doi.org/10.1016/j.imavis.2019.06.005}{doi:\nolinkurl{10.1016/j.imavis.2019.06.005}}


\bibitem[Jahier~Pagliari et~al\mbox{.}(2020)]%
        {bw2}
\bibfield{author}{\bibinfo{person}{Daniele Jahier~Pagliari}, \bibinfo{person}{Francesco Daghero}, {and} \bibinfo{person}{Massimo Poncino}.} \bibinfo{year}{2020}\natexlab{}.
\newblock \showarticletitle{Sequence-To-Sequence Neural Networks Inference on Embedded Processors Using Dynamic Beam Search}.
\newblock \bibinfo{journal}{\emph{Electronics}} \bibinfo{volume}{9}, \bibinfo{number}{2} (\bibinfo{year}{2020}).
\newblock
\showISSN{2079-9292}
\href{https://doi.org/10.3390/electronics9020337}{doi:\nolinkurl{10.3390/electronics9020337}}


\bibitem[Jahier~Pagliari et~al\mbox{.}(2019a)]%
        {pagliari2}
\bibfield{author}{\bibinfo{person}{Daniele Jahier~Pagliari}, \bibinfo{person}{Francesco Panini}, \bibinfo{person}{Enrico Macii}, {and} \bibinfo{person}{Massimo Poncino}.} \bibinfo{year}{2019}\natexlab{a}.
\newblock \showarticletitle{{Dynamic Beam Width Tuning for Energy-Efficient Recurrent Neural Networks}}. In \bibinfo{booktitle}{\emph{Proceedings of the 2019 on Great Lakes Symposium on VLSI}} (Tysons Corner, VA, USA) \emph{(\bibinfo{series}{GLSVLSI '19})}. \bibinfo{publisher}{Association for Computing Machinery}, \bibinfo{address}{New York, NY, USA}, \bibinfo{pages}{69–74}.
\newblock
\showISBNx{9781450362528}
\href{https://doi.org/10.1145/3299874.3317974}{doi:\nolinkurl{10.1145/3299874.3317974}}


\bibitem[Jahier~Pagliari et~al\mbox{.}(2019b)]%
        {bw1}
\bibfield{author}{\bibinfo{person}{Daniele Jahier~Pagliari}, \bibinfo{person}{Francesco Panini}, \bibinfo{person}{Enrico Macii}, {and} \bibinfo{person}{Massimo Poncino}.} \bibinfo{year}{2019}\natexlab{b}.
\newblock \showarticletitle{Dynamic Beam Width Tuning for Energy-Efficient Recurrent Neural Networks}. In \bibinfo{booktitle}{\emph{Proceedings of the 2019 on Great Lakes Symposium on VLSI}} (Tysons Corner, VA, USA) \emph{(\bibinfo{series}{GLSVLSI '19})}. \bibinfo{publisher}{Association for Computing Machinery}, \bibinfo{address}{New York, NY, USA}, \bibinfo{pages}{69–74}.
\newblock
\showISBNx{9781450362528}
\href{https://doi.org/10.1145/3299874.3317974}{doi:\nolinkurl{10.1145/3299874.3317974}}


\bibitem[Jeon et~al\mbox{.}(2023)]%
        {harvnet}
\bibfield{author}{\bibinfo{person}{Seunghyeok Jeon}, \bibinfo{person}{Yonghun Choi}, \bibinfo{person}{Yeonwoo Cho}, {and} \bibinfo{person}{Hojung Cha}.} \bibinfo{year}{2023}\natexlab{}.
\newblock \showarticletitle{{HarvNet: Resource-Optimized Operation of Multi-Exit Deep Neural Networks on Energy Harvesting Devices}}. In \bibinfo{booktitle}{\emph{Proceedings of the 21st Annual International Conference on Mobile Systems, Applications and Services}} (Helsinki, Finland) \emph{(\bibinfo{series}{MobiSys '23})}. \bibinfo{publisher}{Association for Computing Machinery}, \bibinfo{address}{New York, NY, USA}, \bibinfo{pages}{42–55}.
\newblock
\showISBNx{9798400701108}
\href{https://doi.org/10.1145/3581791.3596845}{doi:\nolinkurl{10.1145/3581791.3596845}}


\bibitem[Jiang et~al\mbox{.}(2018)]%
        {jiang}
\bibfield{author}{\bibinfo{person}{Heinrich Jiang}, \bibinfo{person}{Been Kim}, \bibinfo{person}{Melody~Y. Guan}, {and} \bibinfo{person}{Maya Gupta}.} \bibinfo{year}{2018}\natexlab{}.
\newblock \bibinfo{title}{{To Trust Or Not To Trust A Classifier}}.
\newblock
\showeprint[arxiv]{1805.11783}~[stat.ML]


\bibitem[Jokic et~al\mbox{.}(2021)]%
        {face_rec2}
\bibfield{author}{\bibinfo{person}{Petar Jokic}, \bibinfo{person}{Stephane Emery}, {and} \bibinfo{person}{Luca Benini}.} \bibinfo{year}{2021}\natexlab{}.
\newblock \showarticletitle{{Battery-Less Face Recognition at the Extreme Edge}}. In \bibinfo{booktitle}{\emph{2021 19th IEEE International New Circuits and Systems Conference (NEWCAS)}}. \bibinfo{pages}{1--4}.
\newblock
\href{https://doi.org/10.1109/NEWCAS50681.2021.9462787}{doi:\nolinkurl{10.1109/NEWCAS50681.2021.9462787}}


\bibitem[Kang et~al\mbox{.}(2021)]%
        {ecq}
\bibfield{author}{\bibinfo{person}{Beomseok Kang}, \bibinfo{person}{Anni Lu}, \bibinfo{person}{Yun Long}, \bibinfo{person}{Daehyun Kim}, \bibinfo{person}{Shimeng Yu}, {and} \bibinfo{person}{Saibal Mukhopadhyay}.} \bibinfo{year}{2021}\natexlab{}.
\newblock \showarticletitle{{Genetic Algorithm-Based Energy-Aware CNN Quantization for Processing-In-Memory Architecture}}.
\newblock \bibinfo{journal}{\emph{IEEE Journal on Emerging and Selected Topics in Circuits and Systems}} \bibinfo{volume}{11}, \bibinfo{number}{4} (\bibinfo{year}{2021}), \bibinfo{pages}{649--662}.
\newblock
\href{https://doi.org/10.1109/JETCAS.2021.3127129}{doi:\nolinkurl{10.1109/JETCAS.2021.3127129}}


\bibitem[Kang et~al\mbox{.}(2017)]%
        {neurosurgeon}
\bibfield{author}{\bibinfo{person}{Yiping Kang}, \bibinfo{person}{Johann Hauswald}, \bibinfo{person}{Cao Gao}, \bibinfo{person}{Austin Rovinski}, \bibinfo{person}{Trevor Mudge}, \bibinfo{person}{Jason Mars}, {and} \bibinfo{person}{Lingjia Tang}.} \bibinfo{year}{2017}\natexlab{}.
\newblock \showarticletitle{{Neurosurgeon: Collaborative Intelligence Between the Cloud and Mobile Edge}}.
\newblock \bibinfo{journal}{\emph{SIGPLAN Not.}} \bibinfo{volume}{52}, \bibinfo{number}{4} (\bibinfo{date}{apr} \bibinfo{year}{2017}), \bibinfo{pages}{615–629}.
\newblock
\showISSN{0362-1340}
\href{https://doi.org/10.1145/3093336.3037698}{doi:\nolinkurl{10.1145/3093336.3037698}}


\bibitem[Karimi et~al\mbox{.}(2021)]%
        {sched3}
\bibfield{author}{\bibinfo{person}{Mohsen Karimi}, \bibinfo{person}{Hyunjong Choi}, \bibinfo{person}{Yidi Wang}, \bibinfo{person}{Yecheng Xiang}, {and} \bibinfo{person}{Hyoseung Kim}.} \bibinfo{year}{2021}\natexlab{}.
\newblock \showarticletitle{{Real-Time Task Scheduling on Intermittently Powered Batteryless Devices}}.
\newblock \bibinfo{journal}{\emph{IEEE Internet of Things Journal}} \bibinfo{volume}{8}, \bibinfo{number}{17} (\bibinfo{year}{2021}), \bibinfo{pages}{13328--13342}.
\newblock
\href{https://doi.org/10.1109/JIOT.2021.3065947}{doi:\nolinkurl{10.1109/JIOT.2021.3065947}}


\bibitem[Kaya and Dumitras(2018)]%
        {kaya}
\bibfield{author}{\bibinfo{person}{Yigitcan Kaya} {and} \bibinfo{person}{Tudor Dumitras}.} \bibinfo{year}{2018}\natexlab{}.
\newblock \showarticletitle{{How to Stop Off-the-Shelf Deep Neural Networks from Overthinking}}.
\newblock \bibinfo{journal}{\emph{CoRR}}  \bibinfo{volume}{abs/1810.07052} (\bibinfo{year}{2018}).
\newblock
\showeprint[arXiv]{1810.07052}
\urldef\tempurl%
\url{http://arxiv.org/abs/1810.07052}
\showURL{%
\tempurl}


\bibitem[Khoshsirat et~al\mbox{.}(2023)]%
        {divide_and_save}
\bibfield{author}{\bibinfo{person}{Aria Khoshsirat}, \bibinfo{person}{Giovanni Perin}, {and} \bibinfo{person}{Michele Rossi}.} \bibinfo{year}{2023}\natexlab{}.
\newblock \bibinfo{title}{{Divide and Save: Splitting Workload Among Containers in an Edge Device to Save Energy and Time}}.
\newblock
\showeprint[arxiv]{2302.06478}~[cs.DC]


\bibitem[Kim et~al\mbox{.}(2022)]%
        {pim_overview}
\bibfield{author}{\bibinfo{person}{Donghyuk Kim}, \bibinfo{person}{Chengshuo Yu}, \bibinfo{person}{Shanshan Xie}, \bibinfo{person}{Yuzong Chen}, \bibinfo{person}{Joo-Young Kim}, \bibinfo{person}{Bongjin Kim}, \bibinfo{person}{Jaydeep~P. Kulkarni}, {and} \bibinfo{person}{Tony Tae-Hyoung Kim}.} \bibinfo{year}{2022}\natexlab{}.
\newblock \showarticletitle{{An Overview of Processing-in-Memory Circuits for Artificial Intelligence and Machine Learning}}.
\newblock \bibinfo{journal}{\emph{IEEE Journal on Emerging and Selected Topics in Circuits and Systems}} \bibinfo{volume}{12}, \bibinfo{number}{2} (\bibinfo{year}{2022}), \bibinfo{pages}{338--353}.
\newblock
\href{https://doi.org/10.1109/JETCAS.2022.3160455}{doi:\nolinkurl{10.1109/JETCAS.2022.3160455}}


\bibitem[Kirkpatrick et~al\mbox{.}(2016)]%
        {kirkpatrick}
\bibfield{author}{\bibinfo{person}{James Kirkpatrick}, \bibinfo{person}{Razvan Pascanu}, \bibinfo{person}{Neil~C. Rabinowitz}, \bibinfo{person}{Joel Veness}, \bibinfo{person}{Guillaume Desjardins}, \bibinfo{person}{Andrei~A. Rusu}, \bibinfo{person}{Kieran Milan}, \bibinfo{person}{John Quan}, \bibinfo{person}{Tiago Ramalho}, \bibinfo{person}{Agnieszka Grabska{-}Barwinska}, \bibinfo{person}{Demis Hassabis}, \bibinfo{person}{Claudia Clopath}, \bibinfo{person}{Dharshan Kumaran}, {and} \bibinfo{person}{Raia Hadsell}.} \bibinfo{year}{2016}\natexlab{}.
\newblock \showarticletitle{{Overcoming catastrophic forgetting in neural networks}}.
\newblock \bibinfo{journal}{\emph{CoRR}}  \bibinfo{volume}{abs/1612.00796} (\bibinfo{year}{2016}).
\newblock
\showeprint[arXiv]{1612.00796}
\urldef\tempurl%
\url{http://arxiv.org/abs/1612.00796}
\showURL{%
\tempurl}


\bibitem[Krishnamoorthi(2018)]%
        {comp2}
\bibfield{author}{\bibinfo{person}{Raghuraman Krishnamoorthi}.} \bibinfo{year}{2018}\natexlab{}.
\newblock \bibinfo{title}{{Quantizing deep convolutional networks for efficient inference: A whitepaper}}.
\newblock
\showeprint[arxiv]{1806.08342}~[cs.LG]


\bibitem[Krizhevsky et~al\mbox{.}(2012)]%
        {alexnet}
\bibfield{author}{\bibinfo{person}{Alex Krizhevsky}, \bibinfo{person}{Ilya Sutskever}, {and} \bibinfo{person}{Geoffrey~E Hinton}.} \bibinfo{year}{2012}\natexlab{}.
\newblock \showarticletitle{{ImageNet Classification with Deep Convolutional Neural Networks}}. In \bibinfo{booktitle}{\emph{Advances in Neural Information Processing Systems}}, \bibfield{editor}{\bibinfo{person}{F.~Pereira}, \bibinfo{person}{C.J. Burges}, \bibinfo{person}{L.~Bottou}, {and} \bibinfo{person}{K.Q. Weinberger}} (Eds.), Vol.~\bibinfo{volume}{25}. \bibinfo{publisher}{Curran Associates, Inc.}
\newblock
\urldef\tempurl%
\url{https://proceedings.neurips.cc/paper_files/paper/2012/file/c399862d3b9d6b76c8436e924a68c45b-Paper.pdf}
\showURL{%
\tempurl}


\bibitem[Kumar et~al\mbox{.}(2022)]%
        {kumar}
\bibfield{author}{\bibinfo{person}{Ananya Kumar}, \bibinfo{person}{Aditi Raghunathan}, \bibinfo{person}{Robbie Jones}, \bibinfo{person}{Tengyu Ma}, {and} \bibinfo{person}{Percy Liang}.} \bibinfo{year}{2022}\natexlab{}.
\newblock \bibinfo{title}{{Fine-Tuning can Distort Pretrained Features and Underperform Out-of-Distribution}}.
\newblock
\showeprint[arxiv]{2202.10054}~[cs.LG]


\bibitem[Kwon et~al\mbox{.}(2023a)]%
        {lifelearner}
\bibfield{author}{\bibinfo{person}{Young~D. Kwon}, \bibinfo{person}{Jagmohan Chauhan}, \bibinfo{person}{Hong Jia}, \bibinfo{person}{Stylianos~I. Venieris}, {and} \bibinfo{person}{Cecilia Mascolo}.} \bibinfo{year}{2023}\natexlab{a}.
\newblock \bibinfo{title}{{LifeLearner: Hardware-Aware Meta Continual Learning System for Embedded Computing Platforms}}.
\newblock
\showeprint[arxiv]{2311.11420}~[cs.LG]


\bibitem[Kwon et~al\mbox{.}(2023b)]%
        {devt4}
\bibfield{author}{\bibinfo{person}{Young~D. Kwon}, \bibinfo{person}{Rui Li}, \bibinfo{person}{Stylianos~I. Venieris}, \bibinfo{person}{Jagmohan Chauhan}, \bibinfo{person}{Nicholas~D. Lane}, {and} \bibinfo{person}{Cecilia Mascolo}.} \bibinfo{year}{2023}\natexlab{b}.
\newblock \bibinfo{title}{{TinyTrain: Deep Neural Network Training at the Extreme Edge}}.
\newblock
\showeprint[arxiv]{2307.09988}~[cs.LG]


\bibitem[Laskaridis et~al\mbox{.}(2021)]%
        {ee}
\bibfield{author}{\bibinfo{person}{Stefanos Laskaridis}, \bibinfo{person}{Alexandros Kouris}, {and} \bibinfo{person}{Nicholas~D. Lane}.} \bibinfo{year}{2021}\natexlab{}.
\newblock \showarticletitle{{Adaptive Inference through Early-Exit Networks: Design, Challenges and Directions}}. In \bibinfo{booktitle}{\emph{Proceedings of the 5th International Workshop on Embedded and Mobile Deep Learning}} (Virtual, WI, USA) \emph{(\bibinfo{series}{EMDL'21})}. \bibinfo{publisher}{Association for Computing Machinery}, \bibinfo{address}{New York, NY, USA}, \bibinfo{pages}{1–6}.
\newblock
\showISBNx{9781450385978}
\href{https://doi.org/10.1145/3469116.3470012}{doi:\nolinkurl{10.1145/3469116.3470012}}


\bibitem[Laskaridis et~al\mbox{.}(2020)]%
        {ee5}
\bibfield{author}{\bibinfo{person}{Stefanos Laskaridis}, \bibinfo{person}{Stylianos~I. Venieris}, \bibinfo{person}{Hyeji Kim}, {and} \bibinfo{person}{Nicholas~D. Lane}.} \bibinfo{year}{2020}\natexlab{}.
\newblock \showarticletitle{{{HAPI:} Hardware-Aware Progressive Inference}}.
\newblock \bibinfo{journal}{\emph{CoRR}}  \bibinfo{volume}{abs/2008.03997} (\bibinfo{year}{2020}).
\newblock
\showeprint[arXiv]{2008.03997}
\urldef\tempurl%
\url{https://arxiv.org/abs/2008.03997}
\showURL{%
\tempurl}


\bibitem[Lee et~al\mbox{.}(2019)]%
        {elsa}
\bibfield{author}{\bibinfo{person}{Jaejun Lee}, \bibinfo{person}{Raphael Tang}, {and} \bibinfo{person}{Jimmy Lin}.} \bibinfo{year}{2019}\natexlab{}.
\newblock \showarticletitle{{What Would Elsa Do? Freezing Layers During Transformer Fine-Tuning}}.
\newblock \bibinfo{journal}{\emph{CoRR}}  \bibinfo{volume}{abs/1911.03090} (\bibinfo{year}{2019}).
\newblock
\showeprint[arXiv]{1911.03090}
\urldef\tempurl%
\url{http://arxiv.org/abs/1911.03090}
\showURL{%
\tempurl}


\bibitem[Lee et~al\mbox{.}(2020)]%
        {intermittent_learning}
\bibfield{author}{\bibinfo{person}{Seulki Lee}, \bibinfo{person}{Bashima Islam}, \bibinfo{person}{Yubo Luo}, {and} \bibinfo{person}{Shahriar Nirjon}.} \bibinfo{year}{2020}\natexlab{}.
\newblock \showarticletitle{Intermittent Learning: On-Device Machine Learning on Intermittently Powered System}.
\newblock \bibinfo{journal}{\emph{Proc. ACM Interact. Mob. Wearable Ubiquitous Technol.}} \bibinfo{volume}{3}, \bibinfo{number}{4}, Article \bibinfo{articleno}{141} (\bibinfo{date}{sep} \bibinfo{year}{2020}), \bibinfo{numpages}{30}~pages.
\newblock
\href{https://doi.org/10.1145/3369837}{doi:\nolinkurl{10.1145/3369837}}


\bibitem[Lee et~al\mbox{.}(2023)]%
        {surgical}
\bibfield{author}{\bibinfo{person}{Yoonho Lee}, \bibinfo{person}{Annie~S. Chen}, \bibinfo{person}{Fahim Tajwar}, \bibinfo{person}{Ananya Kumar}, \bibinfo{person}{Huaxiu Yao}, \bibinfo{person}{Percy Liang}, {and} \bibinfo{person}{Chelsea Finn}.} \bibinfo{year}{2023}\natexlab{}.
\newblock \bibinfo{title}{{Surgical Fine-Tuning Improves Adaptation to Distribution Shifts}}.
\newblock
\showeprint[arxiv]{2210.11466}~[cs.LG]


\bibitem[Leontiadis et~al\mbox{.}(2021)]%
        {distill4}
\bibfield{author}{\bibinfo{person}{Ilias Leontiadis}, \bibinfo{person}{Stefanos Laskaridis}, \bibinfo{person}{Stylianos~I. Venieris}, {and} \bibinfo{person}{Nicholas~D. Lane}.} \bibinfo{year}{2021}\natexlab{}.
\newblock \showarticletitle{{It's always personal: Using Early Exits for Efficient On-Device {CNN} Personalisation}}.
\newblock \bibinfo{journal}{\emph{CoRR}}  \bibinfo{volume}{abs/2102.01393} (\bibinfo{year}{2021}).
\newblock
\showeprint[arXiv]{2102.01393}
\urldef\tempurl%
\url{https://arxiv.org/abs/2102.01393}
\showURL{%
\tempurl}


\bibitem[Li et~al\mbox{.}(2021)]%
        {hw_nas_bench}
\bibfield{author}{\bibinfo{person}{Chaojian Li}, \bibinfo{person}{Zhongzhi Yu}, \bibinfo{person}{Yonggan Fu}, \bibinfo{person}{Yongan Zhang}, \bibinfo{person}{Yang Zhao}, \bibinfo{person}{Haoran You}, \bibinfo{person}{Qixuan Yu}, \bibinfo{person}{Yue Wang}, {and} \bibinfo{person}{Yingyan Lin}.} \bibinfo{year}{2021}\natexlab{}.
\newblock \showarticletitle{{HW-NAS-Bench: Hardware-Aware Neural Architecture Search Benchmark}}.
\newblock \bibinfo{journal}{\emph{CoRR}}  \bibinfo{volume}{abs/2103.10584} (\bibinfo{year}{2021}).
\newblock
\showeprint[arXiv]{2103.10584}
\urldef\tempurl%
\url{https://arxiv.org/abs/2103.10584}
\showURL{%
\tempurl}


\bibitem[Li et~al\mbox{.}(2017)]%
        {learning1}
\bibfield{author}{\bibinfo{person}{Da Li}, \bibinfo{person}{Yongxin Yang}, \bibinfo{person}{Yi-Zhe Song}, {and} \bibinfo{person}{Timothy~M. Hospedales}.} \bibinfo{year}{2017}\natexlab{}.
\newblock \bibinfo{title}{{Learning to Generalize: Meta-Learning for Domain Generalization}}.
\newblock
\showeprint[arxiv]{1710.03463}~[cs.LG]


\bibitem[Li et~al\mbox{.}(2020)]%
        {fedavg3}
\bibfield{author}{\bibinfo{person}{Xiang Li}, \bibinfo{person}{Kaixuan Huang}, \bibinfo{person}{Wenhao Yang}, \bibinfo{person}{Shusen Wang}, {and} \bibinfo{person}{Zhihua Zhang}.} \bibinfo{year}{2020}\natexlab{}.
\newblock \bibinfo{title}{{On the Convergence of FedAvg on Non-IID Data}}.
\newblock
\showeprint[arxiv]{1907.02189}~[stat.ML]


\bibitem[Li et~al\mbox{.}(2022)]%
        {pred_exit}
\bibfield{author}{\bibinfo{person}{Xiangjie Li}, \bibinfo{person}{Chenfei Lou}, \bibinfo{person}{Zhengping Zhu}, \bibinfo{person}{Yuchi Chen}, \bibinfo{person}{Yingtao Shen}, \bibinfo{person}{Yehan Ma}, {and} \bibinfo{person}{An Zou}.} \bibinfo{year}{2022}\natexlab{}.
\newblock \bibinfo{title}{{Predictive Exit: Prediction of Fine-Grained Early Exits for Computation- and Energy-Efficient Inference}}.
\newblock
\showeprint[arxiv]{2206.04685}~[cs.LG]


\bibitem[Liberis et~al\mbox{.}(2020)]%
        {unas}
\bibfield{author}{\bibinfo{person}{Edgar Liberis}, \bibinfo{person}{Lukasz Dudziak}, {and} \bibinfo{person}{Nicholas~D. Lane}.} \bibinfo{year}{2020}\natexlab{}.
\newblock \showarticletitle{{{\(\mu\)}NAS: Constrained Neural Architecture Search for Microcontrollers}}.
\newblock \bibinfo{journal}{\emph{CoRR}}  \bibinfo{volume}{abs/2010.14246} (\bibinfo{year}{2020}).
\newblock
\showeprint[arXiv]{2010.14246}
\urldef\tempurl%
\url{https://arxiv.org/abs/2010.14246}
\showURL{%
\tempurl}


\bibitem[Lin et~al\mbox{.}(2020)]%
        {lightweight3}
\bibfield{author}{\bibinfo{person}{Ji Lin}, \bibinfo{person}{Wei{-}Ming Chen}, \bibinfo{person}{Yujun Lin}, \bibinfo{person}{John Cohn}, \bibinfo{person}{Chuang Gan}, {and} \bibinfo{person}{Song Han}.} \bibinfo{year}{2020}\natexlab{}.
\newblock \showarticletitle{{MCUNet: Tiny Deep Learning on IoT Devices}}.
\newblock \bibinfo{journal}{\emph{CoRR}}  \bibinfo{volume}{abs/2007.10319} (\bibinfo{year}{2020}).
\newblock
\showeprint[arXiv]{2007.10319}
\urldef\tempurl%
\url{https://arxiv.org/abs/2007.10319}
\showURL{%
\tempurl}


\bibitem[Lin et~al\mbox{.}(2017)]%
        {lin}
\bibfield{author}{\bibinfo{person}{Ji Lin}, \bibinfo{person}{Yongming Rao}, \bibinfo{person}{Jiwen Lu}, {and} \bibinfo{person}{Jie Zhou}.} \bibinfo{year}{2017}\natexlab{}.
\newblock \showarticletitle{{Runtime Neural Pruning}}. In \bibinfo{booktitle}{\emph{Advances in Neural Information Processing Systems}}, \bibfield{editor}{\bibinfo{person}{I.~Guyon}, \bibinfo{person}{U.~Von Luxburg}, \bibinfo{person}{S.~Bengio}, \bibinfo{person}{H.~Wallach}, \bibinfo{person}{R.~Fergus}, \bibinfo{person}{S.~Vishwanathan}, {and} \bibinfo{person}{R.~Garnett}} (Eds.), Vol.~\bibinfo{volume}{30}. \bibinfo{publisher}{Curran Associates, Inc.}
\newblock
\urldef\tempurl%
\url{https://proceedings.neurips.cc/paper_files/paper/2017/file/a51fb975227d6640e4fe47854476d133-Paper.pdf}
\showURL{%
\tempurl}


\bibitem[Lin et~al\mbox{.}(2024)]%
        {256KB}
\bibfield{author}{\bibinfo{person}{Ji Lin}, \bibinfo{person}{Ligeng Zhu}, \bibinfo{person}{Wei-Ming Chen}, \bibinfo{person}{Wei-Chen Wang}, \bibinfo{person}{Chuang Gan}, {and} \bibinfo{person}{Song Han}.} \bibinfo{year}{2024}\natexlab{}.
\newblock \bibinfo{title}{{On-Device Training Under 256KB Memory}}.
\newblock
\showeprint[arxiv]{2206.15472}~[cs.CV]


\bibitem[Liu et~al\mbox{.}(2019)]%
        {split1}
\bibfield{author}{\bibinfo{person}{Qiang Liu}, \bibinfo{person}{Lemeng Wu}, {and} \bibinfo{person}{Dilin Wang}.} \bibinfo{year}{2019}\natexlab{}.
\newblock \showarticletitle{{Splitting Steepest Descent for Growing Neural Architectures}}.
\newblock \bibinfo{journal}{\emph{CoRR}}  \bibinfo{volume}{abs/1910.02366} (\bibinfo{year}{2019}).
\newblock
\showeprint[arXiv]{1910.02366}
\urldef\tempurl%
\url{http://arxiv.org/abs/1910.02366}
\showURL{%
\tempurl}


\bibitem[Liu et~al\mbox{.}(2023a)]%
        {dbs3}
\bibfield{author}{\bibinfo{person}{Weijie Liu}, \bibinfo{person}{Xiaoxi Zhang}, \bibinfo{person}{Jingpu Duan}, \bibinfo{person}{Carlee Joe-Wong}, \bibinfo{person}{Zhi Zhou}, {and} \bibinfo{person}{Xu Chen}.} \bibinfo{year}{2023}\natexlab{a}.
\newblock \showarticletitle{{AdaCoOpt: Leverage the Interplay of Batch Size and Aggregation Frequency for Federated Learning}}. In \bibinfo{booktitle}{\emph{2023 IEEE/ACM 31st International Symposium on Quality of Service (IWQoS)}}. \bibinfo{pages}{1--10}.
\newblock
\href{https://doi.org/10.1109/IWQoS57198.2023.10188807}{doi:\nolinkurl{10.1109/IWQoS57198.2023.10188807}}


\bibitem[Liu et~al\mbox{.}(2023b)]%
        {dbs2}
\bibfield{author}{\bibinfo{person}{Weijie Liu}, \bibinfo{person}{Xiaoxi Zhang}, \bibinfo{person}{Jingpu Duan}, \bibinfo{person}{Carlee Joe-Wong}, \bibinfo{person}{Zhi Zhou}, {and} \bibinfo{person}{Xu Chen}.} \bibinfo{year}{2023}\natexlab{b}.
\newblock \bibinfo{title}{{DYNAMITE: Dynamic Interplay of Mini-Batch Size and Aggregation Frequency for Federated Learning with Static and Streaming Dataset}}.
\newblock
\showeprint[arxiv]{2310.14906}~[cs.LG]


\bibitem[Liu et~al\mbox{.}(2021)]%
        {autofreeze}
\bibfield{author}{\bibinfo{person}{Yuhan Liu}, \bibinfo{person}{Saurabh Agarwal}, {and} \bibinfo{person}{Shivaram Venkataraman}.} \bibinfo{year}{2021}\natexlab{}.
\newblock \showarticletitle{{AutoFreeze: Automatically Freezing Model Blocks to Accelerate Fine-tuning}}.
\newblock \bibinfo{journal}{\emph{CoRR}}  \bibinfo{volume}{abs/2102.01386} (\bibinfo{year}{2021}).
\newblock
\showeprint[arXiv]{2102.01386}
\urldef\tempurl%
\url{https://arxiv.org/abs/2102.01386}
\showURL{%
\tempurl}


\bibitem[Liu et~al\mbox{.}(2020)]%
        {evolutionary}
\bibfield{author}{\bibinfo{person}{Yuqiao Liu}, \bibinfo{person}{Yanan Sun}, \bibinfo{person}{Bing Xue}, \bibinfo{person}{Mengjie Zhang}, {and} \bibinfo{person}{Gary~G. Yen}.} \bibinfo{year}{2020}\natexlab{}.
\newblock \showarticletitle{{A Survey on Evolutionary Neural Architecture Search}}.
\newblock \bibinfo{journal}{\emph{CoRR}}  \bibinfo{volume}{abs/2008.10937} (\bibinfo{year}{2020}).
\newblock
\showeprint[arXiv]{2008.10937}
\urldef\tempurl%
\url{https://arxiv.org/abs/2008.10937}
\showURL{%
\tempurl}


\bibitem[Luo et~al\mbox{.}(2020)]%
        {cefld}
\bibfield{author}{\bibinfo{person}{Bing Luo}, \bibinfo{person}{Xiang Li}, \bibinfo{person}{Shiqiang Wang}, \bibinfo{person}{Jianwei Huang}, {and} \bibinfo{person}{Leandros Tassiulas}.} \bibinfo{year}{2020}\natexlab{}.
\newblock \showarticletitle{{Cost-Effective Federated Learning Design}}.
\newblock \bibinfo{journal}{\emph{CoRR}}  \bibinfo{volume}{abs/2012.08336} (\bibinfo{year}{2020}).
\newblock
\showeprint[arXiv]{2012.08336}
\urldef\tempurl%
\url{https://arxiv.org/abs/2012.08336}
\showURL{%
\tempurl}


\bibitem[Ma et~al\mbox{.}(2017)]%
        {int4}
\bibfield{author}{\bibinfo{person}{Kaisheng Ma}, \bibinfo{person}{Xueqing Li}, \bibinfo{person}{Jinyang Li}, \bibinfo{person}{Yongpan Liu}, \bibinfo{person}{Yuan Xie}, \bibinfo{person}{Jack Sampson}, \bibinfo{person}{Mahmut~Taylan Kandemir}, {and} \bibinfo{person}{Vijaykrishnan Narayanan}.} \bibinfo{year}{2017}\natexlab{}.
\newblock \showarticletitle{{Incidental Computing on IoT Nonvolatile Processors}}. In \bibinfo{booktitle}{\emph{2017 50th Annual IEEE/ACM International Symposium on Microarchitecture (MICRO)}}. \bibinfo{pages}{204--218}.
\newblock


\bibitem[Ma et~al\mbox{.}(2015)]%
        {int3}
\bibfield{author}{\bibinfo{person}{Kaisheng Ma}, \bibinfo{person}{Yang Zheng}, \bibinfo{person}{Shuangchen Li}, \bibinfo{person}{Karthik Swaminathan}, \bibinfo{person}{Xueqing Li}, \bibinfo{person}{Yongpan Liu}, \bibinfo{person}{Jack Sampson}, \bibinfo{person}{Yuan Xie}, {and} \bibinfo{person}{Vijaykrishnan Narayanan}.} \bibinfo{year}{2015}\natexlab{}.
\newblock \showarticletitle{{Architecture exploration for ambient energy harvesting nonvolatile processors}}. In \bibinfo{booktitle}{\emph{2015 IEEE 21st International Symposium on High Performance Computer Architecture (HPCA)}}. \bibinfo{pages}{526--537}.
\newblock
\href{https://doi.org/10.1109/HPCA.2015.7056060}{doi:\nolinkurl{10.1109/HPCA.2015.7056060}}


\bibitem[Ma et~al\mbox{.}(2018)]%
        {lightweight2}
\bibfield{author}{\bibinfo{person}{Ningning Ma}, \bibinfo{person}{Xiangyu Zhang}, \bibinfo{person}{Hai-Tao Zheng}, {and} \bibinfo{person}{Jian Sun}.} \bibinfo{year}{2018}\natexlab{}.
\newblock \bibinfo{title}{{ShuffleNet V2: Practical Guidelines for Efficient CNN Architecture Design}}.
\newblock
\showeprint[arxiv]{1807.11164}~[cs.CV]


\bibitem[Mahajan et~al\mbox{.}(2020)]%
        {causal2}
\bibfield{author}{\bibinfo{person}{Divyat Mahajan}, \bibinfo{person}{Shruti Tople}, {and} \bibinfo{person}{Amit Sharma}.} \bibinfo{year}{2020}\natexlab{}.
\newblock \showarticletitle{{Domain Generalization using Causal Matching}}.
\newblock \bibinfo{journal}{\emph{CoRR}}  \bibinfo{volume}{abs/2006.07500} (\bibinfo{year}{2020}).
\newblock
\showeprint[arXiv]{2006.07500}
\urldef\tempurl%
\url{https://arxiv.org/abs/2006.07500}
\showURL{%
\tempurl}


\bibitem[Maioli and Mottola(2021)]%
        {alfred}
\bibfield{author}{\bibinfo{person}{Andrea Maioli} {and} \bibinfo{person}{Luca Mottola}.} \bibinfo{year}{2021}\natexlab{}.
\newblock \showarticletitle{{ALFRED: Virtual Memory for Intermittent Computing}}. In \bibinfo{booktitle}{\emph{Proceedings of the 19th ACM Conference on Embedded Networked Sensor Systems}} \emph{(\bibinfo{series}{SenSys ’21})}. \bibinfo{publisher}{ACM}.
\newblock
\href{https://doi.org/10.1145/3485730.3485949}{doi:\nolinkurl{10.1145/3485730.3485949}}


\bibitem[Mansilla et~al\mbox{.}(2021)]%
        {gradient1}
\bibfield{author}{\bibinfo{person}{Lucas Mansilla}, \bibinfo{person}{Rodrigo Echeveste}, \bibinfo{person}{Diego~H. Milone}, {and} \bibinfo{person}{Enzo Ferrante}.} \bibinfo{year}{2021}\natexlab{}.
\newblock \showarticletitle{{Domain Generalization via Gradient Surgery}}.
\newblock \bibinfo{journal}{\emph{CoRR}}  \bibinfo{volume}{abs/2108.01621} (\bibinfo{year}{2021}).
\newblock
\showeprint[arXiv]{2108.01621}
\urldef\tempurl%
\url{https://arxiv.org/abs/2108.01621}
\showURL{%
\tempurl}


\bibitem[Marchisio et~al\mbox{.}(2020)]%
        {nascaps}
\bibfield{author}{\bibinfo{person}{Alberto Marchisio}, \bibinfo{person}{Andrea Massa}, \bibinfo{person}{Vojtech Mrazek}, \bibinfo{person}{Beatrice Bussolino}, \bibinfo{person}{Maurizio Martina}, {and} \bibinfo{person}{Muhammad Shafique}.} \bibinfo{year}{2020}\natexlab{}.
\newblock \showarticletitle{{NASCaps: {A} Framework for Neural Architecture Search to Optimize the Accuracy and Hardware Efficiency of Convolutional Capsule Networks}}.
\newblock \bibinfo{journal}{\emph{CoRR}}  \bibinfo{volume}{abs/2008.08476} (\bibinfo{year}{2020}).
\newblock
\showeprint[arXiv]{2008.08476}
\urldef\tempurl%
\url{https://arxiv.org/abs/2008.08476}
\showURL{%
\tempurl}


\bibitem[McMahan et~al\mbox{.}(2016)]%
        {fedavg1}
\bibfield{author}{\bibinfo{person}{H.~Brendan McMahan}, \bibinfo{person}{Eider Moore}, \bibinfo{person}{Daniel Ramage}, {and} \bibinfo{person}{Blaise~Ag{\"{u}}era y Arcas}.} \bibinfo{year}{2016}\natexlab{}.
\newblock \showarticletitle{{Federated Learning of Deep Networks using Model Averaging}}.
\newblock \bibinfo{journal}{\emph{CoRR}}  \bibinfo{volume}{abs/1602.05629} (\bibinfo{year}{2016}).
\newblock
\showeprint[arXiv]{1602.05629}
\urldef\tempurl%
\url{http://arxiv.org/abs/1602.05629}
\showURL{%
\tempurl}


\bibitem[Meng et~al\mbox{.}(2021)]%
        {pbh}
\bibfield{author}{\bibinfo{person}{Lingchen Meng}, \bibinfo{person}{Hengduo Li}, \bibinfo{person}{Bor{-}Chun Chen}, \bibinfo{person}{Shiyi Lan}, \bibinfo{person}{Zuxuan Wu}, \bibinfo{person}{Yu{-}Gang Jiang}, {and} \bibinfo{person}{Ser{-}Nam Lim}.} \bibinfo{year}{2021}\natexlab{}.
\newblock \showarticletitle{{AdaViT: Adaptive Vision Transformers for Efficient Image Recognition}}.
\newblock \bibinfo{journal}{\emph{CoRR}}  \bibinfo{volume}{abs/2111.15668} (\bibinfo{year}{2021}).
\newblock
\showeprint[arXiv]{2111.15668}
\urldef\tempurl%
\url{https://arxiv.org/abs/2111.15668}
\showURL{%
\tempurl}


\bibitem[Michalkiewicz et~al\mbox{.}(2023)]%
        {gradient3}
\bibfield{author}{\bibinfo{person}{Mateusz Michalkiewicz}, \bibinfo{person}{Masoud Faraki}, \bibinfo{person}{Xiang Yu}, \bibinfo{person}{Manmohan Chandraker}, {and} \bibinfo{person}{Mahsa Baktashmotlagh}.} \bibinfo{year}{2023}\natexlab{}.
\newblock \bibinfo{title}{{Domain Generalization Guided by Gradient Signal to Noise Ratio of Parameters}}.
\newblock
\showeprint[arxiv]{2310.07361}~[cs.CV]


\bibitem[Microsoft(2024)]%
        {edgeml}
\bibfield{author}{\bibinfo{person}{Microsoft}.} \bibinfo{year}{2024}\natexlab{}.
\newblock \bibinfo{title}{{EdgeML}}.
\newblock \bibinfo{howpublished}{\url{https://microsoft.github.io/EdgeML/}}.
\newblock


\bibitem[Moloney(2016)]%
        {oscar}
\bibfield{author}{\bibinfo{person}{David Moloney}.} \bibinfo{year}{2016}\natexlab{}.
\newblock \showarticletitle{{Embedded deep neural networks: “The cost of everything and the value of nothing”}}. In \bibinfo{booktitle}{\emph{2016 IEEE Hot Chips 28 Symposium (HCS)}}. \bibinfo{pages}{1--20}.
\newblock
\href{https://doi.org/10.1109/HOTCHIPS.2016.7936219}{doi:\nolinkurl{10.1109/HOTCHIPS.2016.7936219}}


\bibitem[Montanari et~al\mbox{.}(2020)]%
        {eperceptive}
\bibfield{author}{\bibinfo{person}{Alessandro Montanari}, \bibinfo{person}{Manuja Sharma}, \bibinfo{person}{Dainius Jenkus}, \bibinfo{person}{Mohammed Alloulah}, \bibinfo{person}{Lorena Qendro}, {and} \bibinfo{person}{Fahim Kawsar}.} \bibinfo{year}{2020}\natexlab{}.
\newblock \showarticletitle{{ePerceptive: energy reactive embedded intelligence for batteryless sensors}}. In \bibinfo{booktitle}{\emph{Proceedings of the 18th Conference on Embedded Networked Sensor Systems}} (Virtual Event, Japan) \emph{(\bibinfo{series}{SenSys '20})}. \bibinfo{publisher}{Association for Computing Machinery}, \bibinfo{address}{New York, NY, USA}, \bibinfo{pages}{382–394}.
\newblock
\showISBNx{9781450375900}
\href{https://doi.org/10.1145/3384419.3430782}{doi:\nolinkurl{10.1145/3384419.3430782}}


\bibitem[Odena et~al\mbox{.}(2017)]%
        {odena}
\bibfield{author}{\bibinfo{person}{Augustus Odena}, \bibinfo{person}{Dieterich Lawson}, {and} \bibinfo{person}{Christopher Olah}.} \bibinfo{year}{2017}\natexlab{}.
\newblock \bibinfo{title}{{Changing Model Behavior at Test-Time Using Reinforcement Learning}}.
\newblock
\showeprint[arxiv]{1702.07780}~[stat.ML]


\bibitem[Pagliari et~al\mbox{.}(2018)]%
        {pagliari1}
\bibfield{author}{\bibinfo{person}{Daniele~Jahier Pagliari}, \bibinfo{person}{Enrico Macii}, {and} \bibinfo{person}{Massimo Poncino}.} \bibinfo{year}{2018}\natexlab{}.
\newblock \showarticletitle{{Dynamic Bit-width Reconfiguration for Energy-Efficient Deep Learning Hardware}}. In \bibinfo{booktitle}{\emph{Proceedings of the International Symposium on Low Power Electronics and Design}} (Seattle, WA, USA) \emph{(\bibinfo{series}{ISLPED '18})}. \bibinfo{publisher}{Association for Computing Machinery}, \bibinfo{address}{New York, NY, USA}, Article \bibinfo{articleno}{47}, \bibinfo{numpages}{6}~pages.
\newblock
\showISBNx{9781450357043}
\href{https://doi.org/10.1145/3218603.3218611}{doi:\nolinkurl{10.1145/3218603.3218611}}


\bibitem[Panda et~al\mbox{.}(2015a)]%
        {panda}
\bibfield{author}{\bibinfo{person}{Priyadarshini Panda}, \bibinfo{person}{Abhronil Sengupta}, {and} \bibinfo{person}{Kaushik Roy}.} \bibinfo{year}{2015}\natexlab{a}.
\newblock \showarticletitle{{Conditional Deep Learning for Energy-Efficient and Enhanced Pattern Recognition}}.
\newblock \bibinfo{journal}{\emph{CoRR}}  \bibinfo{volume}{abs/1509.08971} (\bibinfo{year}{2015}).
\newblock
\showeprint[arXiv]{1509.08971}
\urldef\tempurl%
\url{http://arxiv.org/abs/1509.08971}
\showURL{%
\tempurl}


\bibitem[Panda et~al\mbox{.}(2015b)]%
        {cdl}
\bibfield{author}{\bibinfo{person}{Priyadarshini Panda}, \bibinfo{person}{Abhronil Sengupta}, {and} \bibinfo{person}{Kaushik Roy}.} \bibinfo{year}{2015}\natexlab{b}.
\newblock \showarticletitle{Conditional Deep Learning for Energy-Efficient and Enhanced Pattern Recognition}.
\newblock \bibinfo{journal}{\emph{CoRR}}  \bibinfo{volume}{abs/1509.08971} (\bibinfo{year}{2015}).
\newblock
\showeprint[arXiv]{1509.08971}
\urldef\tempurl%
\url{http://arxiv.org/abs/1509.08971}
\showURL{%
\tempurl}


\bibitem[Pang et~al\mbox{.}(2019)]%
        {learning4}
\bibfield{author}{\bibinfo{person}{Yanwei Pang}, \bibinfo{person}{Haoran Wang}, \bibinfo{person}{Yunlong Yu}, {and} \bibinfo{person}{Zhong Ji}.} \bibinfo{year}{2019}\natexlab{}.
\newblock \showarticletitle{{A decadal survey of zero-shot image classification}}.
\newblock \bibinfo{journal}{\emph{SCIENTIA SINICA Informationis}} (\bibinfo{year}{2019}).
\newblock
\urldef\tempurl%
\url{https://api.semanticscholar.org/CorpusID:208104980}
\showURL{%
\tempurl}


\bibitem[Park et~al\mbox{.}(2015)]%
        {big_little}
\bibfield{author}{\bibinfo{person}{Eunhyeok Park}, \bibinfo{person}{Dongyoung Kim}, \bibinfo{person}{Soobeom Kim}, \bibinfo{person}{Yong-Deok Kim}, \bibinfo{person}{Gunhee Kim}, \bibinfo{person}{Sungroh Yoon}, {and} \bibinfo{person}{Sungjoo Yoo}.} \bibinfo{year}{2015}\natexlab{}.
\newblock \showarticletitle{{Big/little deep neural network for ultra low power inference}}. In \bibinfo{booktitle}{\emph{2015 International Conference on Hardware/Software Codesign and System Synthesis (CODES+ISSS)}}. \bibinfo{pages}{124--132}.
\newblock
\href{https://doi.org/10.1109/CODESISSS.2015.7331375}{doi:\nolinkurl{10.1109/CODESISSS.2015.7331375}}


\bibitem[Profentzas et~al\mbox{.}(2023)]%
        {minilearn}
\bibfield{author}{\bibinfo{person}{Christos Profentzas}, \bibinfo{person}{Magnus Almgren}, {and} \bibinfo{person}{Olaf Landsiedel}.} \bibinfo{year}{2023}\natexlab{}.
\newblock \showarticletitle{{MiniLearn: On-Device Learning for Low-Power IoT Devices}}. In \bibinfo{booktitle}{\emph{Proceedings of the 2022 International Conference on Embedded Wireless Systems and Networks}} (, Linz, Austria,) \emph{(\bibinfo{series}{EWSN '22})}. \bibinfo{publisher}{Association for Computing Machinery}, \bibinfo{address}{New York, NY, USA}, \bibinfo{pages}{1–11}.
\newblock


\bibitem[Quélennec et~al\mbox{.}(2023)]%
        {neurons_under_budget}
\bibfield{author}{\bibinfo{person}{Aël Quélennec}, \bibinfo{person}{Enzo Tartaglione}, \bibinfo{person}{Pavlo Mozharovskyi}, {and} \bibinfo{person}{Van-Tam Nguyen}.} \bibinfo{year}{2023}\natexlab{}.
\newblock \bibinfo{title}{{Towards On-device Learning on the Edge: Ways to Select Neurons to Update under a Budget Constraint}}.
\newblock
\showeprint[arxiv]{2312.05282}~[cs.LG]


\bibitem[Rajasegaran et~al\mbox{.}(2019)]%
        {deepcaps}
\bibfield{author}{\bibinfo{person}{Jathushan Rajasegaran}, \bibinfo{person}{Vinoj Jayasundara}, \bibinfo{person}{Sandaru Jayasekara}, \bibinfo{person}{Hirunima Jayasekara}, \bibinfo{person}{Suranga Seneviratne}, {and} \bibinfo{person}{Ranga Rodrigo}.} \bibinfo{year}{2019}\natexlab{}.
\newblock \bibinfo{title}{{DeepCaps: Going Deeper with Capsule Networks}}.
\newblock
\showeprint[arxiv]{1904.09546}~[cs.CV]


\bibitem[Ramasesh et~al\mbox{.}(2020)]%
        {catastrophy}
\bibfield{author}{\bibinfo{person}{Vinay~V. Ramasesh}, \bibinfo{person}{Ethan Dyer}, {and} \bibinfo{person}{Maithra Raghu}.} \bibinfo{year}{2020}\natexlab{}.
\newblock \showarticletitle{{Anatomy of Catastrophic Forgetting: Hidden Representations and Task Semantics}}.
\newblock \bibinfo{journal}{\emph{CoRR}}  \bibinfo{volume}{abs/2007.07400} (\bibinfo{year}{2020}).
\newblock
\showeprint[arXiv]{2007.07400}
\urldef\tempurl%
\url{https://arxiv.org/abs/2007.07400}
\showURL{%
\tempurl}


\bibitem[Rastegari et~al\mbox{.}(2016)]%
        {comp3}
\bibfield{author}{\bibinfo{person}{Mohammad Rastegari}, \bibinfo{person}{Vicente Ordonez}, \bibinfo{person}{Joseph Redmon}, {and} \bibinfo{person}{Ali Farhadi}.} \bibinfo{year}{2016}\natexlab{}.
\newblock \bibinfo{title}{{XNOR-Net: ImageNet Classification Using Binary Convolutional Neural Networks}}.
\newblock
\showeprint[arxiv]{1603.05279}~[cs.CV]


\bibitem[Ren et~al\mbox{.}(2020)]%
        {nas_survey}
\bibfield{author}{\bibinfo{person}{Pengzhen Ren}, \bibinfo{person}{Yun Xiao}, \bibinfo{person}{Xiaojun Chang}, \bibinfo{person}{Po{-}Yao Huang}, \bibinfo{person}{Zhihui Li}, \bibinfo{person}{Xiaojiang Chen}, {and} \bibinfo{person}{Xin Wang}.} \bibinfo{year}{2020}\natexlab{}.
\newblock \showarticletitle{{A Comprehensive Survey of Neural Architecture Search: Challenges and Solutions}}.
\newblock \bibinfo{journal}{\emph{CoRR}}  \bibinfo{volume}{abs/2006.02903} (\bibinfo{year}{2020}).
\newblock
\showeprint[arXiv]{2006.02903}
\urldef\tempurl%
\url{https://arxiv.org/abs/2006.02903}
\showURL{%
\tempurl}


\bibitem[Rodrigues et~al\mbox{.}(2018)]%
        {synergy}
\bibfield{author}{\bibinfo{person}{Crefeda~Faviola Rodrigues}, \bibinfo{person}{Graham Riley}, {and} \bibinfo{person}{Mikel Lujan}.} \bibinfo{year}{2018}\natexlab{}.
\newblock \bibinfo{title}{{Fine-Grained Energy and Performance Profiling framework for Deep Convolutional Neural Networks}}.
\newblock
\showeprint[arxiv]{1803.11151}~[cs.PF]


\bibitem[Rouhani et~al\mbox{.}(2016)]%
        {delight}
\bibfield{author}{\bibinfo{person}{Bita~Darvish Rouhani}, \bibinfo{person}{Azalia Mirhoseini}, {and} \bibinfo{person}{Farinaz Koushanfar}.} \bibinfo{year}{2016}\natexlab{}.
\newblock \showarticletitle{DeLight: Adding Energy Dimension To Deep Neural Networks}. In \bibinfo{booktitle}{\emph{Proceedings of the 2016 International Symposium on Low Power Electronics and Design}} (San Francisco Airport, CA, USA) \emph{(\bibinfo{series}{ISLPED '16})}. \bibinfo{publisher}{Association for Computing Machinery}, \bibinfo{address}{New York, NY, USA}, \bibinfo{pages}{112–117}.
\newblock
\showISBNx{9781450341851}
\href{https://doi.org/10.1145/2934583.2934599}{doi:\nolinkurl{10.1145/2934583.2934599}}


\bibitem[Royer and Lampert(2020)]%
        {flextuning}
\bibfield{author}{\bibinfo{person}{Amelie Royer} {and} \bibinfo{person}{Christoph~H. Lampert}.} \bibinfo{year}{2020}\natexlab{}.
\newblock \showarticletitle{{A Flexible Selection Scheme for Minimum-Effort Transfer Learning}}.
\newblock \bibinfo{journal}{\emph{CoRR}}  \bibinfo{volume}{abs/2008.11995} (\bibinfo{year}{2020}).
\newblock
\showeprint[arXiv]{2008.11995}
\urldef\tempurl%
\url{https://arxiv.org/abs/2008.11995}
\showURL{%
\tempurl}


\bibitem[Sabour et~al\mbox{.}(2017)]%
        {capsnet}
\bibfield{author}{\bibinfo{person}{Sara Sabour}, \bibinfo{person}{Nicholas Frosst}, {and} \bibinfo{person}{Geoffrey~E. Hinton}.} \bibinfo{year}{2017}\natexlab{}.
\newblock \showarticletitle{{Dynamic Routing Between Capsules}}.
\newblock \bibinfo{journal}{\emph{CoRR}}  \bibinfo{volume}{abs/1710.09829} (\bibinfo{year}{2017}).
\newblock
\showeprint[arXiv]{1710.09829}
\urldef\tempurl%
\url{http://arxiv.org/abs/1710.09829}
\showURL{%
\tempurl}


\bibitem[Sabovic et~al\mbox{.}(2023)]%
        {sabovic}
\bibfield{author}{\bibinfo{person}{Adnan Sabovic}, \bibinfo{person}{Michiel Aernouts}, \bibinfo{person}{Dragan Subotic}, \bibinfo{person}{Jaron Fontaine}, \bibinfo{person}{Eli {De Poorter}}, {and} \bibinfo{person}{Jeroen Famaey}.} \bibinfo{year}{2023}\natexlab{}.
\newblock \showarticletitle{Towards energy-aware tinyML on battery-less IoT devices}.
\newblock \bibinfo{journal}{\emph{Internet of Things}}  \bibinfo{volume}{22} (\bibinfo{year}{2023}), \bibinfo{pages}{100736}.
\newblock
\showISSN{2542-6605}
\href{https://doi.org/10.1016/j.iot.2023.100736}{doi:\nolinkurl{10.1016/j.iot.2023.100736}}


\bibitem[Sabovic et~al\mbox{.}(2022)]%
        {sched1}
\bibfield{author}{\bibinfo{person}{Adnan Sabovic}, \bibinfo{person}{Ashish~Kumar Sultania}, \bibinfo{person}{Carmen Delgado}, \bibinfo{person}{Lander~De Roeck}, {and} \bibinfo{person}{Jeroen Famaey}.} \bibinfo{year}{2022}\natexlab{}.
\newblock \showarticletitle{{An Energy-Aware Task Scheduler for Energy-Harvesting Batteryless IoT Devices}}.
\newblock \bibinfo{journal}{\emph{IEEE Internet of Things Journal}} \bibinfo{volume}{9}, \bibinfo{number}{22} (\bibinfo{year}{2022}), \bibinfo{pages}{23097--23114}.
\newblock
\href{https://doi.org/10.1109/JIOT.2022.3185321}{doi:\nolinkurl{10.1109/JIOT.2022.3185321}}


\bibitem[Salman et~al\mbox{.}(2020)]%
        {salman}
\bibfield{author}{\bibinfo{person}{Hadi Salman}, \bibinfo{person}{Andrew Ilyas}, \bibinfo{person}{Logan Engstrom}, \bibinfo{person}{Ashish Kapoor}, {and} \bibinfo{person}{Aleksander Madry}.} \bibinfo{year}{2020}\natexlab{}.
\newblock \showarticletitle{{Do Adversarially Robust ImageNet Models Transfer Better?}}
\newblock \bibinfo{journal}{\emph{CoRR}}  \bibinfo{volume}{abs/2007.08489} (\bibinfo{year}{2020}).
\newblock
\showeprint[arXiv]{2007.08489}
\urldef\tempurl%
\url{https://arxiv.org/abs/2007.08489}
\showURL{%
\tempurl}


\bibitem[Samikwa et~al\mbox{.}(2022)]%
        {eeoc}
\bibfield{author}{\bibinfo{person}{Eric Samikwa}, \bibinfo{person}{Antonio Di~Maio}, {and} \bibinfo{person}{Torsten Braun}.} \bibinfo{year}{2022}\natexlab{}.
\newblock \showarticletitle{Adaptive Early Exit of Computation for Energy-Efficient and Low-Latency Machine Learning over IoT Networks}. In \bibinfo{booktitle}{\emph{2022 IEEE 19th Annual Consumer Communications \& Networking Conference (CCNC)}}. \bibinfo{pages}{200--206}.
\newblock
\href{https://doi.org/10.1109/CCNC49033.2022.9700550}{doi:\nolinkurl{10.1109/CCNC49033.2022.9700550}}


\bibitem[Sandler et~al\mbox{.}(2019a)]%
        {lightweight1}
\bibfield{author}{\bibinfo{person}{Mark Sandler}, \bibinfo{person}{Andrew Howard}, \bibinfo{person}{Menglong Zhu}, \bibinfo{person}{Andrey Zhmoginov}, {and} \bibinfo{person}{Liang-Chieh Chen}.} \bibinfo{year}{2019}\natexlab{a}.
\newblock \bibinfo{title}{{MobileNetV2: Inverted Residuals and Linear Bottlenecks}}.
\newblock
\showeprint[arxiv]{1801.04381}~[cs.CV]


\bibitem[Sandler et~al\mbox{.}(2019b)]%
        {mobilenetv2}
\bibfield{author}{\bibinfo{person}{Mark Sandler}, \bibinfo{person}{Andrew Howard}, \bibinfo{person}{Menglong Zhu}, \bibinfo{person}{Andrey Zhmoginov}, {and} \bibinfo{person}{Liang-Chieh Chen}.} \bibinfo{year}{2019}\natexlab{b}.
\newblock \bibinfo{title}{{MobileNetV2: Inverted Residuals and Linear Bottlenecks}}.
\newblock
\showeprint[arxiv]{1801.04381}~[cs.CV]


\bibitem[Sattler et~al\mbox{.}(2019)]%
        {fedavg2}
\bibfield{author}{\bibinfo{person}{Felix Sattler}, \bibinfo{person}{Simon Wiedemann}, \bibinfo{person}{Klaus{-}Robert M{\"{u}}ller}, {and} \bibinfo{person}{Wojciech Samek}.} \bibinfo{year}{2019}\natexlab{}.
\newblock \showarticletitle{{Robust and Communication-Efficient Federated Learning from Non-IID Data}}.
\newblock \bibinfo{journal}{\emph{CoRR}}  \bibinfo{volume}{abs/1903.02891} (\bibinfo{year}{2019}).
\newblock
\showeprint[arXiv]{1903.02891}
\urldef\tempurl%
\url{http://arxiv.org/abs/1903.02891}
\showURL{%
\tempurl}


\bibitem[Shazeer et~al\mbox{.}(2018)]%
        {hydranets}
\bibfield{author}{\bibinfo{person}{Noam Shazeer}, \bibinfo{person}{Kayvon Fatahalian}, \bibinfo{person}{William~R. Mark}, {and} \bibinfo{person}{Ravi~Teja Mullapudi}.} \bibinfo{year}{2018}\natexlab{}.
\newblock \showarticletitle{{HydraNets: Specialized Dynamic Architectures for Efficient Inference}}. In \bibinfo{booktitle}{\emph{2018 IEEE/CVF Conference on Computer Vision and Pattern Recognition}}. \bibinfo{pages}{8080--8089}.
\newblock
\href{https://doi.org/10.1109/CVPR.2018.00843}{doi:\nolinkurl{10.1109/CVPR.2018.00843}}


\bibitem[Shazeer et~al\mbox{.}(2017)]%
        {shazeer}
\bibfield{author}{\bibinfo{person}{Noam Shazeer}, \bibinfo{person}{Azalia Mirhoseini}, \bibinfo{person}{Krzysztof Maziarz}, \bibinfo{person}{Andy Davis}, \bibinfo{person}{Quoc~V. Le}, \bibinfo{person}{Geoffrey~E. Hinton}, {and} \bibinfo{person}{Jeff Dean}.} \bibinfo{year}{2017}\natexlab{}.
\newblock \showarticletitle{{Outrageously Large Neural Networks: The Sparsely-Gated Mixture-of-Experts Layer}}.
\newblock \bibinfo{journal}{\emph{CoRR}}  \bibinfo{volume}{abs/1701.06538} (\bibinfo{year}{2017}).
\newblock
\showeprint[arXiv]{1701.06538}
\urldef\tempurl%
\url{http://arxiv.org/abs/1701.06538}
\showURL{%
\tempurl}


\bibitem[Sheth and Liu(2023)]%
        {causal3}
\bibfield{author}{\bibinfo{person}{Paras Sheth} {and} \bibinfo{person}{Huan Liu}.} \bibinfo{year}{2023}\natexlab{}.
\newblock \bibinfo{booktitle}{\emph{Causal Domain Generalization}}.
\newblock \bibinfo{publisher}{Springer International Publishing}, \bibinfo{address}{Cham}, \bibinfo{pages}{161--185}.
\newblock
\showISBNx{978-3-031-35051-1}
\href{https://doi.org/10.1007/978-3-031-35051-1_8}{doi:\nolinkurl{10.1007/978-3-031-35051-1_8}}


\bibitem[Sheth et~al\mbox{.}(2022)]%
        {causal1}
\bibfield{author}{\bibinfo{person}{Paras Sheth}, \bibinfo{person}{Raha Moraffah}, \bibinfo{person}{K.~Selçuk Candan}, \bibinfo{person}{Adrienne Raglin}, {and} \bibinfo{person}{Huan Liu}.} \bibinfo{year}{2022}\natexlab{}.
\newblock \bibinfo{title}{{Domain Generalization -- A Causal Perspective}}.
\newblock
\showeprint[arxiv]{2209.15177}~[cs.LG]


\bibitem[Shi et~al\mbox{.}(2022a)]%
        {5g}
\bibfield{author}{\bibinfo{person}{Dian Shi}, \bibinfo{person}{Liang Li}, \bibinfo{person}{Rui Chen}, \bibinfo{person}{Pavana Prakash}, \bibinfo{person}{Miao Pan}, {and} \bibinfo{person}{Yuguang Fang}.} \bibinfo{year}{2022}\natexlab{a}.
\newblock \showarticletitle{{Toward Energy-Efficient Federated Learning Over 5G+ Mobile Devices}}.
\newblock \bibinfo{journal}{\emph{IEEE Wireless Communications}} \bibinfo{volume}{29}, \bibinfo{number}{5} (\bibinfo{year}{2022}), \bibinfo{pages}{44--51}.
\newblock
\href{https://doi.org/10.1109/MWC.003.2100028}{doi:\nolinkurl{10.1109/MWC.003.2100028}}


\bibitem[Shi et~al\mbox{.}(2022b)]%
        {dbs1}
\bibfield{author}{\bibinfo{person}{Dian Shi}, \bibinfo{person}{Liang Li}, \bibinfo{person}{Maoqiang Wu}, \bibinfo{person}{Minglei Shu}, \bibinfo{person}{Rong Yu}, \bibinfo{person}{Miao Pan}, {and} \bibinfo{person}{Zhu Han}.} \bibinfo{year}{2022}\natexlab{b}.
\newblock \showarticletitle{{To Talk or to Work: Dynamic Batch Sizes Assisted Time Efficient Federated Learning Over Future Mobile Edge Devices}}.
\newblock \bibinfo{journal}{\emph{IEEE Transactions on Wireless Communications}} \bibinfo{volume}{21}, \bibinfo{number}{12} (\bibinfo{year}{2022}), \bibinfo{pages}{11038--11050}.
\newblock
\href{https://doi.org/10.1109/TWC.2022.3189320}{doi:\nolinkurl{10.1109/TWC.2022.3189320}}


\bibitem[Shi et~al\mbox{.}(2021)]%
        {gradient2}
\bibfield{author}{\bibinfo{person}{Yuge Shi}, \bibinfo{person}{Jeffrey Seely}, \bibinfo{person}{Philip H.~S. Torr}, \bibinfo{person}{N. Siddharth}, \bibinfo{person}{Awni~Y. Hannun}, \bibinfo{person}{Nicolas Usunier}, {and} \bibinfo{person}{Gabriel Synnaeve}.} \bibinfo{year}{2021}\natexlab{}.
\newblock \showarticletitle{{Gradient Matching for Domain Generalization}}.
\newblock \bibinfo{journal}{\emph{CoRR}}  \bibinfo{volume}{abs/2104.09937} (\bibinfo{year}{2021}).
\newblock
\showeprint[arXiv]{2104.09937}
\urldef\tempurl%
\url{https://arxiv.org/abs/2104.09937}
\showURL{%
\tempurl}


\bibitem[Spadaro et~al\mbox{.}(2023)]%
        {shannon_strikes_again}
\bibfield{author}{\bibinfo{person}{G. Spadaro}, \bibinfo{person}{R. Renzulli}, \bibinfo{person}{A. Bragagnolo}, \bibinfo{person}{J.~H. Giraldo}, \bibinfo{person}{A. Fiandrotti}, \bibinfo{person}{M. Grangetto}, {and} \bibinfo{person}{E. Tartaglione}.} \bibinfo{year}{2023}\natexlab{}.
\newblock \showarticletitle{{Shannon Strikes Again! Entropy-based Pruning in Deep Neural Networks for Transfer Learning under Extreme Memory and Computation Budgets}}. In \bibinfo{booktitle}{\emph{2023 IEEE/CVF International Conference on Computer Vision Workshops (ICCVW)}}. \bibinfo{publisher}{IEEE Computer Society}, \bibinfo{address}{Los Alamitos, CA, USA}, \bibinfo{pages}{1510--1514}.
\newblock
\href{https://doi.org/10.1109/ICCVW60793.2023.00165}{doi:\nolinkurl{10.1109/ICCVW60793.2023.00165}}


\bibitem[STMicroelectronics(2024)]%
        {stm32cubemx}
\bibfield{author}{\bibinfo{person}{STMicroelectronics}.} \bibinfo{year}{2024}\natexlab{}.
\newblock \bibinfo{title}{{STM32CubeMX}}.
\newblock \bibinfo{howpublished}{\url{https://www.st.com/en/development-tools/stm32cubemx.html}}.
\newblock


\bibitem[Sultania and Famaey(2022)]%
        {sched2}
\bibfield{author}{\bibinfo{person}{Ashish~Kumar Sultania} {and} \bibinfo{person}{Jeroen Famaey}.} \bibinfo{year}{2022}\natexlab{}.
\newblock \showarticletitle{{Batteryless Bluetooth Low Energy Prototype With Energy-Aware Bidirectional Communication Powered by Ambient Light}}.
\newblock \bibinfo{journal}{\emph{IEEE Sensors Journal}} \bibinfo{volume}{22}, \bibinfo{number}{7} (\bibinfo{year}{2022}), \bibinfo{pages}{6685--6697}.
\newblock
\href{https://doi.org/10.1109/JSEN.2022.3153097}{doi:\nolinkurl{10.1109/JSEN.2022.3153097}}


\bibitem[Sun et~al\mbox{.}(2019)]%
        {denizzz}
\bibfield{author}{\bibinfo{person}{Yuxuan Sun}, \bibinfo{person}{Sheng Zhou}, {and} \bibinfo{person}{Deniz Gündüz}.} \bibinfo{year}{2019}\natexlab{}.
\newblock \bibinfo{title}{{Energy-Aware Analog Aggregation for Federated Learning with Redundant Data}}.
\newblock
\showeprint[arxiv]{1911.00188}~[cs.IT]


\bibitem[Szegedy et~al\mbox{.}(2014)]%
        {googlenet}
\bibfield{author}{\bibinfo{person}{Christian Szegedy}, \bibinfo{person}{Wei Liu}, \bibinfo{person}{Yangqing Jia}, \bibinfo{person}{Pierre Sermanet}, \bibinfo{person}{Scott~E. Reed}, \bibinfo{person}{Dragomir Anguelov}, \bibinfo{person}{Dumitru Erhan}, \bibinfo{person}{Vincent Vanhoucke}, {and} \bibinfo{person}{Andrew Rabinovich}.} \bibinfo{year}{2014}\natexlab{}.
\newblock \showarticletitle{{Going Deeper with Convolutions}}.
\newblock \bibinfo{journal}{\emph{CoRR}}  \bibinfo{volume}{abs/1409.4842} (\bibinfo{year}{2014}).
\newblock
\showeprint[arXiv]{1409.4842}
\urldef\tempurl%
\url{http://arxiv.org/abs/1409.4842}
\showURL{%
\tempurl}


\bibitem[Tan et~al\mbox{.}(2018)]%
        {mnasnet}
\bibfield{author}{\bibinfo{person}{Mingxing Tan}, \bibinfo{person}{Bo Chen}, \bibinfo{person}{Ruoming Pang}, \bibinfo{person}{Vijay Vasudevan}, {and} \bibinfo{person}{Quoc~V. Le}.} \bibinfo{year}{2018}\natexlab{}.
\newblock \showarticletitle{{MnasNet: Platform-Aware Neural Architecture Search for Mobile}}.
\newblock \bibinfo{journal}{\emph{CoRR}}  \bibinfo{volume}{abs/1807.11626} (\bibinfo{year}{2018}).
\newblock
\showeprint[arXiv]{1807.11626}
\urldef\tempurl%
\url{http://arxiv.org/abs/1807.11626}
\showURL{%
\tempurl}


\bibitem[Tan and Le(2019)]%
        {efficientnet}
\bibfield{author}{\bibinfo{person}{Mingxing Tan} {and} \bibinfo{person}{Quoc~V. Le}.} \bibinfo{year}{2019}\natexlab{}.
\newblock \showarticletitle{{EfficientNet: Rethinking Model Scaling for Convolutional Neural Networks}}.
\newblock \bibinfo{journal}{\emph{CoRR}}  \bibinfo{volume}{abs/1905.11946} (\bibinfo{year}{2019}).
\newblock
\showeprint[arXiv]{1905.11946}
\urldef\tempurl%
\url{http://arxiv.org/abs/1905.11946}
\showURL{%
\tempurl}


\bibitem[Tann et~al\mbox{.}(2016)]%
        {tann}
\bibfield{author}{\bibinfo{person}{Hokchhay Tann}, \bibinfo{person}{Soheil Hashemi}, \bibinfo{person}{R.~Iris Bahar}, {and} \bibinfo{person}{Sherief Reda}.} \bibinfo{year}{2016}\natexlab{}.
\newblock \showarticletitle{{Runtime Configurable Deep Neural Networks for Energy-Accuracy Trade-off}}.
\newblock \bibinfo{journal}{\emph{CoRR}}  \bibinfo{volume}{abs/1607.05418} (\bibinfo{year}{2016}).
\newblock
\showeprint[arXiv]{1607.05418}
\urldef\tempurl%
\url{http://arxiv.org/abs/1607.05418}
\showURL{%
\tempurl}


\bibitem[Teerapittayanon et~al\mbox{.}(2017a)]%
        {branchy}
\bibfield{author}{\bibinfo{person}{Surat Teerapittayanon}, \bibinfo{person}{Bradley McDanel}, {and} \bibinfo{person}{H.~T. Kung}.} \bibinfo{year}{2017}\natexlab{a}.
\newblock \showarticletitle{{BranchyNet: Fast Inference via Early Exiting from Deep Neural Networks}}.
\newblock \bibinfo{journal}{\emph{CoRR}}  \bibinfo{volume}{abs/1709.01686} (\bibinfo{year}{2017}).
\newblock
\showeprint[arXiv]{1709.01686}
\urldef\tempurl%
\url{http://arxiv.org/abs/1709.01686}
\showURL{%
\tempurl}


\bibitem[Teerapittayanon et~al\mbox{.}(2017b)]%
        {ee1}
\bibfield{author}{\bibinfo{person}{Surat Teerapittayanon}, \bibinfo{person}{Bradley McDanel}, {and} \bibinfo{person}{H.~T. Kung}.} \bibinfo{year}{2017}\natexlab{b}.
\newblock \showarticletitle{{BranchyNet: Fast Inference via Early Exiting from Deep Neural Networks}}.
\newblock \bibinfo{journal}{\emph{CoRR}}  \bibinfo{volume}{abs/1709.01686} (\bibinfo{year}{2017}).
\newblock
\showeprint[arXiv]{1709.01686}
\urldef\tempurl%
\url{http://arxiv.org/abs/1709.01686}
\showURL{%
\tempurl}


\bibitem[Tekin et~al\mbox{.}(2024)]%
        {energyreview}
\bibfield{author}{\bibinfo{person}{Nazli Tekin}, \bibinfo{person}{Ahmet Aris}, \bibinfo{person}{Abbas Acar}, \bibinfo{person}{Selcuk Uluagac}, {and} \bibinfo{person}{Vehbi~Cagri Gungor}.} \bibinfo{year}{2024}\natexlab{}.
\newblock \showarticletitle{{A review of on-device machine learning for IoT: An energy perspective}}.
\newblock \bibinfo{journal}{\emph{Ad Hoc Networks}}  \bibinfo{volume}{153} (\bibinfo{year}{2024}), \bibinfo{pages}{103348}.
\newblock
\showISSN{1570-8705}
\href{https://doi.org/10.1016/j.adhoc.2023.103348}{doi:\nolinkurl{10.1016/j.adhoc.2023.103348}}


\bibitem[Tran et~al\mbox{.}(2019)]%
        {nguyen}
\bibfield{author}{\bibinfo{person}{Nguyen~H. Tran}, \bibinfo{person}{Wei Bao}, \bibinfo{person}{Albert Zomaya}, \bibinfo{person}{Minh N.~H. Nguyen}, {and} \bibinfo{person}{Choong~Seon Hong}.} \bibinfo{year}{2019}\natexlab{}.
\newblock \showarticletitle{{Federated Learning over Wireless Networks: Optimization Model Design and Analysis}}. In \bibinfo{booktitle}{\emph{IEEE INFOCOM 2019 - IEEE Conference on Computer Communications}}. \bibinfo{pages}{1387--1395}.
\newblock
\href{https://doi.org/10.1109/INFOCOM.2019.8737464}{doi:\nolinkurl{10.1109/INFOCOM.2019.8737464}}


\bibitem[Truong et~al\mbox{.}(2018)]%
        {capband}
\bibfield{author}{\bibinfo{person}{Hoang Truong}, \bibinfo{person}{Shuo Zhang}, \bibinfo{person}{Ufuk Muncuk}, \bibinfo{person}{Phuc Nguyen}, \bibinfo{person}{Nam Bui}, \bibinfo{person}{Anh Nguyen}, \bibinfo{person}{Qin Lv}, \bibinfo{person}{Kaushik Chowdhury}, \bibinfo{person}{Thang Dinh}, {and} \bibinfo{person}{Tam Vu}.} \bibinfo{year}{2018}\natexlab{}.
\newblock \showarticletitle{{CapBand: Battery-free Successive Capacitance Sensing Wristband for Hand Gesture Recognition}}. In \bibinfo{booktitle}{\emph{Proceedings of the 16th ACM Conference on Embedded Networked Sensor Systems}} (Shenzhen, China) \emph{(\bibinfo{series}{SenSys '18})}. \bibinfo{publisher}{Association for Computing Machinery}, \bibinfo{address}{New York, NY, USA}, \bibinfo{pages}{54–67}.
\newblock
\showISBNx{9781450359528}
\href{https://doi.org/10.1145/3274783.3274854}{doi:\nolinkurl{10.1145/3274783.3274854}}


\bibitem[Tschand et~al\mbox{.}(2025)]%
        {mlperfpower}
\bibfield{author}{\bibinfo{person}{Arya Tschand}, \bibinfo{person}{Arun Tejusve~Raghunath Rajan}, \bibinfo{person}{Sachin Idgunji}, \bibinfo{person}{Anirban Ghosh}, \bibinfo{person}{Jeremy Holleman}, \bibinfo{person}{Csaba Kiraly}, \bibinfo{person}{Pawan Ambalkar}, \bibinfo{person}{Ritika Borkar}, \bibinfo{person}{Ramesh Chukka}, \bibinfo{person}{Trevor Cockrell}, \bibinfo{person}{Oliver Curtis}, \bibinfo{person}{Grigori Fursin}, \bibinfo{person}{Miro Hodak}, \bibinfo{person}{Hiwot Kassa}, \bibinfo{person}{Anton Lokhmotov}, \bibinfo{person}{Dejan Miskovic}, \bibinfo{person}{Yuechao Pan}, \bibinfo{person}{Manu~Prasad Manmathan}, \bibinfo{person}{Liz Raymond}, \bibinfo{person}{Tom~St. John}, \bibinfo{person}{Arjun Suresh}, \bibinfo{person}{Rowan Taubitz}, \bibinfo{person}{Sean Zhan}, \bibinfo{person}{Scott Wasson}, \bibinfo{person}{David Kanter}, {and} \bibinfo{person}{Vijay~Janapa Reddi}.} \bibinfo{year}{2025}\natexlab{}.
\newblock \bibinfo{title}{{MLPerf Power: Benchmarking the Energy Efficiency of Machine Learning Systems from Microwatts to Megawatts for Sustainable AI}}.
\newblock
\showeprint[arxiv]{2410.12032}~[cs.AR]
\urldef\tempurl%
\url{https://arxiv.org/abs/2410.12032}
\showURL{%
\tempurl}


\bibitem[Tu et~al\mbox{.}(2023)]%
        {deepen}
\bibfield{author}{\bibinfo{person}{X. Tu}, \bibinfo{person}{A. Mallik}, \bibinfo{person}{D. Chen}, \bibinfo{person}{K. Han}, \bibinfo{person}{O. Altintas}, \bibinfo{person}{H. Wang}, {and} \bibinfo{person}{J. Xie}.} \bibinfo{year}{2023}\natexlab{}.
\newblock \showarticletitle{Unveiling Energy Efficiency in Deep Learning: Measurement, Prediction, and Scoring Across Edge Devices}. In \bibinfo{booktitle}{\emph{2023 IEEE/ACM Symposium on Edge Computing (SEC)}}. \bibinfo{publisher}{IEEE Computer Society}, \bibinfo{address}{Los Alamitos, CA, USA}, \bibinfo{pages}{80--93}.
\newblock
\href{https://doi.org/10.1145/3583740.3628442}{doi:\nolinkurl{10.1145/3583740.3628442}}


\bibitem[Ultralytics(2021)]%
        {yolov5}
\bibfield{author}{\bibinfo{person}{Ultralytics}.} \bibinfo{year}{2021}\natexlab{}.
\newblock \bibinfo{title}{{{YOLOv5}: {A} state-of-the-art real-time object detection system}}.
\newblock \bibinfo{howpublished}{\url{https://docs.ultralytics.com}}.
\newblock
\newblock
\shownote{Accessed: insert date here}.


\bibitem[uTensor(2024)]%
        {utensor}
\bibfield{author}{\bibinfo{person}{uTensor}.} \bibinfo{year}{2024}\natexlab{}.
\newblock \bibinfo{title}{{uTensor}}.
\newblock \bibinfo{howpublished}{\url{https://github.com/uTensor/uTensor}}.
\newblock


\bibitem[Vaishnav et~al\mbox{.}(2023)]%
        {sparse3}
\bibfield{author}{\bibinfo{person}{Shubham Vaishnav}, \bibinfo{person}{Maria Efthymiou}, {and} \bibinfo{person}{Sindri Magnússon}.} \bibinfo{year}{2023}\natexlab{}.
\newblock \showarticletitle{{Energy-Efficient and Adaptive Gradient Sparsification for Federated Learning}}. In \bibinfo{booktitle}{\emph{ICC 2023 - IEEE International Conference on Communications}}. \bibinfo{pages}{1256--1261}.
\newblock
\href{https://doi.org/10.1109/ICC45041.2023.10278999}{doi:\nolinkurl{10.1109/ICC45041.2023.10278999}}


\bibitem[Van Der~Woude and Hicks(2016)]%
        {int1}
\bibfield{author}{\bibinfo{person}{Joel Van Der~Woude} {and} \bibinfo{person}{Matthew Hicks}.} \bibinfo{year}{2016}\natexlab{}.
\newblock \showarticletitle{{Intermittent computation without hardware support or programmer intervention}}. In \bibinfo{booktitle}{\emph{Proceedings of the 12th USENIX Conference on Operating Systems Design and Implementation}} (Savannah, GA, USA) \emph{(\bibinfo{series}{OSDI'16})}. \bibinfo{publisher}{USENIX Association}, \bibinfo{address}{USA}, \bibinfo{pages}{17–32}.
\newblock
\showISBNx{9781931971331}


\bibitem[van Kempen et~al\mbox{.}(2023)]%
        {mlonmcu}
\bibfield{author}{\bibinfo{person}{Philipp van Kempen}, \bibinfo{person}{Rafael Stahl}, \bibinfo{person}{Daniel Mueller-Gritschneder}, {and} \bibinfo{person}{Ulf Schlichtmann}.} \bibinfo{year}{2023}\natexlab{}.
\newblock \showarticletitle{{MLonMCU: TinyML Benchmarking with Fast Retargeting}}. In \bibinfo{booktitle}{\emph{Proceedings of the 2023 Workshop on Compilers, Deployment, and Tooling for Edge AI}} \emph{(\bibinfo{series}{CODAI ’23})}. \bibinfo{publisher}{ACM}.
\newblock
\href{https://doi.org/10.1145/3615338.3618128}{doi:\nolinkurl{10.1145/3615338.3618128}}


\bibitem[Wang et~al\mbox{.}(2019a)]%
        {split2}
\bibfield{author}{\bibinfo{person}{Dilin Wang}, \bibinfo{person}{Meng Li}, \bibinfo{person}{Lemeng Wu}, \bibinfo{person}{Vikas Chandra}, {and} \bibinfo{person}{Qiang Liu}.} \bibinfo{year}{2019}\natexlab{a}.
\newblock \showarticletitle{{Energy-Aware Neural Architecture Optimization with Fast Splitting Steepest Descent}}.
\newblock \bibinfo{journal}{\emph{CoRR}}  \bibinfo{volume}{abs/1910.03103} (\bibinfo{year}{2019}).
\newblock
\showeprint[arXiv]{1910.03103}
\urldef\tempurl%
\url{http://arxiv.org/abs/1910.03103}
\showURL{%
\tempurl}


\bibitem[Wang et~al\mbox{.}(2022)]%
        {devt5}
\bibfield{author}{\bibinfo{person}{Qipeng Wang}, \bibinfo{person}{Mengwei Xu}, \bibinfo{person}{Chao Jin}, \bibinfo{person}{Xinran Dong}, \bibinfo{person}{Jinliang Yuan}, \bibinfo{person}{Xin Jin}, \bibinfo{person}{Gang Huang}, \bibinfo{person}{Yunxin Liu}, {and} \bibinfo{person}{Xuanzhe Liu}.} \bibinfo{year}{2022}\natexlab{}.
\newblock \showarticletitle{Melon: breaking the memory wall for resource-efficient on-device machine learning}. In \bibinfo{booktitle}{\emph{Proceedings of the 20th Annual International Conference on Mobile Systems, Applications and Services}} (Portland, Oregon) \emph{(\bibinfo{series}{MobiSys '22})}. \bibinfo{publisher}{Association for Computing Machinery}, \bibinfo{address}{New York, NY, USA}, \bibinfo{pages}{450–463}.
\newblock
\showISBNx{9781450391856}
\href{https://doi.org/10.1145/3498361.3538928}{doi:\nolinkurl{10.1145/3498361.3538928}}


\bibitem[Wang et~al\mbox{.}(2019b)]%
        {agg}
\bibfield{author}{\bibinfo{person}{Shiqiang Wang}, \bibinfo{person}{Tiffany Tuor}, \bibinfo{person}{Theodoros Salonidis}, \bibinfo{person}{Kin~K. Leung}, \bibinfo{person}{Christian Makaya}, \bibinfo{person}{Ting He}, {and} \bibinfo{person}{Kevin Chan}.} \bibinfo{year}{2019}\natexlab{b}.
\newblock \showarticletitle{{Adaptive Federated Learning in Resource Constrained Edge Computing Systems}}.
\newblock \bibinfo{journal}{\emph{IEEE Journal on Selected Areas in Communications}} \bibinfo{volume}{37}, \bibinfo{number}{6} (\bibinfo{year}{2019}), \bibinfo{pages}{1205--1221}.
\newblock
\href{https://doi.org/10.1109/JSAC.2019.2904348}{doi:\nolinkurl{10.1109/JSAC.2019.2904348}}


\bibitem[Wang et~al\mbox{.}(2019c)]%
        {learning3}
\bibfield{author}{\bibinfo{person}{Wei Wang}, \bibinfo{person}{Vincent~W. Zheng}, \bibinfo{person}{Han Yu}, {and} \bibinfo{person}{Chunyan Miao}.} \bibinfo{year}{2019}\natexlab{c}.
\newblock \showarticletitle{A Survey of Zero-Shot Learning: Settings, Methods, and Applications}.
\newblock \bibinfo{journal}{\emph{ACM Trans. Intell. Syst. Technol.}} \bibinfo{volume}{10}, \bibinfo{number}{2}, Article \bibinfo{articleno}{13} (\bibinfo{date}{jan} \bibinfo{year}{2019}), \bibinfo{numpages}{37}~pages.
\newblock
\showISSN{2157-6904}
\href{https://doi.org/10.1145/3293318}{doi:\nolinkurl{10.1145/3293318}}


\bibitem[Wang et~al\mbox{.}(2017)]%
        {skipnet}
\bibfield{author}{\bibinfo{person}{Xin Wang}, \bibinfo{person}{Fisher Yu}, \bibinfo{person}{Zi{-}Yi Dou}, {and} \bibinfo{person}{Joseph~E. Gonzalez}.} \bibinfo{year}{2017}\natexlab{}.
\newblock \showarticletitle{{SkipNet: Learning Dynamic Routing in Convolutional Networks}}.
\newblock \bibinfo{journal}{\emph{CoRR}}  \bibinfo{volume}{abs/1711.09485} (\bibinfo{year}{2017}).
\newblock
\showeprint[arXiv]{1711.09485}
\urldef\tempurl%
\url{http://arxiv.org/abs/1711.09485}
\showURL{%
\tempurl}


\bibitem[Wang et~al\mbox{.}(2021)]%
        {patches}
\bibfield{author}{\bibinfo{person}{Yulin Wang}, \bibinfo{person}{Rui Huang}, \bibinfo{person}{Shiji Song}, \bibinfo{person}{Zeyi Huang}, {and} \bibinfo{person}{Gao Huang}.} \bibinfo{year}{2021}\natexlab{}.
\newblock \showarticletitle{{Not All Images are Worth 16x16 Words: Dynamic Vision Transformers with Adaptive Sequence Length}}.
\newblock \bibinfo{journal}{\emph{CoRR}}  \bibinfo{volume}{abs/2105.15075} (\bibinfo{year}{2021}).
\newblock
\showeprint[arXiv]{2105.15075}
\urldef\tempurl%
\url{https://arxiv.org/abs/2105.15075}
\showURL{%
\tempurl}


\bibitem[Watkins and Dayan(1992)]%
        {qlearning}
\bibfield{author}{\bibinfo{person}{Christopher J. C.~H. Watkins} {and} \bibinfo{person}{Peter Dayan}.} \bibinfo{year}{1992}\natexlab{}.
\newblock \showarticletitle{{Q-learning}}.
\newblock \bibinfo{journal}{\emph{Machine Learning}} \bibinfo{volume}{8}, \bibinfo{number}{3} (\bibinfo{date}{01 May} \bibinfo{year}{1992}), \bibinfo{pages}{279--292}.
\newblock
\showISSN{1573-0565}
\href{https://doi.org/10.1007/BF00992698}{doi:\nolinkurl{10.1007/BF00992698}}


\bibitem[Wei et~al\mbox{.}(2024)]%
        {sparse2}
\bibfield{author}{\bibinfo{person}{Kang Wei}, \bibinfo{person}{Jun Li}, \bibinfo{person}{Chuan Ma}, \bibinfo{person}{Ming Ding}, \bibinfo{person}{Feng Shu}, \bibinfo{person}{Haitao Zhao}, \bibinfo{person}{Wen Chen}, {and} \bibinfo{person}{Hongbo Zhu}.} \bibinfo{year}{2024}\natexlab{}.
\newblock \showarticletitle{{Gradient sparsification for efficient wireless federated learning with differential privacy}}.
\newblock \bibinfo{journal}{\emph{Science China Information Sciences}} \bibinfo{volume}{67}, \bibinfo{number}{4} (\bibinfo{date}{March} \bibinfo{year}{2024}).
\newblock
\showISSN{1869-1919}
\href{https://doi.org/10.1007/s11432-023-3918-9}{doi:\nolinkurl{10.1007/s11432-023-3918-9}}


\bibitem[Wiles et~al\mbox{.}(2021)]%
        {wiles}
\bibfield{author}{\bibinfo{person}{Olivia Wiles}, \bibinfo{person}{Sven Gowal}, \bibinfo{person}{Florian Stimberg}, \bibinfo{person}{Sylvestre{-}Alvise Rebuffi}, \bibinfo{person}{Ira Ktena}, \bibinfo{person}{Krishnamurthy Dvijotham}, {and} \bibinfo{person}{A.~Taylan Cemgil}.} \bibinfo{year}{2021}\natexlab{}.
\newblock \showarticletitle{{A Fine-Grained Analysis on Distribution Shift}}.
\newblock \bibinfo{journal}{\emph{CoRR}}  \bibinfo{volume}{abs/2110.11328} (\bibinfo{year}{2021}).
\newblock
\showeprint[arXiv]{2110.11328}
\urldef\tempurl%
\url{https://arxiv.org/abs/2110.11328}
\showURL{%
\tempurl}


\bibitem[Williams(1992)]%
        {reinforce}
\bibfield{author}{\bibinfo{person}{R.~J. Williams}.} \bibinfo{year}{1992}\natexlab{}.
\newblock \showarticletitle{{Simple statistical gradient-following algorithms for connectionist reinforcement learning}}.
\newblock \bibinfo{journal}{\emph{Machine Learning}}  \bibinfo{volume}{8} (\bibinfo{year}{1992}), \bibinfo{pages}{229--256}.
\newblock


\bibitem[Wolczyk et~al\mbox{.}(2021)]%
        {ztw}
\bibfield{author}{\bibinfo{person}{Maciej Wolczyk}, \bibinfo{person}{Bartosz W{\'{o}}jcik}, \bibinfo{person}{Klaudia Balazy}, \bibinfo{person}{Igor~T. Podolak}, \bibinfo{person}{Jacek Tabor}, \bibinfo{person}{Marek Smieja}, {and} \bibinfo{person}{Tomasz Trzcinski}.} \bibinfo{year}{2021}\natexlab{}.
\newblock \showarticletitle{{Zero Time Waste: Recycling Predictions in Early Exit Neural Networks}}.
\newblock \bibinfo{journal}{\emph{CoRR}}  \bibinfo{volume}{abs/2106.05409} (\bibinfo{year}{2021}).
\newblock
\showeprint[arXiv]{2106.05409}
\urldef\tempurl%
\url{https://arxiv.org/abs/2106.05409}
\showURL{%
\tempurl}


\bibitem[Wu et~al\mbox{.}(2019a)]%
        {fbnet}
\bibfield{author}{\bibinfo{person}{Bichen Wu}, \bibinfo{person}{Xiaoliang Dai}, \bibinfo{person}{Peizhao Zhang}, \bibinfo{person}{Yanghan Wang}, \bibinfo{person}{Fei Sun}, \bibinfo{person}{Yiming Wu}, \bibinfo{person}{Yuandong Tian}, \bibinfo{person}{Peter Vajda}, \bibinfo{person}{Yangqing Jia}, {and} \bibinfo{person}{Kurt Keutzer}.} \bibinfo{year}{2019}\natexlab{a}.
\newblock \bibinfo{title}{{FBNet: Hardware-Aware Efficient ConvNet Design via Differentiable Neural Architecture Search}}.
\newblock
\showeprint[arxiv]{1812.03443}~[cs.CV]


\bibitem[Wu et~al\mbox{.}(2019b)]%
        {blockdrop}
\bibfield{author}{\bibinfo{person}{Zuxuan Wu}, \bibinfo{person}{Tushar Nagarajan}, \bibinfo{person}{Abhishek Kumar}, \bibinfo{person}{Steven Rennie}, \bibinfo{person}{Larry~S. Davis}, \bibinfo{person}{Kristen Grauman}, {and} \bibinfo{person}{Rogerio Feris}.} \bibinfo{year}{2019}\natexlab{b}.
\newblock \bibinfo{title}{{BlockDrop: Dynamic Inference Paths in Residual Networks}}.
\newblock
\showeprint[arxiv]{1711.08393}~[cs.CV]


\bibitem[Xie et~al\mbox{.}(2020)]%
        {innout}
\bibfield{author}{\bibinfo{person}{Sang~Michael Xie}, \bibinfo{person}{Ananya Kumar}, \bibinfo{person}{Robbie Jones}, \bibinfo{person}{Fereshte Khani}, \bibinfo{person}{Tengyu Ma}, {and} \bibinfo{person}{Percy Liang}.} \bibinfo{year}{2020}\natexlab{}.
\newblock \showarticletitle{{In-N-Out: Pre-Training and Self-Training using Auxiliary Information for Out-of-Distribution Robustness}}.
\newblock \bibinfo{journal}{\emph{CoRR}}  \bibinfo{volume}{abs/2012.04550} (\bibinfo{year}{2020}).
\newblock
\showeprint[arXiv]{2012.04550}
\urldef\tempurl%
\url{https://arxiv.org/abs/2012.04550}
\showURL{%
\tempurl}


\bibitem[Xin et~al\mbox{.}(2020)]%
        {transformer2}
\bibfield{author}{\bibinfo{person}{Ji Xin}, \bibinfo{person}{Raphael Tang}, \bibinfo{person}{Jaejun Lee}, \bibinfo{person}{Yaoliang Yu}, {and} \bibinfo{person}{Jimmy Lin}.} \bibinfo{year}{2020}\natexlab{}.
\newblock \showarticletitle{{DeeBERT: Dynamic Early Exiting for Accelerating {BERT} Inference}}.
\newblock \bibinfo{journal}{\emph{CoRR}}  \bibinfo{volume}{abs/2004.12993} (\bibinfo{year}{2020}).
\newblock
\showeprint[arXiv]{2004.12993}
\urldef\tempurl%
\url{https://arxiv.org/abs/2004.12993}
\showURL{%
\tempurl}


\bibitem[Xin et~al\mbox{.}(2021)]%
        {transformer1}
\bibfield{author}{\bibinfo{person}{Ji Xin}, \bibinfo{person}{Raphael Tang}, \bibinfo{person}{Yaoliang Yu}, {and} \bibinfo{person}{Jimmy~J. Lin}.} \bibinfo{year}{2021}\natexlab{}.
\newblock \showarticletitle{{BERxiT: Early Exiting for BERT with Better Fine-Tuning and Extension to Regression}}. In \bibinfo{booktitle}{\emph{Conference of the European Chapter of the Association for Computational Linguistics}}.
\newblock
\urldef\tempurl%
\url{https://api.semanticscholar.org/CorpusID:233189542}
\showURL{%
\tempurl}


\bibitem[Xu et~al\mbox{.}(2020)]%
        {deepwear}
\bibfield{author}{\bibinfo{person}{Mengwei Xu}, \bibinfo{person}{Feng Qian}, \bibinfo{person}{Mengze Zhu}, \bibinfo{person}{Feifan Huang}, \bibinfo{person}{Saumay Pushp}, {and} \bibinfo{person}{Xuanzhe Liu}.} \bibinfo{year}{2020}\natexlab{}.
\newblock \showarticletitle{{DeepWear: Adaptive Local Offloading for On-Wearable Deep Learning}}.
\newblock \bibinfo{journal}{\emph{IEEE Transactions on Mobile Computing}} \bibinfo{volume}{19}, \bibinfo{number}{2} (\bibinfo{year}{2020}), \bibinfo{pages}{314--330}.
\newblock
\href{https://doi.org/10.1109/TMC.2019.2893250}{doi:\nolinkurl{10.1109/TMC.2019.2893250}}


\bibitem[Yang et~al\mbox{.}(2021)]%
        {sched4}
\bibfield{author}{\bibinfo{person}{Fan Yang}, \bibinfo{person}{Ashok~Samraj Thangarajan}, \bibinfo{person}{Gowri~Sankar Ramachandran}, \bibinfo{person}{Wouter Joosen}, {and} \bibinfo{person}{Danny Hughes}.} \bibinfo{year}{2021}\natexlab{}.
\newblock \showarticletitle{{AsTAR: Sustainable Energy Harvesting for the Internet of Things through Adaptive Task Scheduling}}.
\newblock \bibinfo{journal}{\emph{ACM Trans. Sen. Netw.}} \bibinfo{volume}{18}, \bibinfo{number}{1}, Article \bibinfo{articleno}{4} (\bibinfo{date}{oct} \bibinfo{year}{2021}), \bibinfo{numpages}{34}~pages.
\newblock
\showISSN{1550-4859}
\href{https://doi.org/10.1145/3467894}{doi:\nolinkurl{10.1145/3467894}}


\bibitem[Yang et~al\mbox{.}(2016a)]%
        {eap}
\bibfield{author}{\bibinfo{person}{Tien{-}Ju Yang}, \bibinfo{person}{Yu{-}Hsin Chen}, {and} \bibinfo{person}{Vivienne Sze}.} \bibinfo{year}{2016}\natexlab{a}.
\newblock \showarticletitle{{Designing Energy-Efficient Convolutional Neural Networks using Energy-Aware Pruning}}.
\newblock \bibinfo{journal}{\emph{CoRR}}  \bibinfo{volume}{abs/1611.05128} (\bibinfo{year}{2016}).
\newblock
\showeprint[arXiv]{1611.05128}
\urldef\tempurl%
\url{http://arxiv.org/abs/1611.05128}
\showURL{%
\tempurl}


\bibitem[Yang et~al\mbox{.}(2016b)]%
        {tien_ju1}
\bibfield{author}{\bibinfo{person}{Tien{-}Ju Yang}, \bibinfo{person}{Yu{-}Hsin Chen}, {and} \bibinfo{person}{Vivienne Sze}.} \bibinfo{year}{2016}\natexlab{b}.
\newblock \showarticletitle{{Designing Energy-Efficient Convolutional Neural Networks using Energy-Aware Pruning}}.
\newblock \bibinfo{journal}{\emph{CoRR}}  \bibinfo{volume}{abs/1611.05128} (\bibinfo{year}{2016}).
\newblock
\showeprint[arXiv]{1611.05128}
\urldef\tempurl%
\url{http://arxiv.org/abs/1611.05128}
\showURL{%
\tempurl}


\bibitem[Yang et~al\mbox{.}(2017a)]%
        {eems}
\bibfield{author}{\bibinfo{person}{Tien-Ju Yang}, \bibinfo{person}{Yu-Hsin Chen}, \bibinfo{person}{Joel Emer}, {and} \bibinfo{person}{Vivienne Sze}.} \bibinfo{year}{2017}\natexlab{a}.
\newblock \showarticletitle{{A method to estimate the energy consumption of deep neural networks}}. In \bibinfo{booktitle}{\emph{2017 51st Asilomar Conference on Signals, Systems, and Computers}}. \bibinfo{pages}{1916--1920}.
\newblock
\href{https://doi.org/10.1109/ACSSC.2017.8335698}{doi:\nolinkurl{10.1109/ACSSC.2017.8335698}}


\bibitem[Yang et~al\mbox{.}(2017b)]%
        {tien_ju2}
\bibfield{author}{\bibinfo{person}{Tien-Ju Yang}, \bibinfo{person}{Yu-Hsin Chen}, \bibinfo{person}{Joel Emer}, {and} \bibinfo{person}{Vivienne Sze}.} \bibinfo{year}{2017}\natexlab{b}.
\newblock \showarticletitle{{A method to estimate the energy consumption of deep neural networks}}. In \bibinfo{booktitle}{\emph{2017 51st Asilomar Conference on Signals, Systems, and Computers}}. \bibinfo{pages}{1916--1920}.
\newblock
\href{https://doi.org/10.1109/ACSSC.2017.8335698}{doi:\nolinkurl{10.1109/ACSSC.2017.8335698}}


\bibitem[Yang et~al\mbox{.}(2019)]%
        {zhaohui}
\bibfield{author}{\bibinfo{person}{Zhaohui Yang}, \bibinfo{person}{Mingzhe Chen}, \bibinfo{person}{Walid Saad}, \bibinfo{person}{Choong~Seon Hong}, {and} \bibinfo{person}{Mohammad Shikh{-}Bahaei}.} \bibinfo{year}{2019}\natexlab{}.
\newblock \showarticletitle{{Energy Efficient Federated Learning Over Wireless Communication Networks}}.
\newblock \bibinfo{journal}{\emph{CoRR}}  \bibinfo{volume}{abs/1911.02417} (\bibinfo{year}{2019}).
\newblock
\showeprint[arXiv]{1911.02417}
\urldef\tempurl%
\url{http://arxiv.org/abs/1911.02417}
\showURL{%
\tempurl}


\bibitem[Yao et~al\mbox{.}(2020)]%
        {off}
\bibfield{author}{\bibinfo{person}{Shuochao Yao}, \bibinfo{person}{Jinyang Li}, \bibinfo{person}{Dongxin Liu}, \bibinfo{person}{Tianshi Wang}, \bibinfo{person}{Shengzhong Liu}, \bibinfo{person}{Huajie Shao}, {and} \bibinfo{person}{Tarek Abdelzaher}.} \bibinfo{year}{2020}\natexlab{}.
\newblock \showarticletitle{Deep compressive offloading: speeding up neural network inference by trading edge computation for network latency}. In \bibinfo{booktitle}{\emph{Proceedings of the 18th Conference on Embedded Networked Sensor Systems}} (Virtual Event, Japan) \emph{(\bibinfo{series}{SenSys '20})}. \bibinfo{publisher}{Association for Computing Machinery}, \bibinfo{address}{New York, NY, USA}, \bibinfo{pages}{476–488}.
\newblock
\showISBNx{9781450375900}
\href{https://doi.org/10.1145/3384419.3430898}{doi:\nolinkurl{10.1145/3384419.3430898}}


\bibitem[Yeom et~al\mbox{.}(2021)]%
        {nuke_norm}
\bibfield{author}{\bibinfo{person}{Seul{-}Ki Yeom}, \bibinfo{person}{Kyung{-}Hwan Shim}, {and} \bibinfo{person}{Jee{-}Hyun Hwang}.} \bibinfo{year}{2021}\natexlab{}.
\newblock \showarticletitle{{Toward Compact Deep Neural Networks via Energy-Aware Pruning}}.
\newblock \bibinfo{journal}{\emph{CoRR}}  \bibinfo{volume}{abs/2103.10858} (\bibinfo{year}{2021}).
\newblock
\showeprint[arXiv]{2103.10858}
\urldef\tempurl%
\url{https://arxiv.org/abs/2103.10858}
\showURL{%
\tempurl}


\bibitem[Yousefzadeh et~al\mbox{.}(2022)]%
        {alu}
\bibfield{author}{\bibinfo{person}{Amirreza Yousefzadeh}, \bibinfo{person}{Jan Stuijt}, \bibinfo{person}{Martijn Hijdra}, \bibinfo{person}{Hsiao-Hsuan Liu}, \bibinfo{person}{Anteneh Gebregiorgis}, \bibinfo{person}{Abhairaj Singh}, \bibinfo{person}{Said Hamdioui}, {and} \bibinfo{person}{Francky Catthoor}.} \bibinfo{year}{2022}\natexlab{}.
\newblock \showarticletitle{{Energy-efficient In-Memory Address Calculation}}.
\newblock \bibinfo{journal}{\emph{ACM Trans. Archit. Code Optim.}} \bibinfo{volume}{19}, \bibinfo{number}{4}, Article \bibinfo{articleno}{52} (\bibinfo{date}{sep} \bibinfo{year}{2022}), \bibinfo{numpages}{16}~pages.
\newblock
\showISSN{1544-3566}
\href{https://doi.org/10.1145/3546071}{doi:\nolinkurl{10.1145/3546071}}


\bibitem[Yu et~al\mbox{.}(2018)]%
        {slimmable}
\bibfield{author}{\bibinfo{person}{Jiahui Yu}, \bibinfo{person}{Linjie Yang}, \bibinfo{person}{Ning Xu}, \bibinfo{person}{Jianchao Yang}, {and} \bibinfo{person}{Thomas~S. Huang}.} \bibinfo{year}{2018}\natexlab{}.
\newblock \showarticletitle{{Slimmable Neural Networks}}.
\newblock \bibinfo{journal}{\emph{CoRR}}  \bibinfo{volume}{abs/1812.08928} (\bibinfo{year}{2018}).
\newblock
\showeprint[arXiv]{1812.08928}
\urldef\tempurl%
\url{http://arxiv.org/abs/1812.08928}
\showURL{%
\tempurl}


\bibitem[Yu and Li(2021)]%
        {rong}
\bibfield{author}{\bibinfo{person}{Rong Yu} {and} \bibinfo{person}{Peichun Li}.} \bibinfo{year}{2021}\natexlab{}.
\newblock \showarticletitle{{Toward Resource-Efficient Federated Learning in Mobile Edge Computing}}.
\newblock \bibinfo{journal}{\emph{IEEE Network}} \bibinfo{volume}{35}, \bibinfo{number}{1} (\bibinfo{year}{2021}), \bibinfo{pages}{148--155}.
\newblock
\href{https://doi.org/10.1109/MNET.011.2000295}{doi:\nolinkurl{10.1109/MNET.011.2000295}}


\bibitem[Yuan et~al\mbox{.}(2024)]%
        {qfedupdate}
\bibfield{author}{\bibinfo{person}{Jinliang Yuan}, \bibinfo{person}{Shangguang Wang}, \bibinfo{person}{Hongyu Li}, \bibinfo{person}{Daliang Xu}, \bibinfo{person}{Yuanchun Li}, \bibinfo{person}{Mengwei Xu}, {and} \bibinfo{person}{Xuanzhe Liu}.} \bibinfo{year}{2024}\natexlab{}.
\newblock \showarticletitle{{Towards Energy-efficient Federated Learning via INT8-based Training on Mobile DSPs}}. In \bibinfo{booktitle}{\emph{Proceedings of the ACM Web Conference 2024}} (Singapore, Singapore) \emph{(\bibinfo{series}{WWW '24})}. \bibinfo{publisher}{Association for Computing Machinery}, \bibinfo{address}{New York, NY, USA}, \bibinfo{pages}{2786–2794}.
\newblock
\showISBNx{9798400701719}
\href{https://doi.org/10.1145/3589334.3645341}{doi:\nolinkurl{10.1145/3589334.3645341}}


\bibitem[Zela et~al\mbox{.}(2020)]%
        {fs2}
\bibfield{author}{\bibinfo{person}{Arber Zela}, \bibinfo{person}{Julien Siems}, {and} \bibinfo{person}{Frank Hutter}.} \bibinfo{year}{2020}\natexlab{}.
\newblock \showarticletitle{NAS-Bench-1Shot1: Benchmarking and Dissecting One-shot Neural Architecture Search}.
\newblock \bibinfo{journal}{\emph{CoRR}}  \bibinfo{volume}{abs/2001.10422} (\bibinfo{year}{2020}).
\newblock
\showeprint[arXiv]{2001.10422}
\urldef\tempurl%
\url{https://arxiv.org/abs/2001.10422}
\showURL{%
\tempurl}


\bibitem[Zeng et~al\mbox{.}(2019b)]%
        {boomerang}
\bibfield{author}{\bibinfo{person}{Liekang Zeng}, \bibinfo{person}{En Li}, \bibinfo{person}{Zhi Zhou}, {and} \bibinfo{person}{Xu Chen}.} \bibinfo{year}{2019}\natexlab{b}.
\newblock \showarticletitle{{Boomerang: On-Demand Cooperative Deep Neural Network Inference for Edge Intelligence on the Industrial Internet of Things}}.
\newblock \bibinfo{journal}{\emph{IEEE Network}} \bibinfo{volume}{33}, \bibinfo{number}{5} (\bibinfo{year}{2019}), \bibinfo{pages}{96--103}.
\newblock
\href{https://doi.org/10.1109/MNET.001.1800506}{doi:\nolinkurl{10.1109/MNET.001.1800506}}


\bibitem[Zeng et~al\mbox{.}(2019a)]%
        {radio}
\bibfield{author}{\bibinfo{person}{Qunsong Zeng}, \bibinfo{person}{Yuqing Du}, \bibinfo{person}{Kin~K. Leung}, {and} \bibinfo{person}{Kaibin Huang}.} \bibinfo{year}{2019}\natexlab{a}.
\newblock \bibinfo{title}{{Energy-Efficient Radio Resource Allocation for Federated Edge Learning}}.
\newblock
\showeprint[arxiv]{1907.06040}~[cs.IT]


\bibitem[Zhang et~al\mbox{.}(2022)]%
        {growth}
\bibfield{author}{\bibinfo{person}{Yu Zhang}, \bibinfo{person}{Tao Gu}, {and} \bibinfo{person}{Xi Zhang}.} \bibinfo{year}{2022}\natexlab{}.
\newblock \showarticletitle{{MDLdroidLite: A Release-and-Inhibit Control Approach to Resource-Efficient Deep Neural Networks on Mobile Devices}}.
\newblock \bibinfo{journal}{\emph{IEEE Transactions on Mobile Computing}} \bibinfo{volume}{21}, \bibinfo{number}{10} (\bibinfo{year}{2022}), \bibinfo{pages}{3670--3686}.
\newblock
\href{https://doi.org/10.1109/TMC.2021.3062575}{doi:\nolinkurl{10.1109/TMC.2021.3062575}}


\bibitem[Zhao et~al\mbox{.}(2022)]%
        {underwater}
\bibfield{author}{\bibinfo{person}{Yuchen Zhao}, \bibinfo{person}{Sayed~Saad Afzal}, \bibinfo{person}{Waleed Akbar}, \bibinfo{person}{Osvy Rodriguez}, \bibinfo{person}{Fan Mo}, \bibinfo{person}{David Boyle}, \bibinfo{person}{Fadel Adib}, {and} \bibinfo{person}{Hamed Haddadi}.} \bibinfo{year}{2022}\natexlab{}.
\newblock \showarticletitle{{Towards battery-free machine learning and inference in underwater environments}}. In \bibinfo{booktitle}{\emph{Proceedings of the 23rd Annual International Workshop on Mobile Computing Systems and Applications}} \emph{(\bibinfo{series}{HotMobile ’22})}. \bibinfo{publisher}{ACM}.
\newblock
\href{https://doi.org/10.1145/3508396.3512877}{doi:\nolinkurl{10.1145/3508396.3512877}}


\bibitem[Zhao et~al\mbox{.}(2020)]%
        {fs1}
\bibfield{author}{\bibinfo{person}{Yiyang Zhao}, \bibinfo{person}{Linnan Wang}, \bibinfo{person}{Yuandong Tian}, \bibinfo{person}{Rodrigo Fonseca}, {and} \bibinfo{person}{Tian Guo}.} \bibinfo{year}{2020}\natexlab{}.
\newblock \showarticletitle{{Few-shot Neural Architecture Search}}.
\newblock \bibinfo{journal}{\emph{CoRR}}  \bibinfo{volume}{abs/2006.06863} (\bibinfo{year}{2020}).
\newblock
\showeprint[arXiv]{2006.06863}
\urldef\tempurl%
\url{https://arxiv.org/abs/2006.06863}
\showURL{%
\tempurl}


\bibitem[Zhao et~al\mbox{.}(2018)]%
        {deepthings}
\bibfield{author}{\bibinfo{person}{Zhuoran Zhao}, \bibinfo{person}{Kamyar~Mirzazad Barijough}, {and} \bibinfo{person}{Andreas Gerstlauer}.} \bibinfo{year}{2018}\natexlab{}.
\newblock \showarticletitle{{DeepThings: Distributed Adaptive Deep Learning Inference on Resource-Constrained IoT Edge Clusters}}.
\newblock \bibinfo{journal}{\emph{IEEE Transactions on Computer-Aided Design of Integrated Circuits and Systems}} \bibinfo{volume}{37}, \bibinfo{number}{11} (\bibinfo{year}{2018}), \bibinfo{pages}{2348--2359}.
\newblock
\href{https://doi.org/10.1109/TCAD.2018.2858384}{doi:\nolinkurl{10.1109/TCAD.2018.2858384}}


\bibitem[Zhu et~al\mbox{.}(2020)]%
        {signsgd}
\bibfield{author}{\bibinfo{person}{Guangxu Zhu}, \bibinfo{person}{Yong Wang}, {and} \bibinfo{person}{Kaibin Huang}.} \bibinfo{year}{2020}\natexlab{}.
\newblock \showarticletitle{{Broadband Analog Aggregation for Low-Latency Federated Edge Learning}}.
\newblock \bibinfo{journal}{\emph{IEEE Transactions on Wireless Communications}} \bibinfo{volume}{19}, \bibinfo{number}{1} (\bibinfo{year}{2020}), \bibinfo{pages}{491--506}.
\newblock
\href{https://doi.org/10.1109/TWC.2019.2946245}{doi:\nolinkurl{10.1109/TWC.2019.2946245}}


\end{thebibliography}

\end{document}